# A multi-strategy improved snake optimizer for three-dimensional UAV path planning and engineering problems

Genliang Li[1,2,*], Yaxin Cui[1,*], Jinyu Su[1,2]

[1]College of Computer and Data Science, Putian University, Putian 351100, Fujian, China

[2]New Engineering Industry College, Putian University, Putian 351100, Fujian, China

*These authors contributed to the work equally and should be regarded as co-first authors;

Corresponding author：Yaxin Cui.

Genliang Li (Email: humrry@foxmail.com)

Yaxin Cui (Email: cuiyaxin19@mails.ucas.edu.cn)

Jinyu Su (Email: sujingyu@bit.edu.cn)

**Abstract**

Metaheuristic algorithms have gained widespread application across various fields owing to their ability to generate diverse solutions. One such algorithm is the Snake Optimizer (SO), a progressive optimization approach. However, SO suffers from the issues of slow convergence speed and susceptibility to local optima. In light of these shortcomings, we propose a novel Multi-strategy Improved Snake Optimizer (MISO). Firstly, we propose a new adaptive random disturbance strategy based on sine function to alleviate the risk of getting trapped in a local optimum. Secondly, we introduce adaptive Levy flight strategy based on scale factor and leader and endow the male snake leader with flight capability, which makes it easier for the algorithm to leap out of the local optimum and find the global optimum. More importantly, we put forward a position update strategy combining elite leadership and Brownian motion, effectively accelerating the convergence speed while ensuring precision. Finally, to demonstrate the performance of MISO, we utilize 30 CEC2017 test functions and the CEC2022 test suite, comparing it with 11 popular algorithms across different dimensions to validate its effectiveness. Moreover, Unmanned Aerial Vehicle (UAV) has been widely used in various fields due to its advantages of low cost, high mobility and easy operation. However, the UAV path planning problem is crucial for flight safety and efficiency, and there are still challenges in establishing and optimizing the path model. Therefore, we apply MISO to the UAV 3D path planning problem as well as 6 engineering design problems to assess its feasibility in practical applications. The experimental results demonstrate that MISO exceeds other competitive algorithms in terms of solution quality and stability, establishing its strong potential for application.

## 1. Introduction

With the advancement of society, optimization problems are prevalent across various real-world fields [1]. However, when dealing with complex nonlinear, non-convex, and non-differentiable problems, traditional mathematical methods face challenges in finding solutions, often leading to entrapment in local optima, which fail to meet the existing demands [2]. The utilization of metaheuristic algorithms to address optimization problems has garnered significant attention from researchers [3]. This is due to several advantages offered by metaheuristic optimization algorithms, including simplicity, flexibility, lack of reliance on derivatives, and the ability to avoid local optima [4].

Metaheuristic algorithms are proposed by mimicking natural phenomena [5]. They are broadly categorized into four groups [6]: Evolutionary Algorithms (EA), Physics-based Algorithms (PhA), Human-based algorithms, and Swarm Intelligence (SI) algorithms. EA is a probabilistic optimization method inspired by the principles of natural evolution. The most popular EA is the Genetic Algorithm (GA), which was presented by Holland [7] and draws inspiration from Darwin's theory of evolution. PhA is usually inspired by physical phenomena, with the Simulated Annealing (SA) algorithm proposed by Kirkpatrick et al. [8] being a classical example, based on the principle of solid annealing in metallurgy. Human-based algorithms are generally enlightened by human activity [9]. For example, Kumar et al. [10] proposed the Social Evolution and Learning Optimization Algorithm (SELOA) based on human social learning pattern. SI algorithms draw inspiration from the social behavior of miscellaneous biological organizations in nature [11]. The Particle Swarm Optimization (PSO) algorithm developed by Eberhart and Kennedy [12] takes inspiration from the foraging pattern of birds, which abstracts birds as particles randomly searching for food (objective function) in a forest (search space). In each iteration, the movement of each particle considers its own optimal value and the best solution obtained by the swarm. Eventually, the particles will gather to the position where the food is the most (optimal solution).

The SI algorithm, as a highly representative branch of metaheuristic algorithms, has experienced rapid development over the past two decades [13]. For example, Dorigo et al. [14] introduced the Ant Colony Optimization (ACO) algorithm, which simulating the foraging behavior of the ant colony. The Artificial Bee Colony (ABC) [15] algorithm takes inspiration from the honey-collecting behavior of bees. Social Spider Optimization (SSO) was proposed to emulate the cooperative behavior of spiders [16]. Cuckoo Search (CS) primarily imitates the living behavior of birds [17]. Bansal et al. [18] developed the Spider Monkey Optimization (SMO) algorithm by drawing from the social structure of spider monkeys during their foraging process. The Grey Wolf Optimizer (GWO) was inspired by the hunting activity of grey wolf [19]. Bird Swarm Algorithm (BSA) was proposed by learning the social performance of bird [20]. The Dragonfly Algorithm (DA) developed by Mirjalili [21], draws its main inspiration from the collective behavior of dragonflies. Similarly, the Whale Optimization Algorithm (WOA) was presented by Mirjalili et al. [4] , simulates the behavior of humpback whales, while the Sperm Whale Algorithm (SWA) takes cues from the living habits of sperm whales [22]. In the Crow Search Algorithm (CSA), social activities of crows are imitated [23], Jangir et al.[24] presented the Moth-Flame Optimization (MFO) algorithm, based on the navigation of moths in nature. The Rhinoceros Search Algorithm (RSA) primarily simulates the behavior characteristics of rhinos [25]. Fox Hunting Algorithm (FHA) mainly imitates the hunting activities of fox [26]. The Salp Swarm Algorithm (SSA) proposed by Mirjalili et al. [27], finds inspiration in the survival behavior of salps in the ocean. Grasshopper Optimization Algorithm (GOA) learns from the foraging habits of grasshoppers [28]. Wang et al. [29] presented Earthworm Optimisation Algorithm (EWA) according to the contribution of earthworms in nature. The Emperor Penguin Optimizer (EPO) takes inspiration from the behavior of penguins [30], and the Moth Search (MS) algorithm, developed by Wang [31], leverages the unique performance of moths. The Squirrel Search Algorithm (SSA) was proposed by simulating the living habits of squirrels [32], while the Harris Hawks Optimization (HHO) algorithm, displayed by Heidari et al. [33], is based on the behavioral characteristics of Harris hawks in nature. The Mayfly Algorithm (MA) was proposed by imitating the natural performance of mayflies [34]. The Tunicate Swarm Algorithm (TSA) proposed by Kaur et al. [35], draws inspiration from the living habits of tunicates. The Black Widow Optimization (BWO) algorithm was inspired by the reproduction mode of the black widow spider [36], Chimp Optimization Algorithm (ChOA) inspired by the special behavior of chimpanzee [37], and the Marine Predators Algorithm (MPA) was presented by Faramarzi et al. [38], this approach predominantly simulated the special hunting performance of marine predators. The Sandpiper Optimization Algorithm (SOA) is mainly inspired by the sandpiper attack and migration behavior [39], and the Rat Swarm Optimizer (RSO) was developed by Dhiman et al. [40], takes its cues from the living behavior of mice. The Horse herd Optimization Algorithm (HOA) proposed by MiarNaeimi et al. [41] takes inspiration from the specific habit of the horse herd. The African Vultures Optimization Algorithm (AVOA) was presented by Abdollahzadeh et al. [42] based on the living habits of vultures. The artificial Jellyfish Search (JS) optimizer simulates the behavior of jellyfish in the ocean [43], while the Chameleon Swarm Algorithm (CSA) draws inspiration from chameleon hunting behavior [44]. The Golden Eagle Optimizer (GEO) proposed by Mohammadi et al. [45], takes inspiration from the intelligent hunting performance of golden eagles, the Pelican Optimization Algorithm (POA) derives its design from the intelligent hunting behavior of pelicans [46], Honey Badger Algorithm (HBA) is mainly proposed based on the foraging behavior of honey badgers [6]. The Artificial Hummingbird Algorithm (AHA) is primarily inspired by the special activity of hummingbirds [47]. Abualigah et al. [48] presented the Reptile Search Algorithm (RSA) based on the behavior of reptiles, and the main design idea of Sand Cat Swarm Optimization (SCSO) is to imitate the living habits of sand cats [49]. Wang et al. [50] designed the Artificial Rabbits Optimization (ARO) algorithm according to the living habits of rabbits, the Prairie Dog Optimization (PDO) was mainly designed based on the intelligent behavior of prairie dogs [51], and Zamani et al. [52] developed the Starling Murmuration Optimizer (SMO) inspired by the social activities of starlings. The White Shark Optimizer (WSO) was designed based on the foraging activity of the great white shark [53]. Piranha Foraging Optimization Algorithm (PFOA) is designed to study the survival mode of piranha [54]. Based on the living habits of Dung beetle, Xue et al. [55] developed the Dung Beetle Optimizer (DBO), and the Jumping Spider Optimization Algorithm (JSOA) is proposed based on the survival mode of spider [56]. The Ebola Optimization Search Algorithm (EOSA) is primarily inspired by the transmission mode of ebola virus [57]. Agushaka et al. [58] designed the Dwarf Mongoose Optimization (DMO) algorithm according to the life style of the dwarf mongoose. The Snake Optimizer (SO) is designed to simulate the special behavior of snakes [59]. The inspiration of Golden Jackal Optimization (GJO) algorithm comes from the hunting process of golden jackals [60], the Beluga Whale Optimization (BWO) is proposed by Zhong et al. [61] to simulate the intelligent performance of whale, the Nutcracker Optimization Algorithm (NOA) draws its design inspiration from the two different survival modes of nutcrackers [62]. Regardless of the differences in inspiration and definition of these algorithms, they all have a common essential feature, dividing the search process into two stages [63]: exploration and exploitation. The exploration phase means that the algorithm explores various areas in the search space to discover as many promising regions as possible. The exploitation stage focuses on continuously searching for the optimal solution within the potential areas identified during the exploration phase. Striking the right balance between these two phases poses a challenging task.

In addition, the NFL theorem [64] logically demonstrates that no single algorithm can excel in solving all optimization problems simultaneously. This motivates many scholars to continuously enhance existing algorithms to address optimization problems [65, 66]. For instance, Rezaei et al. [67] developed a Guided Adaptive Search-based Particle Swarm Optimizer (GuASPSO), and Sedighizadeh et al. [68] presented a Generalized Particle Swarm Optimization (GEPSO) algorithm. Zhang et al. [69] designed the Dynamic Grey Wolf Optimizer (DGWO) to eliminate waiting times among grey wolves, resulting in accelerated convergence. Ma et al. [70] developed an improved GWO algorithm (AGWO) by giving the grey wolf flight ability to enhance exploration. Fan et al. [71] proposed the Beetle antenna-based Grey Wolf Optimization (BGWO) method by augmenting the hearing capability of the leader grey wolf to improve global exploration. Li et al. [72] presented the Modified Whale Optimization Algorithm (MWOA) integrating chaotic mapping and a selection method. Guo et al. [73] introduced an improved Whale Optimization Algorithm (PAWOA) by using a new strategy of random jump change method and random control parameter strategy. Ding et al. [74] designed an improved Whale Optimization Algorithm (LNAWOA) by utilizing three strategies: chaos initialization, nonlinear control factor, and chaos inertia weight. Fan et al. [65, 66] proposed a Modified Marine Predator Algorithm (MMPA) based on opposition-based learning, and designed a modified Equilibrium Optimizer (m-EO) based on reverse learning and a new update method.

Recently, Hashim et al. [59] proposed the Snake Optimizer (SO) by simulating the living habits of snakes, and compared it with 11 competitive algorithms, including L-SHADE [75], LSHADE-EpSin [76], CMAES [77], COA [78], MFO [24], HHO [33], TEO [79], GOA [28] and WOA [4]. SO shows strong competitiveness. However, the NFL theorem [64] encourages us to improve this new population-based optimizer. Although SO has advantages over other competitors, it still suffers from slow convergence and susceptibility to local optima.

In view of the aforementioned shortcomings of the SO, we develop a Multi-strategy Improved Snake Optimizer (MISO). The main contributions of this manuscript are as follows:

1) A stochastic disturbance strategy based on sine function is introduced to make it easier for the algorithm to jump out of the local optimum.
2) Adaptive Levy flight strategy based on scale factor and leader is introduced to give the male snake flight ability to enhance the exploration capability of the algorithm.
3) Adaptive position update strategy combining elite leadership and Brownian motion is proposed to enhance the exploitation of the algorithm.
4) To evaluate the performance of MISO, we compare it with 11 popular algorithms on the CEC2017 and CEC2022 test suites.
5) The effectiveness and accuracy of MISO in optimization applications are verified by 3D UAV trajectory planning and 6 engineering application problems.

The rest of the manuscript is structured as follows. A brief description of the SO is given in Section 2. Section 3 presents the strategy adopted by MISO. In Section 4, we carried out the corresponding simulation experiments and detailed analysis. In Section 5, we apply MISO to solve 3D UAV path planning and 6 engineering design problems. Finally, Section 6 concludes the paper with a summary and prospect.

## 2. Snake Optimizer

This section provides a succinct description of the inspiration and mathematical model of SO.

### 2.1 The inspiration of SO

The snake's unique mating pattern is influenced by various factors, and mating between male and female snakes typically occurs during late spring and early summer. The mating process is dependent not only on temperature but also on food availability. When the temperature is suitable and food is sufficient, the male snakes attract female attention through fighting. Following successful mating, female snakes lay eggs in a nest and promptly depart. Drawing inspiration from snake mating behavior, the Snake Optimizer (SO) divides its search process into two stages: exploration and exploitation. In the exploration phase, the snake seeks a comfortable environment with plentiful food. During the exploitation stage, several transitional periods exist. If the food is sufficient but the temperature is high, snakes focus solely on consuming available food. On the other hand, if food is ample and the temperature is low, the mating process ensues. This mating process comprises two situations: fighting and mating mode. Finally, the female snakes may lay eggs and hatch them into new snakes [80, 81]. Fig 1 shows the mating process of snakes.

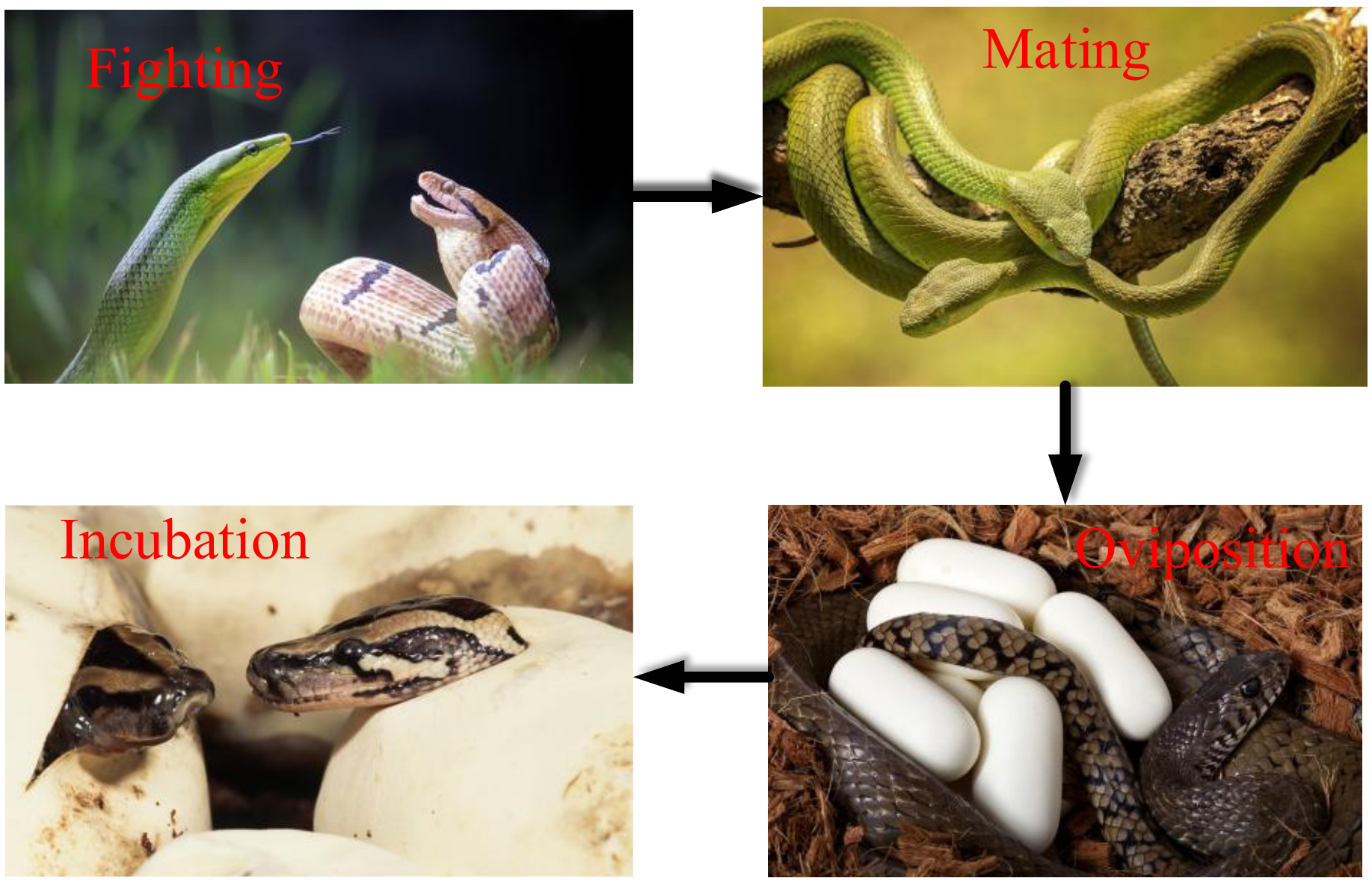

Fig 1. Schematic diagram of the mating process of snakes.

### 2.2 The mathematical model of SO

This subsection introduces the mathematical model of SO.

#### 2.2.1 Initialization process

The initial population of snakes is shown in Eq. (1).

$$X_i = X_{min} + r \times (X_{max} - X_{min}) \tag{1}$$

Where $X_i$ represents the position of the $i^{th}$ snake, $X_{min}$ and $X_{max}$ are the lower and upper bounds, respectively, and $r$ denotes a random number between 0 and 1. The snake population is equally divided into male and female parts, as reported in Eq. (2) and Eq. (3).

$$N_m \approx \frac{N}{2} \tag{2}$$

$$N_f = N - N_m \tag{3}$$

Where $N$ is the population size, $N_m$ represents the number of male snakes, and $N_f$ denotes the number of female snakes.

Whether the snakes mate depends on the adequacy of food and the level of temperature. Therefore, in SO, the search process is affected by food and temperature. Food and temperature are defined as Eq. (4) and Eq. (5).

$$Temp = exp\left(\frac{-t}{T}\right) \tag{4}$$

$$Q = c_1 \times \exp\left(\frac{t-T}{T}\right) \tag{5}$$

Where $t$ denotes the current iteration number, $T$ indicates the maximum iteration number, and $c_1$=0.5.

2.2.2 Exploration process

When the amount of food $Q$ < 0.25, snakes randomly select locations in the search space to update their locations in order to find food. The update formula of male and female snakes is described by Eq. (6).

$$\begin{aligned} X_{i,m}(t+1) &= X_{rand,m}(t) \pm c_2 \times A_m \times \left((X_{max} - X_{min}) \times rand + X_{min}\right) \\ X_{i,f}(t+1) &= X_{rand,f}(t) \pm c_2 \times A_f \times \left((X_{max} - X_{min}) \times rand + X_{min}\right) \end{aligned} \tag{6}$$

Where $X_{i,m}$ and $X_{i,f}$ represent the position of the $i^{th}$ male and female snake, respectively. $X_{rand,m}$ and $X_{rand,f}$ indicate the random individual of male and female snakes, respectively. $rand$ refers to a random number from 0 to 1. $X_{min}$ and $X_{max}$ are the lower and upper bounds, respectively. $c_2$ = 0.05. $A_m$ and $A_f$ denote the ability of male and female snakes to find food, respectively. The calculation formula is expressed as Eq. (7).

$$\begin{aligned} A_m &= exp\left(\frac{-f_{rand,m}}{f_{i,m}}\right) \\ A_f &= exp\left(\frac{-f_{rand,f}}{f_{i,f}}\right) \end{aligned} \tag{7}$$

Where $f_{rand,m}$ and $f_{rand,f}$ denote the fitness of random individual in male and female snakes, respectively, and $f_{i,m}$ and $f_{i,f}$ represent the fitness of the $i^{th}$ individual in male and female snakes, respectively.

2.2.3 Exploitation process

When the amount of food is sufficient, if the temperature $T$ > 0.6, the snake will only move in the direction of food. The mathematical model is defined by Eq. (8).

$$X_{i,j}(t+1) = X_{food} \pm c_3 \times \text{Temp} \times \text{rand} \times \left(X_{food} - X_{i,j}(t)\right) \tag{8}$$

Where $X_{i,j}$ represents the individual solution of male and female snakes, $X_{food}$ refers to the best agent, and $c_3$=2.

When $T$ < 0.6, the snake will be in battle mode or mating mode. The mathematical model of snake fighting mode is given in Eq. (9).

$$X_{i,m}(t+1) = X_{i,m}(t) + c_3 \times FM \times rand \times (Q \times X_{best,f} - X_{i,m}(t))$$
$$X_{i,f}(t+1) = X_{i,f}(t+1) + c_3 \times FF \times rand \times (Q \times X_{best,m} - X_{i,F}(t+1)) \quad (9)$$

Where $X_{i,m}$ and $X_{i,f}$ refer to the position of the $i^{th}$ male and female snake, respectively. $X_{best,f}$ and $X_{best,m}$ denote the best solution in the group of female and male snakes respectively, and $c_3$=2, $FM$ and $FF$ indicate the fighting ability of male and female snakes, respectively, which is described by Eq. (10).

$$\mathrm{FM} = exp(\frac{-f_{best,f}}{f_i})$$
$$\mathrm{FF} = exp(\frac{-f_{best,m}}{f_i}) \quad (10)$$

Where $f_{best,f}$ and $f_{best,m}$ represent the fitness of the best agent of the female snakes and the male snakes, respectively, and $f$ is the individual fitness.

The mathematical model of snake mating pattern is shown in Eq. (11).

$$X_{i,m}(t+1) = X_{i,m}(t) + c_3 \times M_m \times rand \times (Q \times X_{i,f}(t) - X_{i,m}(t))$$
$$X_{i,f}(t+1) = X_{i,f}(t) + c_3 \times M_f \times rand \times (Q \times X_{i,m}(t) - X_{i,f}(t)) \quad (11)$$

Where $X_{i,m}$ and $X_{i,f}$ refer to the location of the $i^{th}$ male and female snakes in the population, respectively, and $M_m$ and $M_f$ denote the mating ability of males and females, respectively, as follows Eq. (12).

$$M_m = \exp\left(-\frac{f_{i,f}}{f_{i,m}}\right)$$
$$M_f = \exp\left(-\frac{f_{i,m}}{f_{i,f}}\right) \quad (12)$$

After mating, if the snake eggs hatch, the new individual will replace the worst male and female in the population. The mathematical expression is described by Eq. (13).

$$X_{worst,m} = X_{min} + rand \times (X_{max} - X_{min})$$
$$X_{worst,f} = X_{min} + rand \times (X_{max} - X_{min}) \quad (13)$$

When $Q$ < 0.25, the SO explores in the search space according to Eq. (6). If $Q$ > 0.25, the Snake exploits in the promising area using Eq. (8), when $T$ > 0.6 and $rand$ > 0.6, the snake is in fighting mode through Eq. (9). Conversely, when $T$ < 0.6 and $rand$ < 0.6, the snake utilizes Eq. (11) to initiate the mating mode. Finally, snake egg hatching uses Eq. (13) to replace the worst individual in the population. In summary, the flowchart of so is shown in Fig. 2.

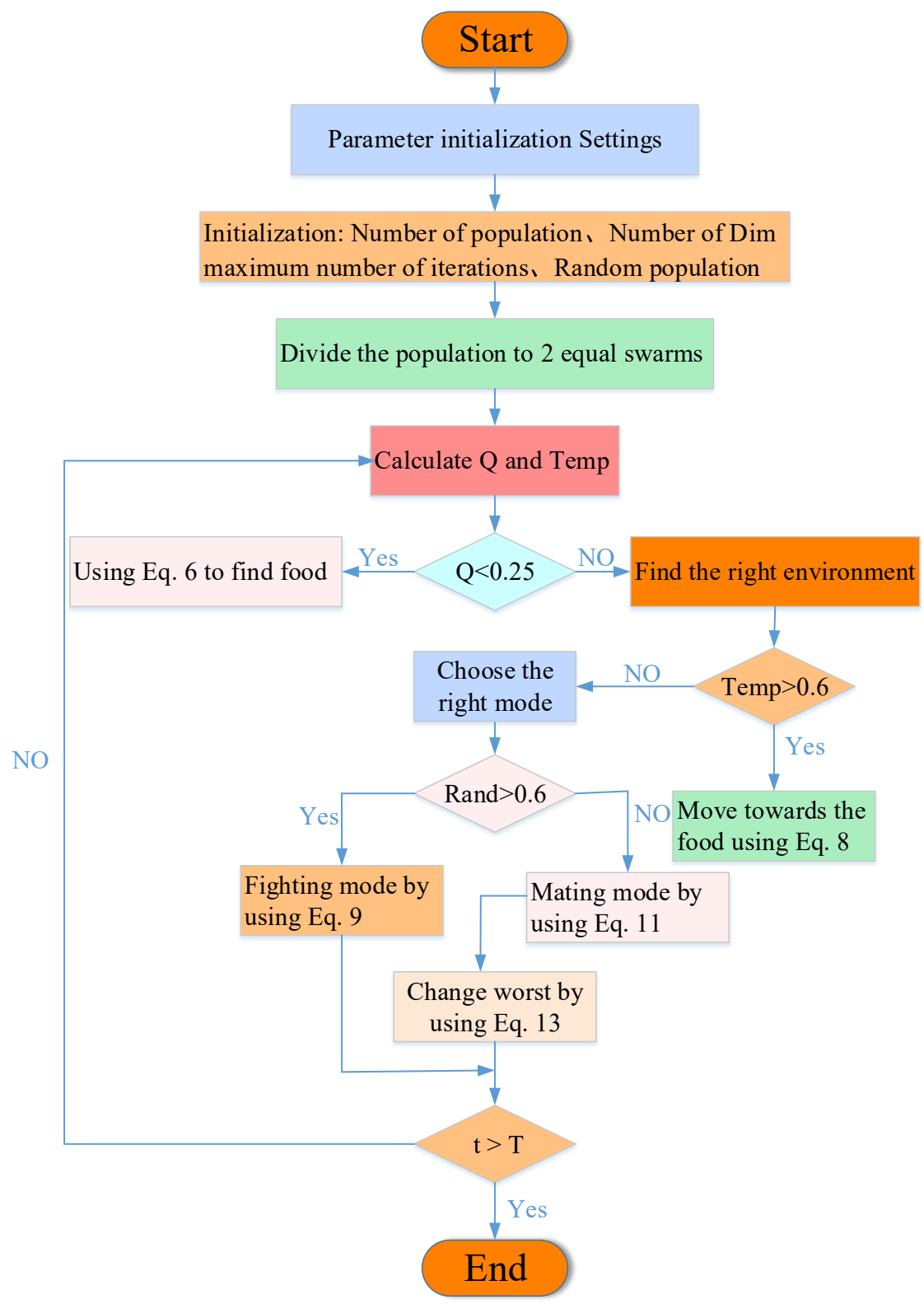

Fig. 2. The flowchart of SO.

### 3. The proposed MISO

This section describes our proposed MISO. Because SO has problems such as low accuracy and easy to stagnate in the local optimum in the optimization process. Therefore, we propose three improvement strategies to enhance the performance of SO, and the specific details of the improvements are elaborated in the subsequent chapters.

### 3.1 Adaptive random disturbance strategy based on sine function (DSO)

Due to the tendency of SO to become trapped in local minima during iterations, we propose a novel perturbation strategy. This strategy aims to assist the optimizer in escaping local optima more effectively throughout the search process. It is well-known that the sine function exhibits oscillatory behavior on the coordinate axis [82]. This observation serves as inspiration for introducing a novel Disturbance Factor ($DF$) that relies on the sine function. The mathematical model of this factor is provided in Eq. (14).

$$DF = \left( [\sin \times (2 \times rand) + 1] \times \left(1 - \frac{t}{T}\right)\right) \tag{14}$$

By exploiting the volatility of the $DF$, the optimizer can effectively locate the global optimum during the search process. Consequently, we have redefined the update rule of the SO when $t < \frac{T}{2}$. When $Q < 0.25$, the update formula of male and female snakes is shown in Eq. (15).

$$\begin{aligned} X_{i,m}(t+1) &= X_{rand,m}(t) \pm DF \times c_2 \times A_m \times \left((X_{max} - X_{min}) \times rand + X_{min}\right) \\ X_{i,f}(t+1) &= X_{rand,f}(t) \pm DF \times c_2 \times A_f \times \left((X_{max} - X_{min}) \times rand + X_{min}\right) \end{aligned} \tag{15}$$

When the food is sufficient, if the temperature $T$ > 0.6, the mathematical model is described by Eq. (16).

$$X_{i,j}(t+1) = X_{food} \pm DF \times c_3 \times \text{Temp} \times \text{rand} \times \left(X_{food} - X_{i,j}(t)\right) \tag{16}$$

When the $T$ < 0.6, the fighting model of snake is defined as Eq. (17).

$$\begin{aligned} X_{i,m}(t+1) &= X_{i,m}(t) + DF \times c_3 \times FM \times rand \times (Q \times X_{best,f} - X_{i,m}(t)) \\ X_{i,f}(t+1) &= X_{i,f}(t+1) + DF \times c_3 \times FF \times rand \times (Q \times X_{best,m} - X_{i,F}(t+1)) \end{aligned} \tag{17}$$

The mating pattern of snake is provided by Eq. (18).

$$\begin{aligned} X_{i,m}(t+1) &= X_{i,m}(t) + DF \times c_3 \times M_m \times rand \times (Q \times X_{i,f}(t) - X_{i,m}(t)) \\ X_{i,f}(t+1) &= X_{i,f}(t) + DF \times c_3 \times M_f \times rand \times (Q \times X_{i,m}(t) - X_{i,f}(t)) \end{aligned} \tag{18}$$

### 3.2 Adaptive Levy flight strategy based on scale factor and leader (LSO)

In the random search process, due to insufficient diversity of snake population. SO is easy to fall into the trap of local solutions in the later iteration, resulting in premature convergence of the algorithm. In order to address this issue, we propose adaptive Levy flight [83, 84] strategy based on scale factor and leader to improve the global search capability of the optimizer, mitigate the risk of SO stagnation in local optima, and improve the convergence accuracy of the algorithm. Levy flight is a random movement strategy characterized by small, continuous steps and occasional long jumps over short time intervals. This strategy can be likened to granting snakes the ability to fly, thereby enhancing their exploration of the search space. The long jumps facilitate the algorithm in exploring the global range within the search space, while the small step movements aid in improving the optimization accuracy.

Considering the characteristics of Levy flight, we introduced a novel update rule in the exploitation phase of the SO, equipping male snake leaders with flight ability to enhance the search efficiency. This update rule is presented in Eq. (19).

$$X_{i,m}(t+1) = X_{food} + CF \times \left(RL \times \left(RL \times X_{food} - X_{i,m}(t)\right)\right) \tag{19}$$

Where $X_{i,m}$ represents the individual position of the male snake, and $X_{food}$ refers to the best agent. $CF$ is the nonlinear Convergence Factor proposed by us, defined as Eq. (20). In order to improve the optimization accuracy of the optimizer, we utilize the weighted levy flight, expressed as $'RL'$, and the calculation method is denoted by Eq. (21).

$$CF = \left(cos\left(\frac{\pi}{2} \times \frac{t}{T}\right) \times \left(1 - \frac{t}{T}\right)^{\frac{2\times t}{T}}\right) \tag{20}$$

$$RL = 0.05 \times Levy(D) \tag{21}$$

Where $D$ denotes the dimensional space, and $Levy(D)$ is the Levy flight distribution function [85]. The calculation method defined by Eq. (22) [86].

$$\text{Levy}(D) = s \times \frac{u \times \sigma}{|v|^{\frac{1}{\eta}}} \tag{22}$$

Where $s$ and $\eta$ are the fixed constant 0.01 and 1.5, respectively. $u$ and $v$ are random numbers in the interval [0,1]. The calculation formula of $\sigma$ is as follows :

$$\sigma = \left(\frac{\Gamma(1+\eta) \times \sin\left(\frac{\pi\eta}{2}\right)}{\Gamma\left(\frac{1+\eta}{2}\right) \times \eta \times 2\left(\frac{\eta-1}{2}\right)}\right)^{\frac{1}{\eta}} \tag{23}$$

Where, $\Gamma$ indicates the gamma function. The value of $\eta$ is 1.5.

### 3.3 Adaptive position update strategy combining elite leadership and Brownian motion (BSO)

In order to reduce the risk of SO becoming trapped in local optimum, we propose a new position update strategy combining elite leadership and Brownian motion [87] to enhance the local search capability of the optimizer. Brownian motion is a continuous random motion model, characterized by particles undergoing continuous and irregular motion in a random manner over short time intervals. The randomness of the strategy enables the search agent to conduct irregular diffusion in the local area, enhancing the exploration of the local search space. Brownian motion can explore a wider search space and reduce the risk of getting stuck in a local optimum.

Based on the characteristics of Brownian motion, we propose a new update method that endows female snake leaders with the ability of Brownian motion in the exploitation stage of SO. The calculation rule is given in Eq. (24).

$$X_{i,f}(t+1) = X_{food} + CF \times \left(RB \times \left(RB \times X_{food} - X_{i,f}(t)\right)\right) \tag{24}$$

Where $X_{i,f}$ denotes the individual position of the female snake, and $X_{food}$ refers to the best agent. $CF$ is the nonlinear Convergence Factor, as shown in Eq. (20). In order to improve the exploitation accuracy of the optimizer, we utilize the weighted Brownian motion, expressed as $'RB'$, and its calculation method is given in Eq. (25).

$$RB = 0.05 \times Randn(D) \tag{25}$$

Where $D$ denotes the dimensional space, and $Randn(D)$ is a function of Brownian motion. The calculation method is defined by Eq. (26) [38].

$$Randn(D) = \frac{1}{\sqrt{2\pi\sigma^2}} \exp\left(-\frac{(x-\mu)^2}{2\sigma^2}\right) = \frac{1}{\sqrt{2\pi}} \exp\left(-\frac{x^2}{2}\right) \tag{26}$$

Where $\mu$=0 and $\sigma$=1.

To sum up, we show the flowchart of MISO in Fig. 3. Algorithm 1 provides the pseudo-code for MISO.

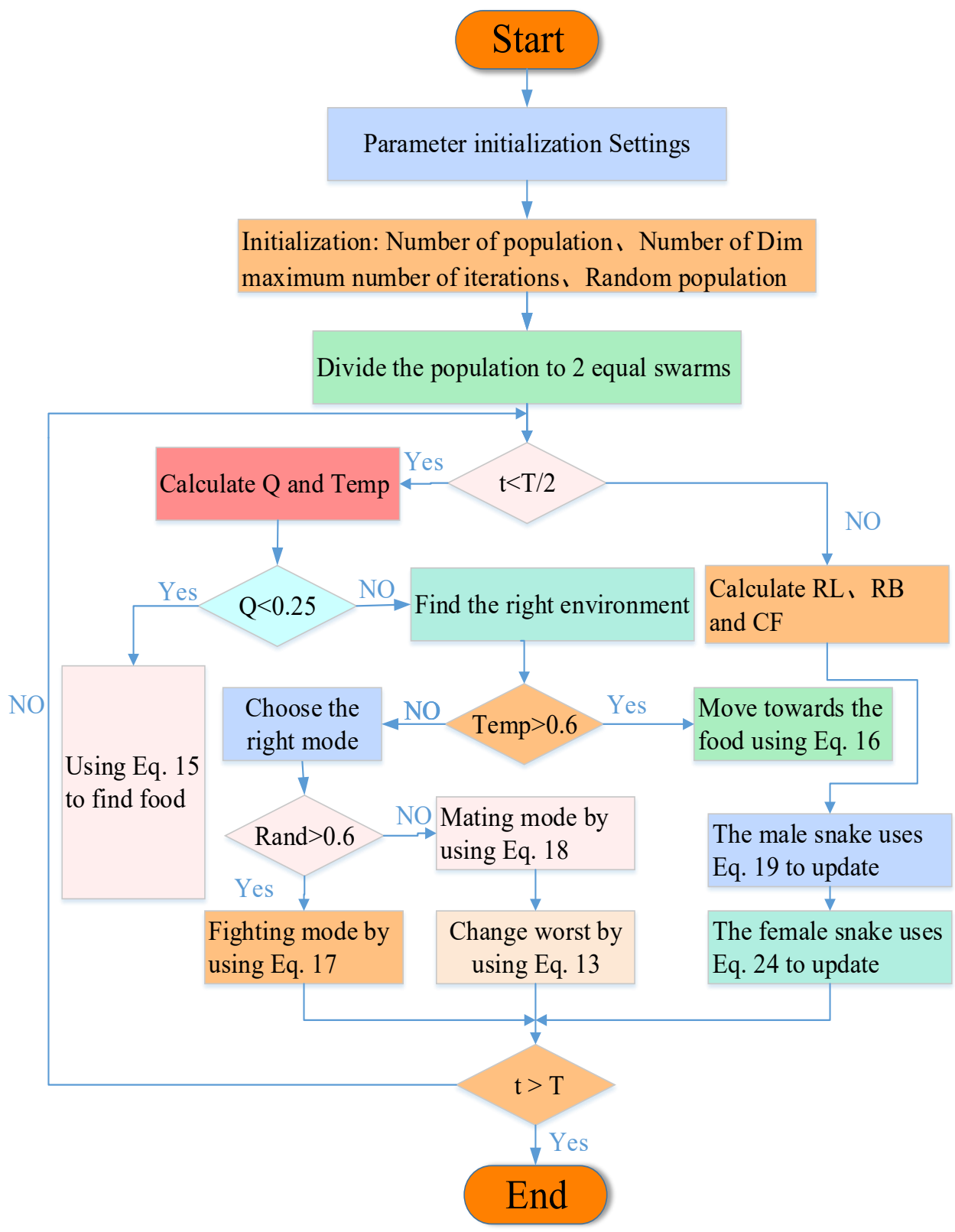

Fig. 3. The flowchart of MISO.

**Algorithm 1** Pseudo-Code of MISO Algorithm.

***Begin***

*Initialize the snake population N*

*Initialize related parameters of the snake*

*Divide population N to 2 equal groups Nm and Nf using Eq. (2) and Eq. (3).*

***While*** *t < T*

  ***if*** *t<$\frac{T}{2}$* ***do***

    *Calculate the fitness of the each search agent*

    *Simulate Temp using Eq. (4).*

    *Evaluate food Quantity Q using Eq. (5).*

    ***if*** *(Q < 0.25)*

      *Calculate **DF** using (14)*

      *Operate exploration using Eq. (15)*

    ***else if*** *(Q > 0.6)*

      *Perform exploitation using Eq. (16)*

    ***else***

      ***if*** *(rand > 0.6)*

        *Simulate Fight Mode using Eq. (17)*

      ***else***

        *Snakes in Mating Mode Eq. (18)*

        *Change the worst search agent using Eq. (13)*

      ***end if***

    ***end if***

  ***else***

    *Calculate **CF** using (20)*

    *Define **RL** using (21)*

    *Define **RB** using (25)*

    *Update the positions of the snake using Eq. (19) and Eq. (24)*

  ***end if***

***end while***

***End***

### 3.4 Time complexity analysis

The computational complexity of an algorithm is a crucial metric for evaluating its execution time. In this manuscript, we adopt the Big $O$ [88, 89] notation to analyze and compare the time complexity of the SO and MISO algorithms. The snake population size is $N$, $Nm = Nf = N/2$ ( $Nm$ is the number of male snakes, $Nf$ is the number of female snakes ), the dimension is $D$, and the maximum number of iterations is $T$. The time complexity of initializing population is $O(N \times D)$. The time complexity of finding the optimal individual position (food location) is $O(N)$. When there is a lack of food, the time complexity of finding food is $O(Nm \times D + Nf \times D)$; if the food is sufficient and the temperature is hot, the time complexity of finding a suitable environment is also $O(Nm \times D + Nf \times D)$. When the environment is suitable, the time complexity of snake competition and mating is $O(Nm \times D + Nf \times D)$. In summary, the total time complexity of the MISO iteration process is $O(T \times N \times D)$. Because no new cycles are introduced in MISO, the time complexity of MISO is the same as that of SO.

## 4. Experimental result analysis

In this section, we assess the capability of the MISO algorithm on different function test sets. To validate the effectiveness of MISO, we implemented experiments to estimate convergence, exploration and exploitation, population diversity and the impact of improvement methods, and compared with other 11 progressive algorithms. Subsequently, we employed two nonparametric tests, namely the Wilcoxon and Friedman tests, to analyze the distinctions among the competitors and their overall performance. The experiments were performed on a laptop equipped with an Intel Core i7-12700F, 2.10 GHz CPU, and 16 GB RAM, utilizing the MATLAB 2022b platform.

### 4.1 The benchmark test function

Benchmark test functions are crucial for assessing the algorithm's feasibility. These functions offer a platform to evaluate and test various optimization methods. We initially employ the CEC2017 test suite [90] to evaluate the performance of the proposed MISO method with dimensions set to 30, 50, and 100. In addition to the unimodal function, increasing dimensions will also lead to an increase in the number of local optimal solutions. This enables the test suite to effectively evaluate the algorithm's global optimization capability. Detailed information about the CEC2017 test suite is provided in Table 1. Finally, to validate MISO's ability to solve complex problems, we analyze its performance using the CEC2022 suite [91] with a dimension of 20. The specific function details are presented in Table 2.

Table 1
The CEC2017 test suite.

| Type | ID | CEC2017 Function name | Rang | Dimension | $f_{min}$ |
|---|---|---|---|---|---|
| Unimodal | F1 | Shifted and Rotated Bent Cigar Function | [-100,100] | 30/50/100 | 100 |
| | F2 | Shifted and Rotated Sum of Different Power Function | [-100,100] | 30/50/100 | 200 |
| | F3 | Shifted and Rotated Zakharov Function | [-100,100] | 30/50/100 | 300 |
| Multimodal | F4 | Shifted and Rotated Rosenbrock’s Function | [-100,100] | 30/50/100 | 400 |
| | F5 | Shifted and Rotated Rastrigin’s Function | [-100,100] | 30/50/100 | 500 |
| | F6 | Shifted and Rotated Expanded Scaffer’s F6 Function | [-100,100] | 30/50/100 | 600 |
| | F7 | Shifted and Rotated Lunacek Bi_Rastrigin Function | [-100,100] | 30/50/100 | 700 |
| | F8 | Shifted and Rotated Non-Continuous Rastrigin’s Function | [-100,100] | 30/50/100 | 800 |
| | F9 | Shifted and Rotated Levy Function | [-100,100] | 30/50/100 | 900 |
| | F10 | Shifted and Rotated Schwefel’s Function | [-100,100] | 30/50/100 | 1000 |
| Hybrid | F11 | Hybrid Function 1 (N=3) | [-100,100] | 30/50/100 | 1100 |
| | F12 | Hybrid Function 2 (N=3) | [-100,100] | 30/50/100 | 1200 |
| | F13 | Hybrid Function 3 (N=3) | [-100,100] | 30/50/100 | 1300 |
| | F14 | Hybrid Function 4 (N=4) | [-100,100] | 30/50/100 | 1400 |
| | F15 | Hybrid Function 5 (N=4) | [-100,100] | 30/50/100 | 1500 |
| | F16 | Hybrid Function 6 (N=4) | [-100,100] | 30/50/100 | 1600 |
| | F17 | Hybrid Function 6 (N=5) | [-100,100] | 30/50/100 | 1700 |
| | F18 | Hybrid Function 6 (N=5) | [-100,100] | 30/50/100 | 1800 |
| | F19 | Hybrid Function 6 (N=5) | [-100,100] | 30/50/100 | 1900 |
| | F20 | Hybrid Function 6 (N=6) | [-100,100] | 30/50/100 | 2000 |
| Composition | F21 | Composition Function 1 (N=3) | [-100,100] | 30/50/100 | 2100 |
| | F22 | Composition Function 2 (N=3) | [-100,100] | 30/50/100 | 2200 |
| | F23 | Composition Function 3 (N=4) | [-100,100] | 30/50/100 | 2300 |
| | F24 | Composition Function 4 (N=4) | [-100,100] | 30/50/100 | 2400 |
| | F25 | Composition Function 5 (N=5) | [-100,100] | 30/50/100 | 2500 |
| | F26 | Composition Function 6 (N=5) | [-100,100] | 30/50/100 | 2600 |
| | F27 | Composition Function 7 (N=6) | [-100,100] | 30/50/100 | 2700 |
| | F28 | Composition Function 8 (N=6) | [-100,100] | 30/50/100 | 2800 |
| | F29 | Composition Function 9 (N=3) | [-100,100] | 30/50/100 | 2900 |
| | F30 | Composition Function 10 (N=3) | [-100,100] | 30/50/100 | 3000 |

Table 2

The CEC2022 test suite.

| Type | ID | Description | Range | Dimension | $f_{min}$ |
|---|---|---|---|---|---|
| Unimodal | F1 | Shifted and full Rotated Zakharov Function | [-100,100] | 10/20 | 300 |
| Multimodal | F2 | Shifted and full Rotated Rosenbrock's Function | [-100,100] | 10/20 | 400 |
| | F3 | Shifted and full Rotated Rastrigin's Function | [-100,100] | 10/20 | 600 |
| | F4 | Shifted and full Rotated Non-Continuous Rastrigin's Function | [-100,100] | 10/20 | 800 |
| | F5 | Shifted and full Rotated Levy Function | [-100,100] | 10/20 | 900 |
| Hybrid | F6 | Hybrid Function 1 (N=3) | [-100,100] | 10/20 | 1800 |
| | F7 | Hybrid Function 2 (N=6) | [-100,100] | 10/20 | 2000 |
| | F8 | Hybrid Function 3 (N=5) | [-100,100] | 10/20 | 2200 |
| Composition | F9 | Composition Function 1 (N=5) | [-100,100] | 10/20 | 2300 |
| | F10 | Composition Function 2 (N=4) | [-100,100] | 10/20 | 2400 |
| | F11 | Composition Function 3 (N=5) | [-100,100] | 10/20 | 2600 |
| | F12 | Composition Function 4 (N=6) | [-100,100] | 10/20 | 2700 |

### 4.2 Competitor algorithm parameter setting

MISO is compared with 11 other well-known optimizers, namely GWO [19], SCA [82], WOA [4], HHO [33], MFO [24], AVOA [42], CSA [44], CPSOGSA [92], AO [85], DBO [55], and the original SO. The parameter settings for these competitors are presented in Table 3. The maximum number of iterations and population size are set to 500 and 30, respectively, and each algorithm is independently run 30 times. The Standard deviation (Std) and the Average (Ave) were recorded, and the best results are highlighted in bold.

Table 3

Parameter Settings for the selected competitive algorithm.

| Algorithms | Name of the parameter | Value of the parameter |
|---|---|---|
| GWO | a | [0,2] |
| SCA | a | 2 |
| WOA | a, a2, b | [0,2], [-1,-2], 1 |
| HHO | $E_0$, $E_1$ | [-1,1], [0,2] |
| MFO | $\varphi_1$, $\varphi_2$ | 2.05, 2.05 |
| AVOA | $L_1$, $L_2$, w, $p_1$, $p_2$, $p_3$ | 0.8, 0.2, 2.5, 0.6, 0.4, 0.6 |
| CSA | α, δ, r1, U=0, ω | 0.1, 0.1, 10, 0.00565, 0.005 |
| CPSOGSA | $\varphi_1$, $\varphi_2$ | 2.05, 2.05 |
| AO | alpha, delta | 0.1， 0.1 |
| DBO | P_percent | 0.2 |
| SO | Q, T, c1, c2, c3 | 0.25, 0.6, 0.5, 0.05, 2 |
| MISO | DF，CF | [0,1], [0,1] |

### 4.3 Qualitative analysis

In this section, we perform a qualitative analysis of the MISO through the design of three experiments: population diversity, exploration and exploitation, and convergence behavior.

4.3.1 Convergence behavior analysis

To investigate the convergence of the MISO, we arrange experiments to interpret its convergence characteristic. As plotted in Fig. 4, the first column represents the two-dimensional search space of the test function, vividly showcasing the complexity of the objective function. The second column denotes the search history of the agent. It is evident that the majority of individuals are concentrated near the optimal solution and distributed across the entire search area, demonstrating MISO's adeptness at avoiding local optima. The third column illustrates the change in the mean fitness value of the search agent. Initially, this value is large, signifying extensive exploration of the entire search space. Subsequently, it rapidly decreases, suggesting that most individuals have the potential to find the optimal value. The fourth column represents the search trajectory of the individual, transitioning from fluctuation to stability. This process reflects the search agent's shift from global exploration to local exploitation, facilitating the attainment of the global optimal solution. The final column presents the convergence curve of MISO. For unimodal functions, the curve continuously decreases, indicating the algorithm's ability to find the optimal value with sufficient iterations. For multimodal functions, the curve decreases step by step, indicating the algorithm's capability to consistently escape local optima and obtain the global optimum.

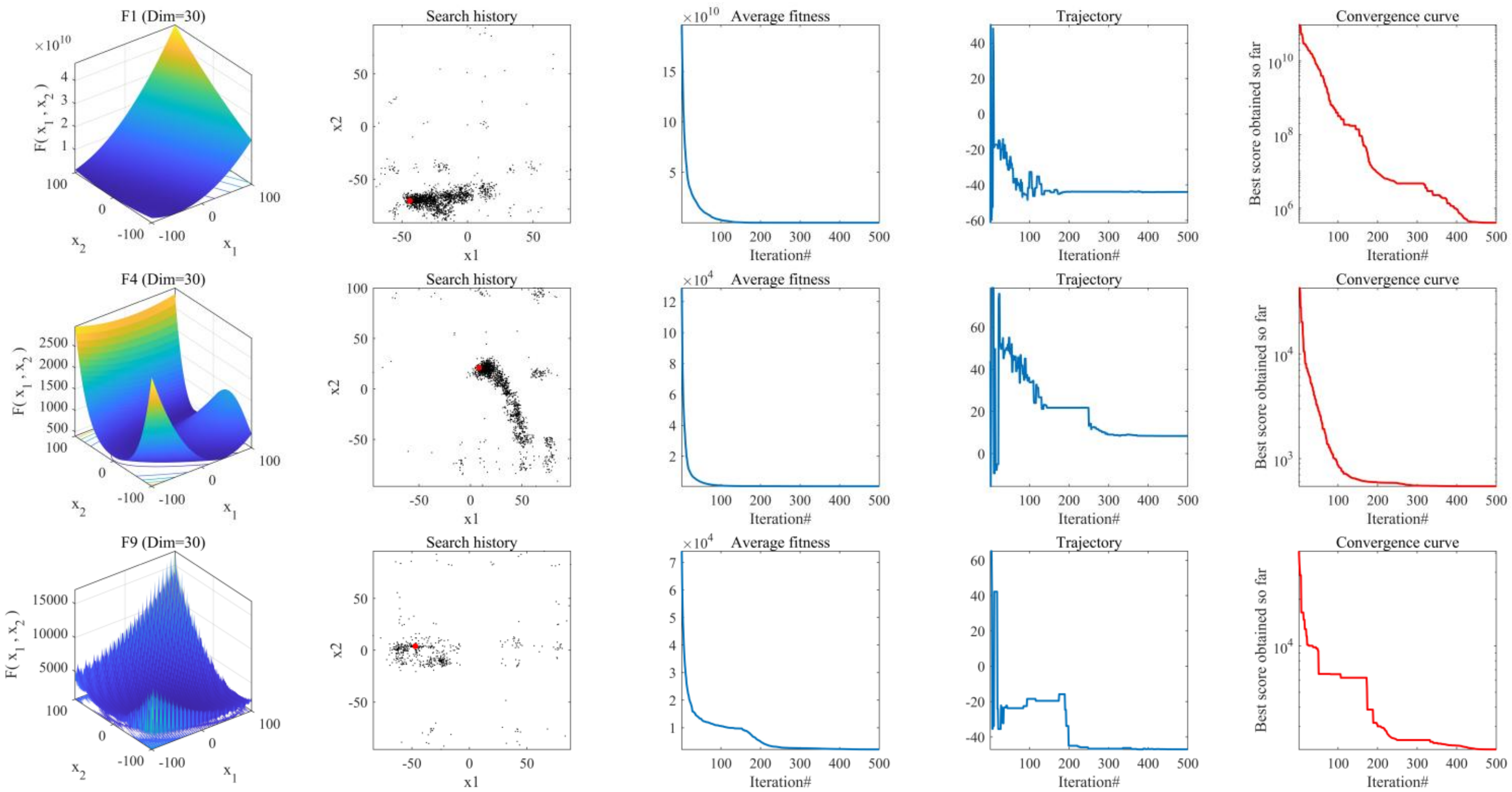

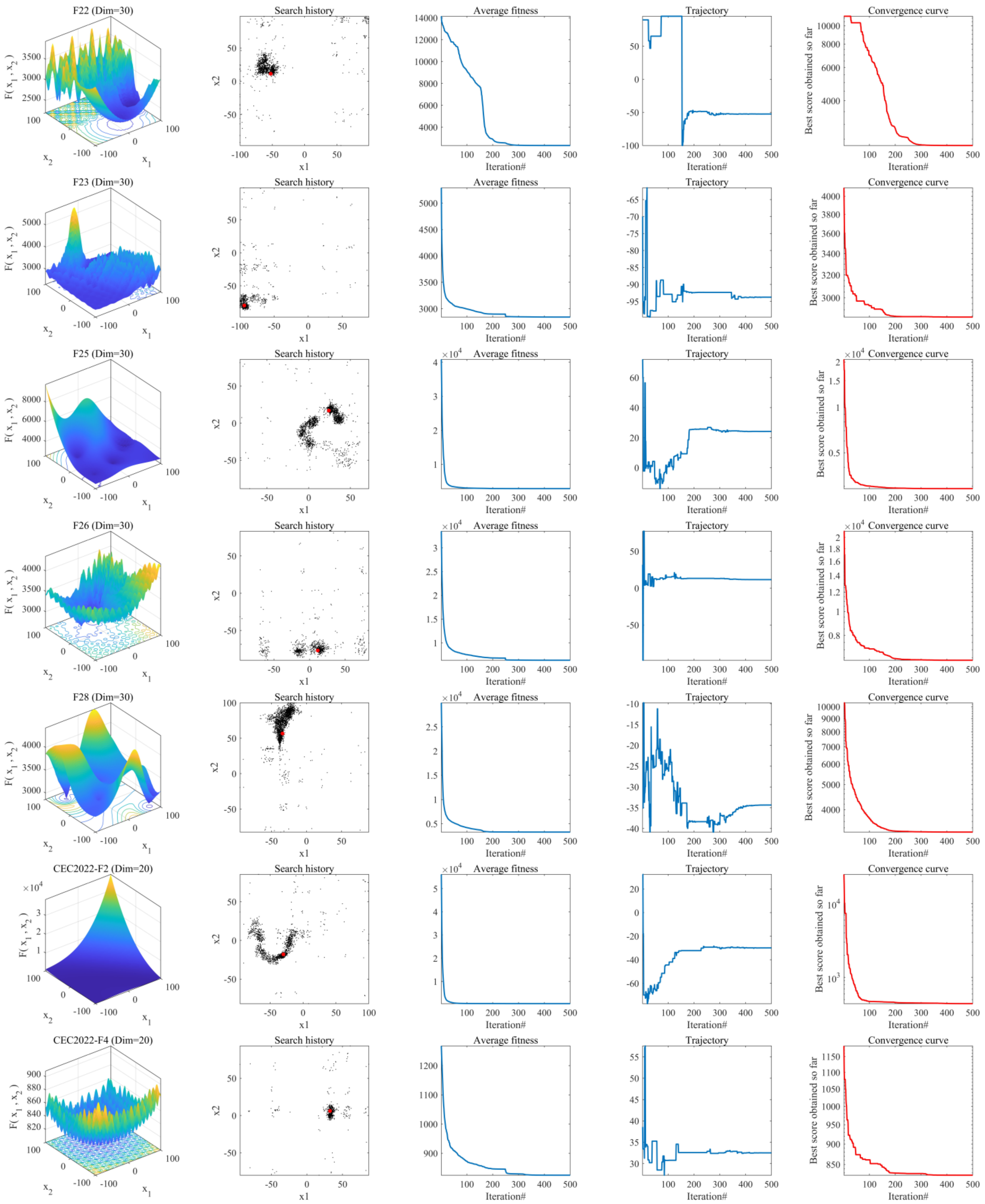

F22 (Dim=30)
Search history
Average fitness
Trajectory
Convergence curve
F23 (Dim=30)
F25 (Dim=30)
F26 (Dim=30)
F28 (Dim=30)
CEC2022-F2 (Dim=20)
CEC2022-F4 (Dim=20)
F( x1 , x2 )
x1
x2
Iteration#
Best score obtained so far

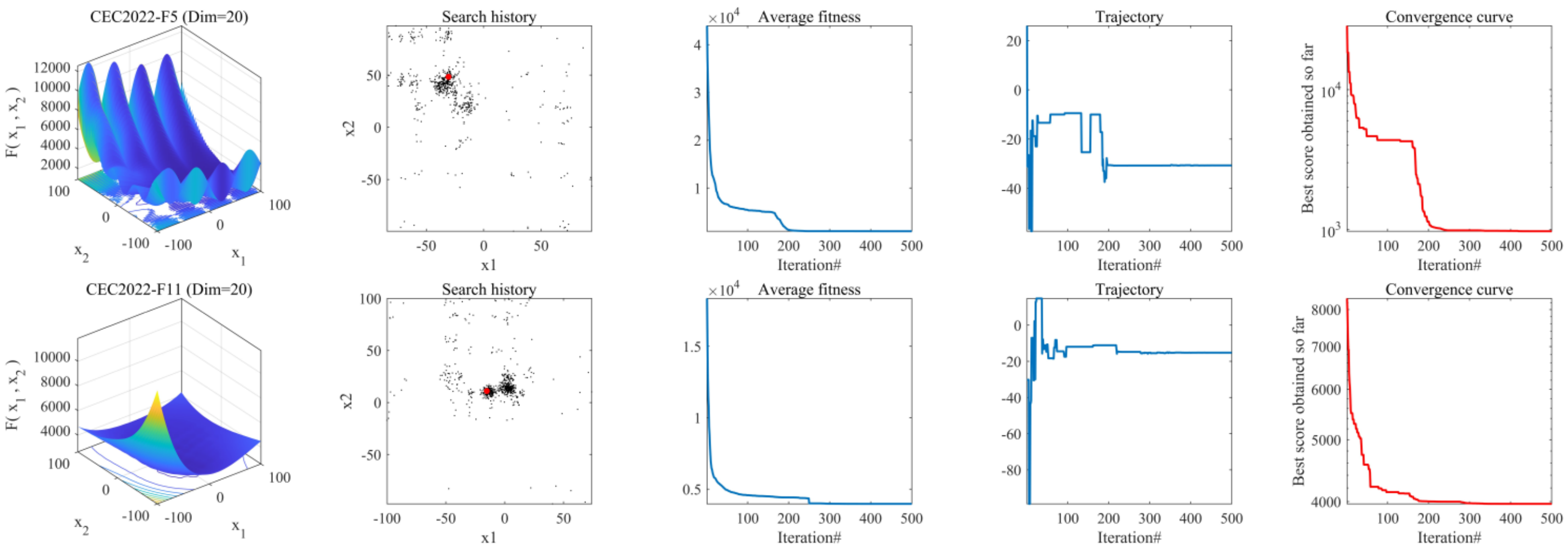

Fig. 4. Convergence behavior of MISO.

4.3.2 Exploration and exploitation analysis

Swarm intelligence algorithms possess an essential characteristic of dividing the search process into two stages: exploration and exploitation. Exploration entails searching for new solutions in an unknown space to find improved solutions in undiscovered regions. Exploitation involves discovering better fitness values within a known solution space by exploring local regions. Striking a balance between exploration and exploitation enhances the optimizer's effectiveness and prevents premature convergence. Accordingly, we calculate the percentage of exploration and exploitation using Eq. (27) and Eq. (28), respectively. $Div(t)$ represents the dimension of diversity measure, calculated by the Eq. (29) [93]. Where $x_{id}$ is the position of the $i^{\text{th}}$ agent in the $D^{\text{th}}$ dimension, and $Div_{max}$ denotes the maximum diversity during the whole iteration.

$$Exploration(\%) = \frac{Div(t)}{Div_{max}} \times 100 \tag{27}$$

$$Exploitation(\%) = \frac{|Div(t) - Div_{max}|}{Div_{max}} \times 100 \tag{28}$$

$$Div(t) = \frac{1}{D}\sum_{d=1}^{D}\frac{1}{N}\sum_{i=1}^{N}\left|median\left(x_d(t)\right) - x_{id}(t)\right| \tag{29}$$

Fig. 5 illustrates the experimental results of MISO on CEC2017 (30 dim) and CEC2022 (20 dim) suites. The horizontal axis represents the number of iterations, while the vertical axis denotes the percentage of exploration and exploitation during the iterative process. The intersection of MISO exploration and exploitation ratio mainly occurs in the middle of the iteration. The algorithm exhibits a substantial exploration proportion in the early stages of iteration, signifying superior exploration ability. Moreover, it maintains a high exploitation ratio in the later stages, which aids in improving the convergence speed and accuracy of the problem. The exploration and exploitation of the MISO algorithm in the iterative process are basically in a dynamic balance. Consequently, MISO has outstanding advantages in avoiding local optima and premature convergence.

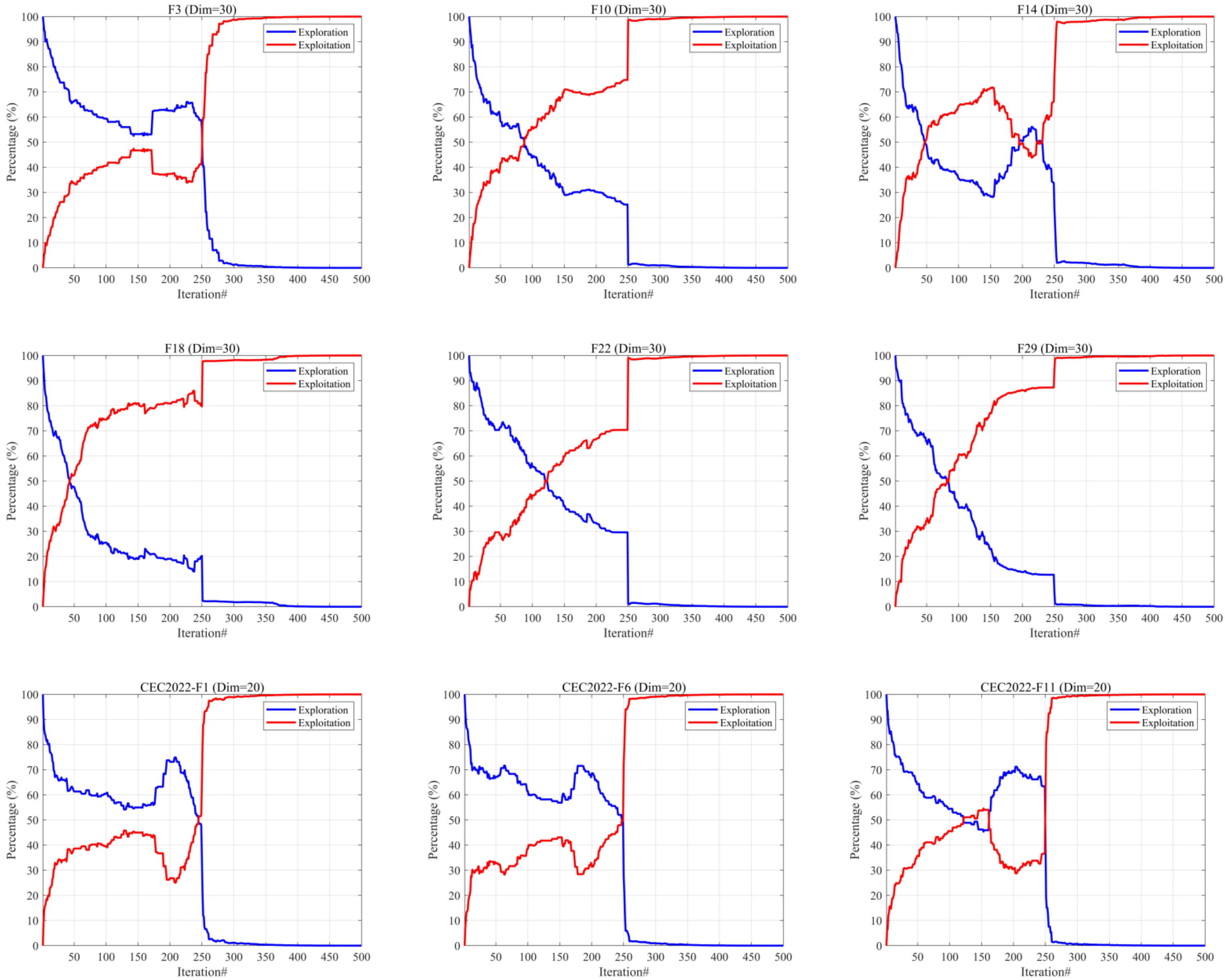

Fig. 5. Exploration and exploitation of MISO.

### 4.3.3 Population diversity analysis

Population diversity plays a pivotal role in swarm intelligence algorithms. High population diversity enables broader exploration in the search space and prevents premature convergence to local optima. To contrast the population size difference between SO and MISO, we executed population diversity experiments on the CEC2017 and CEC2022 test suites with dimensions set to 30 and 20, respectively. The population diversity is calculated using the moment of inertia $I_c$, described in Eq. (30), where $c_d$ represents the scatter of the population from its centroid $c$ in each update, and defined by Eq. (31), where $c_d$ denotes the value of the $D^{\text{th}}$ dimension of the $i^{\text{th}}$ agent at iteration $t^{\text{th}}$ [94].

$$I_c(t) = \sqrt{\sum_{i=1}^{N}\sum_{d=1}^{D}\left(x_{id}(t) - c_d(t)\right)^2} \tag{30}$$

$$c_d(t) = \frac{1}{D}\sum_{i=1}^{N} x_{id}(t) \tag{31}$$

The experimental results are presented in Fig. 6. Evidently, MISO exhibits higher population diversity than SO throughout the entire iteration process. This suggests that the enhanced optimizer can explore the search space more extensively, effectively avoiding premature convergence and local stagnation. Therefore, MISO has greater potential to achieve the global optimum.

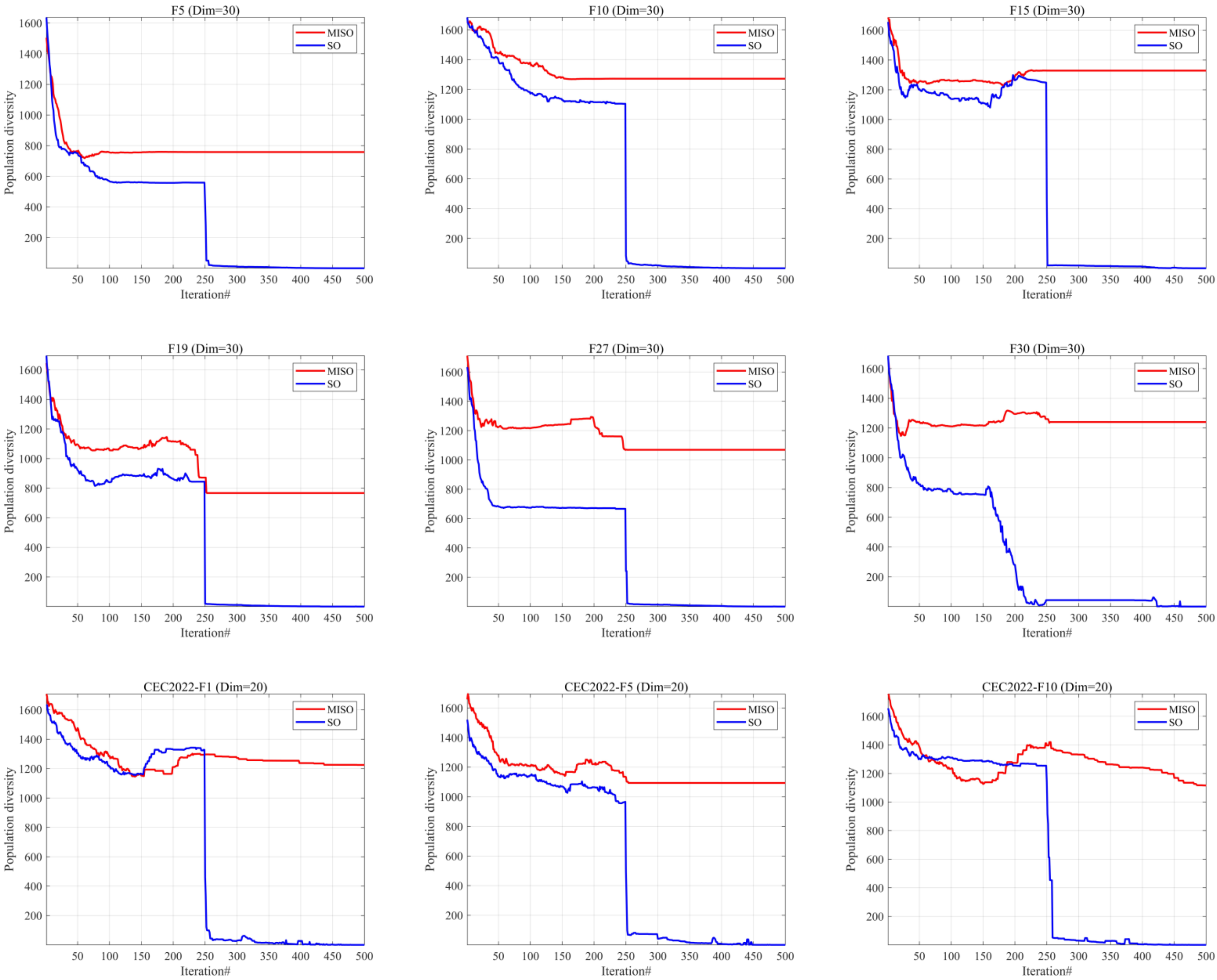

Fig. 6. Population diversity between SO and MISO.

### 4.4 Quantitative analysis

In this section, we utilize the CEC 2017 and CEC 2022 test suites to evaluate the effectiveness of MISO.

### 4.4.1 Comparison with other competitive algorithms in CEC 2017

To showcase the competitiveness of the proposed MISO, we implement performance examination using the CEC 2017 suite with dimensions set to 30, 50, and 100, respectively. The test results are reported in Table 4, Table 5, and Table 6. With increasing dimensions, the results of other optimizers are significantly affected, leading to possible convergence on local optima. In contrast, MISO remains less affected, demonstrating excellent stability and robustness. The last line of the table denotes the three labels (W|T|L), which represent the quantity of functions with the best (win), indistinguishable (tie) and worst (loss) results of the algorithm, respectively. The experimental results show that MISO has achieved the highest number of wins in all dimensions, and there is no worst number. Fig. 7 depicts the convergence curves of different competitors in different dimensions. As the dimension increases, the optimization difficulty also enhances. Apparently, other optimizers tend to stagnate on local optima. In contrast, most of MISO's curves exhibit a continuous downward trend, signifying MISO's superior potential to achieve the best solution. Fig. 8 shows the Friedman ranking of MISO and its comparison algorithms on the CEC2017 test set with 30, 50, and 100 dimensions, respectively, using the Sankey Diagram. It is worth noting that in all three dimensions, MISO obtains the highest number of first places. Fig. 9 shows the results of the 12 algorithms on the three dimensions of the CEC2017 test set in a more comprehensive way in the form of box plots, and it can be seen that MISO achieves excellent results. This proves that the algorithm has excellent global exploration and local exploitation performance, and verifies the effectiveness and accuracy of the proposed MISO algorithm.

Table 4

Experimental results of 12 algorithms on CEC 2017 (Dim=30).

| ID | | GWO | SCA | WOA | HHO | MFO | AVOA | CSA | CPSOGSA | AO | DBO | SO | MISO |
|---|---|---|---|---|---|---|---|---|---|---|---|---|---|
| F1 | Ave | 2.3946E+09 | 2.1217E+10 | 5.7257E+09 | 5.4303E+08 | 1.0021E+10 | 4.5279E+06 | 2.0239E+08 | 1.2565E+09 | 4.4163E+09 | 2.7839E+08 | 1.1847E+07 | **2.3682E+05** |
| | Std | 1.7937E+09 | 4.1600E+09 | 2.1124E+09 | 2.9039E+08 | 7.6720E+09 | 1.2932E+07 | 1.2462E+08 | 1.3352E+09 | 1.5885E+09 | 2.0169E+08 | 1.2784E+07 | **4.4552E+05** |
| F2 | Ave | 3.3484E+32 | 3.6210E+38 | 5.6043E+35 | 2.1907E+35 | 1.7271E+38 | 3.9514E+26 | 1.3604E+29 | 4.3052E+37 | 6.2736E+37 | 7.3353E+32 | 1.8395E+27 | **1.0979E+20** |
| | Std | 1.2285E+33 | 1.1632E+39 | 2.6687E+36 | 8.9277E+35 | 8.7643E+38 | 2.0104E+27 | 6.6743E+29 | 1.8084E+38 | 1.9406E+38 | 2.9949E+33 | 9.9378E+27 | **2.6272E+20** |
| F3 | Ave | 5.9488E+04 | 8.3489E+04 | 2.8247E+05 | 5.6536E+04 | 1.6423E+05 | **5.5321E+04** | 7.6713E+04 | 1.7296E+05 | 6.8343E+04 | 8.7848E+04 | 7.2616E+04 | 9.6692E+04 |
| | Std | 9.2556E+03 | 1.7091E+04 | 7.7278E+04 | 6.8757E+03 | 4.7821E+04 | 8.7718E+03 | 2.3432E+04 | 5.3072E+04 | **6.4895E+03** | 1.9480E+04 | 9.9689E+03 | 3.0920E+04 |
| F4 | Ave | 7.3738E+02 | 3.0136E+03 | 1.4388E+03 | 7.3956E+02 | 1.1412E+03 | 5.3510E+02 | 5.8855E+02 | 8.0596E+02 | 1.1209E+03 | 7.0823E+02 | 5.6463E+02 | **5.1139E+02** |
| | Std | 3.6381E+02 | 9.3079E+02 | 4.3607E+02 | 1.1202E+02 | 6.1568E+02 | **2.6253E+01** | 6.4862E+01 | 2.8663E+02 | 2.6560E+02 | 1.4182E+02 | 4.5646E+01 | 2.8592E+01 |
| F5 | Ave | 6.2085E+02 | 8.3171E+02 | 8.7207E+02 | 7.7576E+02 | 7.2228E+02 | 7.2868E+02 | 6.5451E+02 | 7.8422E+02 | 7.2858E+02 | 7.4398E+02 | 6.0666E+02 | **5.8775E+02** |
| | Std | 3.1693E+01 | 2.9168E+01 | 5.6527E+01 | 2.9246E+01 | 5.4273E+01 | 4.5288E+01 | 3.2735E+01 | 4.8639E+01 | 3.7025E+01 | 5.5313E+01 | 2.6388E+01 | **1.7919E+01** |
| F6 | Ave | 6.1273E+02 | 6.6465E+02 | 6.8048E+02 | 6.7113E+02 | 6.3814E+02 | 6.5426E+02 | 6.3673E+02 | 6.6862E+02 | 6.5641E+02 | 6.4980E+02 | 6.1958E+02 | **6.0957E+02** |
| | Std | **4.1188E+00** | 5.6757E+00 | 9.3432E+00 | 5.7447E+00 | 1.2548E+01 | 9.9374E+00 | 7.1202E+00 | 9.9177E+00 | 9.9712E+00 | 1.2426E+01 | 7.3011E+00 | 4.3834E+00 |
| F7 | Ave | 9.1309E+02 | 1.2463E+03 | 1.3187E+03 | 1.3216E+03 | 1.1623E+03 | 1.1666E+03 | 1.0041E+03 | 1.6096E+03 | 1.1649E+03 | 1.0015E+03 | 9.0433E+02 | **8.3264E+02** |
| | Std | 4.8928E+01 | 6.7611E+01 | 8.3721E+01 | 6.7599E+01 | 1.5282E+02 | 8.9296E+01 | 6.3034E+01 | 2.2345E+02 | 6.2149E+01 | 7.8690E+01 | 3.9700E+01 | **2.2403E+01** |
| F8 | Ave | 9.0088E+02 | 1.0985E+03 | 1.0768E+03 | 9.8957E+02 | 1.0360E+03 | 9.7201E+02 | 9.0796E+02 | 1.0246E+03 | 9.8216E+02 | 1.0218E+03 | 8.9761E+02 | **8.8846E+02** |
| | Std | 2.2467E+01 | 2.6322E+01 | 4.9938E+01 | 2.9040E+01 | 5.0272E+01 | 3.4750E+01 | 2.2638E+01 | 3.9986E+01 | 2.7000E+01 | 5.5683E+01 | 2.2527E+01 | **2.0502E+01** |
| F9 | Ave | 2.5196E+03 | 8.2736E+03 | 1.1569E+04 | 8.4281E+03 | 6.8688E+03 | 5.3327E+03 | 2.9241E+03 | 7.4280E+03 | 6.9023E+03 | 6.8615E+03 | 2.3237E+03 | **1.5968E+03** |
| | Std | 9.2495E+02 | 2.1983E+03 | 2.9931E+03 | 1.1693E+03 | 1.9599E+03 | 7.1345E+02 | 6.5397E+02 | 1.7208E+03 | 1.4409E+03 | 2.1066E+03 | 7.9919E+02 | **2.8261E+02** |
| F10 | Ave | 5.2766E+03 | 8.8867E+03 | 7.4874E+03 | 6.0354E+03 | 5.5221E+03 | 5.4315E+03 | 5.7908E+03 | 5.3919E+03 | 6.4701E+03 | 6.1174E+03 | 4.3826E+03 | **4.0139E+03** |
| | Std | 1.2576E+03 | **3.3930E+02** | 6.8817E+02 | 6.7954E+02 | 7.2287E+02 | 7.7783E+02 | 7.5659E+02 | 7.1019E+02 | 7.5154E+02 | 9.5577E+02 | 1.3132E+03 | 6.2111E+02 |
| F11 | Ave | 2.7539E+03 | 3.6552E+03 | 1.0305E+04 | 1.5163E+03 | 6.9273E+03 | 1.3492E+03 | 1.5489E+03 | 1.8048E+03 | 4.2574E+03 | 2.1830E+03 | 1.4655E+03 | **1.3322E+03** |
| | Std | 1.3963E+03 | 7.4135E+02 | 5.3988E+03 | 1.8912E+02 | 7.0406E+03 | 1.1841E+02 | 1.9091E+02 | 5.6452E+02 | 1.5780E+03 | 1.1944E+03 | 2.4649E+02 | **6.3937E+01** |
| F12 | Ave | 1.4211E+08 | 2.8269E+09 | 4.3518E+08 | 8.6556E+07 | 3.9216E+08 | 1.6624E+07 | 4.0546E+07 | 2.5940E+07 | 3.1600E+08 | 5.3554E+07 | 5.3677E+06 | **2.3211E+06** |
| | Std | 3.1584E+08 | 1.1496E+09 | 3.2406E+08 | 5.0332E+07 | 7.0933E+08 | 1.2741E+07 | 3.1939E+07 | 1.9134E+07 | 1.6909E+08 | 6.0064E+07 | 5.8948E+06 | **2.1415E+06** |
| F13 | Ave | 1.3075E+07 | 1.1203E+09 | 1.2213E+07 | 3.0469E+06 | 1.3502E+08 | 1.4893E+05 | 1.0130E+05 | 6.6064E+05 | 2.1917E+07 | 1.2052E+07 | **3.9629E+04** | 4.8582E+04 |
| | Std | 3.1777E+07 | 4.3869E+08 | 1.4770E+07 | 1.0643E+07 | 4.9208E+08 | 8.7809E+04 | 6.7818E+04 | 2.6377E+06 | 2.8565E+07 | 2.2419E+07 | **3.1537E+04** | 3.1912E+04 |
| F14 | Ave | 4.8756E+05 | 8.6282E+05 | 2.1220E+06 | 1.4090E+06 | 6.5812E+05 | 8.2671E+05 | **2.0730E+04** | 5.8453E+05 | 1.2922E+06 | 3.2934E+05 | 7.3660E+04 | 5.1399E+04 |
| | Std | 6.6658E+05 | 7.6917E+05 | 2.8177E+06 | 1.1697E+06 | 9.9573E+05 | 9.8181E+05 | **3.2712E+04** | 7.3786E+05 | 1.0650E+06 | 4.1415E+05 | 6.7077E+04 | 6.7966E+04 |
| F15 | Ave | 1.4272E+06 | 6.0875E+07 | 1.0382E+07 | 1.2125E+05 | 5.1146E+04 | 5.2926E+04 | 3.7884E+04 | 3.0017E+04 | 2.0758E+05 | 5.3700E+06 | **1.6785E+04** | 1.9974E+04 |
| | Std | 2.6736E+06 | 4.3285E+07 | 1.4804E+07 | 7.1484E+04 | 3.7973E+04 | 3.9183E+04 | 2.7485E+04 | 2.3231E+04 | 1.1694E+05 | 2.7963E+07 | **1.2262E+04** | 1.7158E+04 |
| F16 | Ave | 2.6888E+03 | 4.1296E+03 | 4.3403E+03 | 3.7533E+03 | 3.0846E+03 | 3.2251E+03 | 2.8389E+03 | 3.3398E+03 | 3.5002E+03 | 3.3993E+03 | 2.5768E+03 | **2.4375E+03** |
| | Std | 4.9521E+02 | 2.8877E+02 | 6.6986E+02 | 5.0563E+02 | 3.8005E+02 | 3.4617E+02 | 3.5102E+02 | 4.2894E+02 | 4.4128E+02 | 4.1284E+02 | **2.2669E+02** | 2.8557E+02 |
| F17 | Ave | 2.1148E+03 | 2.8572E+03 | 2.7005E+03 | 2.6457E+03 | 2.4435E+03 | 2.4793E+03 | **2.0930E+03** | 2.8581E+03 | 2.5881E+03 | 2.6770E+03 | 2.2237E+03 | 2.1484E+03 |
| | Std | 1.7803E+02 | 2.0838E+02 | 3.3196E+02 | 2.7215E+02 | 3.3068E+02 | 2.5213E+02 | 1.9512E+02 | 2.8600E+02 | 2.6431E+02 | 2.8306E+02 | 1.9839E+02 | **1.7718E+02** |
| F18 | Ave | 2.4117E+06 | 1.5862E+07 | 1.5718E+07 | 5.4743E+06 | 3.4784E+06 | 1.7389E+06 | **3.9828E+05** | 7.9018E+05 | 7.0564E+06 | 2.3075E+06 | 1.3903E+06 | 6.0262E+05 |
| | Std | 2.2018E+06 | 1.0487E+07 | 1.4403E+07 | 7.0648E+06 | 5.0050E+06 | 1.7906E+06 | 5.8223E+05 | 9.2124E+05 | 5.6397E+06 | 3.5629E+06 | 1.6726E+06 | **5.5229E+05** |
| F19 | Ave | 1.4670E+06 | 1.3368E+08 | 2.3504E+07 | 1.7186E+06 | 5.7401E+06 | 1.3057E+05 | 1.2674E+06 | 2.5703E+04 | 3.4776E+06 | 4.2094E+06 | **1.1042E+04** | 1.9745E+04 |
| | Std | 2.6807E+06 | 1.0489E+08 | 2.1100E+07 | 1.6548E+06 | 1.8748E+07 | 1.4944E+05 | 1.2104E+06 | 2.5615E+04 | 3.3981E+06 | 9.4499E+06 | **1.0809E+04** | 2.6662E+04 |
| F20 | Ave | **2.4654E+03** | 2.9485E+03 | 2.8875E+03 | 2.8492E+03 | 2.7276E+03 | 2.8245E+03 | 2.4960E+03 | 2.8905E+03 | 2.6911E+03 | 2.7468E+03 | 2.4890E+03 | 2.4798E+03 |
| | Std | 1.8571E+02 | 1.9991E+02 | 2.1514E+02 | 2.0497E+02 | 2.0776E+02 | 2.2409E+02 | 1.8314E+02 | 2.6472E+02 | 1.7882E+02 | 2.2501E+02 | 1.3545E+02 | **1.2596E+02** |
| F21 | Ave | 2.4038E+03 | 2.5964E+03 | 2.6416E+03 | 2.5807E+03 | 2.5087E+03 | 2.5497E+03 | 2.4228E+03 | 2.5940E+03 | 2.5161E+03 | 2.5555E+03 | 2.4053E+03 | **2.3897E+03** |
| | Std | 3.8583E+01 | 2.3696E+01 | 7.0562E+01 | 4.3479E+01 | 4.1002E+01 | 5.7302E+01 | 3.1665E+01 | 7.0574E+01 | 4.7758E+01 | 5.1019E+01 | **2.0743E+01** | 2.2796E+01 |
| F22 | Ave | 5.6000E+03 | 1.0275E+04 | 8.4160E+03 | 7.2253E+03 | 6.5539E+03 | 5.1503E+03 | 4.9175E+03 | 6.4982E+03 | **3.3994E+03** | 5.1048E+03 | 3.9555E+03 | 3.6629E+03 |
| | Std | 2.0472E+03 | **3.8315E+02** | 1.4508E+03 | 1.5861E+03 | 1.2901E+03 | 2.2313E+03 | 2.4351E+03 | 1.5529E+03 | 5.0945E+02 | 2.1921E+03 | 1.6079E+03 | 1.4462E+03 |
| F23 | Ave | **2.7816E+03** | 3.0714E+03 | 3.1365E+03 | 3.2916E+03 | 2.8309E+03 | 2.9908E+03 | 2.8298E+03 | 3.0889E+03 | 3.0304E+03 | 3.0065E+03 | 2.8036E+03 | 2.7880E+03 |
| | Std | 3.2953E+01 | **2.7538E+01** | 1.1900E+02 | 1.4625E+02 | 4.4741E+01 | 6.8102E+01 | 3.9511E+01 | 8.3661E+01 | 9.6873E+01 | 9.5631E+01 | 3.4350E+01 | 3.6185E+01 |
| F24 | Ave | 2.9643E+03 | 3.2471E+03 | 3.2422E+03 | 3.4791E+03 | 2.9921E+03 | 3.1596E+03 | 2.9608E+03 | 3.3061E+03 | 3.1518E+03 | 3.1606E+03 | 2.9517E+03 | **2.9494E+03** |
| | Std | 5.5701E+01 | 4.6227E+01 | 8.0425E+01 | 1.5958E+02 | **2.7338E+01** | 8.7263E+01 | 3.2743E+01 | 9.5599E+01 | 5.1593E+01 | 8.0117E+01 | 3.8163E+01 | 3.3827E+01 |
| F25 | Ave | 3.0208E+03 | 3.6714E+03 | 3.2205E+03 | 3.0164E+03 | 3.3123E+03 | 2.9469E+03 | 3.0167E+03 | 3.0259E+03 | 3.0993E+03 | 3.0702E+03 | 2.9370E+03 | **2.9087E+03** |
| | Std | 5.2490E+01 | 2.5907E+02 | 1.0710E+02 | 2.9430E+01 | 3.5217E+02 | 3.5094E+01 | 3.0187E+01 | 7.3393E+01 | 7.5098E+01 | 3.3917E+02 | 3.9656E+01 | **2.2704E+01** |
| F26 | Ave | **5.0588E+03** | 7.9045E+03 | 8.7849E+03 | 7.7998E+03 | 5.8921E+03 | 7.0150E+03 | 5.5155E+03 | 7.3735E+03 | 6.5270E+03 | 6.7746E+03 | 5.5256E+03 | 5.3374E+03 |
| | Std | 6.0558E+02 | 3.9736E+02 | 8.5986E+02 | 1.5798E+03 | 4.4319E+02 | 1.1437E+03 | 6.0180E+02 | 1.0329E+03 | 1.2038E+03 | 1.1156E+03 | **3.5165E+02** | 3.7633E+02 |
| F27 | Ave | 3.2705E+03 | 3.5432E+03 | 3.4774E+03 | 3.6188E+03 | **3.2474E+03** | 3.2949E+03 | 3.3298E+03 | 3.6443E+03 | 3.4492E+03 | 3.3307E+03 | 3.2980E+03 | 3.2736E+03 |
| | Std | 2.9768E+01 | 6.5080E+01 | 1.2529E+02 | 2.1081E+02 | 3.0390E+01 | 4.8996E+01 | 5.1814E+01 | 2.4776E+02 | 7.6960E+01 | 7.1019E+01 | 3.3879E+01 | **1.6357E+01** |
| F28 | Ave | 3.5612E+03 | 4.4333E+03 | 3.9953E+03 | 3.5030E+03 | 4.4304E+03 | 3.3162E+03 | 3.3944E+03 | 3.3516E+03 | 3.8405E+03 | 3.6552E+03 | 3.3635E+03 | **3.2952E+03** |
| | Std | 2.5851E+02 | 3.6560E+02 | 7.6879E+02 | 8.4709E+01 | 9.5833E+02 | **3.6581E+01** | 5.9573E+01 | 4.9033E+01 | 2.3406E+02 | 6.9210E+02 | 5.6576E+01 | 4.5545E+01 |
| F29 | Ave | 3.9735E+03 | 5.2094E+03 | 5.4046E+03 | 5.1221E+03 | 4.1937E+03 | 4.3967E+03 | 4.3817E+03 | 4.5706E+03 | 4.8403E+03 | 4.4849E+03 | 4.0925E+03 | **3.9647E+03** |
| | Std | **1.6739E+02** | 3.6876E+02 | 5.2493E+02 | 6.3559E+02 | 3.6295E+02 | 2.6977E+02 | 4.1042E+02 | 3.2837E+02 | 3.6551E+02 | 4.0498E+02 | 2.1733E+02 | 2.6790E+02 |
| F30 | Ave | 1.1195E+07 | 1.8918E+08 | 7.5216E+07 | 8.6254E+06 | 8.4937E+05 | 1.4797E+06 | 5.8505E+06 | 8.9697E+05 | 3.3755E+07 | 4.3829E+06 | 2.0597E+05 | **1.7091E+05** |
| | Std | 9.6382E+06 | 6.4441E+07 | 6.5274E+07 | 6.9356E+06 | 1.2435E+06 | 1.0952E+06 | 3.0122E+06 | 8.0814E+05 | 2.7329E+07 | 4.8227E+06 | 2.5749E+05 | **1.6827E+05** |
| (W\|T\|L) | | (4\|26\|0) | (0\|15\|15) | (0\|21\|9) | (1\|25\|4) | (1\|29\|0) | (0\|30\|0) | (3\|27\|0) | (0\|28\|2) | (0\|30\|0) | (0\|30\|0) | (3\|27\|0) | (**18**\|12\|0) |

Table 5

Experimental results of 12 algorithms on CEC 2017 (Dim=50).

| ID | | GWO | SCA | WOA | HHO | MFO | AVOA | CSA | CPSOGSA | AO | DBO | SO | MISO |
|---|---|---|---|---|---|---|---|---|---|---|---|---|---|
| F1 | Ave | 2.3946E+09 | 2.1217E+10 | 5.7257E+09 | 5.4303E+08 | 1.0021E+10 | 4.5279E+06 | 2.0239E+08 | 1.2565E+09 | 4.4163E+09 | 2.7839E+08 | 1.1847E+07 | **2.3682E+05** |
| | Std | 4.6288E+09 | 8.4804E+09 | 4.3689E+09 | 2.1195E+09 | 2.1409E+10 | 4.3653E+08 | 1.2912E+09 | 3.5382E+09 | 3.2333E+09 | 1.4528E+10 | 2.7273E+08 | **1.9440E+07** |
| F2 | Ave | 3.3484E+32 | 3.6210E+38 | 5.6043E+35 | 2.1907E+35 | 1.7271E+38 | 3.9514E+26 | 1.3604E+29 | 4.3052E+37 | 6.2736E+37 | 7.3353E+32 | 1.8395E+27 | **1.0979E+20** |
| | Std | 1.3608E+63 | 2.5904E+72 | 7.0135E+80 | 7.3974E+66 | 3.1629E+77 | 2.0869E+52 | 8.1369E+59 | 5.1245E+72 | 1.0665E+68 | 9.0951E+64 | 2.9649E+54 | **1.3286E+50** |
| F3 | Ave | 5.9488E+04 | 8.3489E+04 | 2.8247E+05 | 5.6536E+04 | 1.6423E+05 | **5.5321E+04** | 7.6713E+04 | 1.7296E+05 | 6.8343E+04 | 8.7848E+04 | 7.2616E+04 | 9.6692E+04 |
| | Std | 3.5739E+04 | 4.1686E+04 | 8.0322E+04 | 1.6396E+04 | 9.4155E+04 | 3.2542E+04 | 3.9990E+04 | 6.5250E+04 | 1.0734E+05 | 7.9007E+04 | **1.5935E+04** | 4.8312E+04 |
| F4 | Ave | 7.3738E+02 | 3.0136E+03 | 1.4388E+03 | 7.3956E+02 | 1.1412E+03 | 5.3510E+02 | 5.8855E+02 | 8.0596E+02 | 1.1209E+03 | 7.0823E+02 | 5.6463E+02 | **5.1139E+02** |
| | Std | 8.2580E+02 | 3.4194E+03 | 1.7160E+03 | 5.9032E+02 | 3.7363E+03 | 1.0493E+02 | 2.4834E+02 | 1.1391E+03 | 9.1958E+02 | 5.4384E+02 | 1.1150E+02 | **4.1795E+01** |
| F5 | Ave | 7.5121E+02 | 1.1469E+03 | 1.1344E+03 | 9.2969E+02 | 9.9085E+02 | 8.5738E+02 | 8.2348E+02 | 1.0513E+03 | 9.4358E+02 | 1.0041E+03 | 7.2272E+02 | **6.8283E+02** |
| | Std | 5.5777E+01 | 3.6598E+01 | 8.7867E+01 | 3.4023E+01 | 1.0918E+02 | 3.9165E+01 | 3.9360E+01 | 6.8568E+01 | 4.5394E+01 | 1.0054E+02 | 3.4488E+01 | **2.7886E+01** |
| F6 | Ave | 6.1273E+02 | 6.6465E+02 | 6.8048E+02 | 6.7113E+02 | 6.3814E+02 | 6.5426E+02 | 6.3673E+02 | 6.6862E+02 | 6.5641E+02 | 6.4980E+02 | 6.1958E+02 | **6.0957E+02** |
| | Std | 5.8817E+00 | 4.8796E+00 | 1.2466E+01 | 5.2281E+00 | 8.6947E+00 | 5.8843E+00 | 6.7046E+00 | 9.2776E+00 | 6.4277E+00 | 1.2546E+01 | 6.2282E+00 | **4.5033E+00** |
| F7 | Ave | 9.1309E+02 | 1.2463E+03 | 1.3187E+03 | 1.3216E+03 | 1.1623E+03 | 1.1666E+03 | 1.0041E+03 | 1.6096E+03 | 1.1649E+03 | 1.0015E+03 | 9.0433E+02 | **8.3264E+02** |
| | Std | 7.1746E+01 | 1.0174E+02 | 1.1634E+02 | 7.8016E+01 | 3.4829E+02 | 1.0784E+02 | 1.2377E+02 | 3.3506E+02 | 1.2850E+02 | 1.4386E+02 | 7.7069E+01 | **4.6235E+01** |
| F8 | Ave | 9.0088E+02 | 1.0985E+03 | 1.0768E+03 | 9.8957E+02 | 1.0360E+03 | 9.7201E+02 | 9.0796E+02 | 1.0246E+03 | 9.8216E+02 | 1.0218E+03 | 8.9761E+02 | **8.8846E+02** |
| | Std | 6.2178E+01 | 4.8975E+01 | 8.2654E+01 | 4.4244E+01 | 1.0232E+02 | 5.6882E+01 | 3.9512E+01 | 7.6928E+01 | 5.3045E+01 | 1.0709E+02 | 3.5276E+01 | **3.4237E+01** |
| F9 | Ave | 2.5196E+03 | 8.2736E+03 | 1.1569E+04 | 8.4281E+03 | 6.8688E+03 | 5.3327E+03 | 2.9241E+03 | 7.4280E+03 | 6.9023E+03 | 6.8615E+03 | 2.3237E+03 | **1.5968E+03** |
| | Std | 5.4791E+03 | 6.7560E+03 | 1.1773E+04 | 4.3656E+03 | 5.3954E+03 | 1.8690E+03 | 2.9266E+03 | 3.8154E+03 | 4.2263E+03 | 8.5041E+03 | 2.3378E+03 | **7.6145E+02** |
| F10 | Ave | 5.2766E+03 | 8.8867E+03 | 7.4874E+03 | 6.0354E+03 | 5.5221E+03 | 5.4315E+03 | 5.7908E+03 | 5.3919E+03 | 6.4701E+03 | 6.1174E+03 | 4.3826E+03 | **4.0139E+03** |
| | Std | 2.2043E+03 | **3.8548E+02** | 9.0852E+02 | 1.0854E+03 | 1.1452E+03 | 1.1149E+03 | 1.2867E+03 | 9.0735E+02 | 1.0287E+03 | 2.1489E+03 | 2.7767E+03 | 9.4873E+02 |
| F11 | Ave | 2.7539E+03 | 3.6552E+03 | 1.0305E+04 | 1.5163E+03 | 6.9273E+03 | 1.3492E+03 | 1.5489E+03 | 1.8048E+03 | 4.2574E+03 | 2.1830E+03 | 1.4655E+03 | **1.3322E+03** |
| | Std | 2.3757E+03 | 3.3561E+03 | 2.3750E+03 | 8.3815E+02 | 1.1324E+04 | 4.6274E+02 | 2.5448E+03 | 5.3215E+03 | 1.5208E+03 | 3.7694E+03 | 2.4376E+03 | **2.4124E+02** |
| F12 | Ave | 1.4211E+08 | 2.8269E+09 | 4.3518E+08 | 8.6556E+07 | 3.9216E+08 | 1.6624E+07 | 4.0546E+07 | 2.5940E+07 | 3.1600E+08 | 5.3554E+07 | 5.3677E+06 | **2.3211E+06** |
| | Std | 1.9600E+09 | 4.0134E+09 | 1.9085E+09 | 6.4927E+08 | 5.6935E+09 | 5.1559E+07 | 2.7743E+08 | 1.0568E+09 | 2.6893E+09 | 7.6436E+08 | 5.5943E+07 | **1.6504E+07** |
| F13 | Ave | 1.3075E+07 | 1.1203E+09 | 1.2213E+07 | 3.0469E+06 | 1.3502E+08 | 1.4893E+05 | 1.0130E+05 | 6.6064E+05 | 2.1917E+07 | 1.2052E+07 | **3.9629E+04** | 4.8582E+04 |
| | Std | 7.1506E+08 | 2.9804E+09 | 2.8830E+08 | 1.9078E+07 | 1.7131E+09 | 1.2151E+05 | 3.4420E+05 | 2.5208E+08 | 7.6163E+08 | 1.0071E+08 | 2.3434E+05 | **6.7725E+04** |
| F14 | Ave | 4.8756E+05 | 8.6282E+05 | 2.1220E+06 | 1.4090E+06 | 6.5812E+05 | 8.2671E+05 | **2.0730E+04** | 5.8453E+05 | 1.2922E+06 | 3.2934E+05 | 7.3660E+04 | 5.1399E+04 |
| | Std | 1.5853E+06 | 5.5861E+06 | 8.7213E+06 | 7.8687E+06 | 9.3837E+06 | 2.0961E+06 | 4.5579E+05 | 1.8233E+06 | 6.7386E+06 | 5.7859E+06 | 8.0672E+05 | **2.8814E+05** |
| F15 | Ave | 1.4272E+06 | 6.0875E+07 | 1.0382E+07 | 1.2125E+05 | 5.1146E+04 | 5.2926E+04 | 3.7884E+04 | 3.0017E+04 | 2.0758E+05 | 5.3700E+06 | **1.6785E+04** | 1.9974E+04 |
| | Std | 2.3043E+08 | 3.4885E+08 | 9.4802E+07 | 3.6555E+07 | 3.5332E+08 | 6.6381E+04 | 4.8776E+04 | 7.6362E+07 | 1.6196E+07 | 1.1154E+08 | **1.8493E+04** | 2.2136E+04 |
| F16 | Ave | 2.6888E+03 | 4.1296E+03 | 4.3403E+03 | 3.7533E+03 | 3.0846E+03 | 3.2251E+03 | 2.8389E+03 | 3.3398E+03 | 3.5002E+03 | 3.3993E+03 | 2.5768E+03 | **2.4375E+03** |
| | Std | 5.1645E+02 | 5.0755E+02 | 9.2415E+02 | 5.8165E+02 | 3.8954E+02 | 5.6647E+02 | 4.1295E+02 | 6.9031E+02 | 8.4726E+02 | 5.7096E+02 | 3.8842E+02 | **3.7319E+02** |
| F17 | Ave | 2.1148E+03 | 2.8572E+03 | 2.7005E+03 | 2.6457E+03 | 2.4435E+03 | 2.4793E+03 | **2.0930E+03** | 2.8581E+03 | 2.5881E+03 | 2.6770E+03 | 2.2237E+03 | 2.1484E+03 |
| | Std | 4.0437E+02 | 3.4730E+02 | 6.0897E+02 | 5.1312E+02 | 4.4973E+02 | 5.0392E+02 | 3.1025E+02 | 4.5197E+02 | 4.5741E+02 | 4.6258E+02 | 3.1224E+02 | **3.0428E+02** |
| F18 | Ave | 1.0061E+07 | 6.3666E+07 | 7.2694E+07 | 1.0801E+07 | 1.8620E+07 | 5.9770E+06 | **1.9053E+06** | 3.2908E+06 | 2.4740E+07 | 1.4100E+07 | 4.6011E+06 | 2.5689E+06 |
| | Std | 8.9886E+06 | 2.7950E+07 | 6.7823E+07 | 9.8561E+06 | 3.3045E+07 | 5.9123E+06 | **1.4410E+06** | 3.0899E+06 | 1.9386E+07 | 1.4291E+07 | 3.2072E+06 | 2.0559E+06 |
| F19 | Ave | 1.4670E+06 | 1.3368E+08 | 2.3504E+07 | 1.7186E+06 | 5.7401E+06 | 1.3057E+05 | 1.2674E+06 | 2.5703E+04 | 3.4776E+06 | 4.2094E+06 | **1.1042E+04** | 1.9745E+04 |
| | Std | 2.0703E+07 | 2.6852E+08 | 1.7881E+07 | 1.8612E+06 | 3.4235E+07 | 4.0305E+05 | 4.2250E+06 | 3.2706E+05 | 9.1725E+06 | 6.6660E+06 | 5.3719E+04 | **2.5081E+04** |
| F20 | Ave | **2.4654E+03** | 2.9485E+03 | 2.8875E+03 | 2.8492E+03 | 2.7276E+03 | 2.8245E+03 | 2.4960E+03 | 2.8905E+03 | 2.6911E+03 | 2.7468E+03 | 2.4890E+03 | 2.4798E+03 |
| | Std | 4.2559E+02 | 2.2653E+02 | 3.6004E+02 | 2.8928E+02 | 3.7176E+02 | 3.6235E+02 | 2.7598E+02 | 3.4803E+02 | 3.4650E+02 | 3.4821E+02 | 4.1220E+02 | **1.9963E+02** |
| F21 | Ave | 2.4038E+03 | 2.5964E+03 | 2.6416E+03 | 2.5807E+03 | 2.5087E+03 | 2.5497E+03 | 2.4228E+03 | 2.5940E+03 | 2.5161E+03 | 2.5555E+03 | 2.4053E+03 | **2.3897E+03** |
| | Std | 7.2695E+01 | 4.8087E+01 | 1.1493E+02 | 1.1676E+02 | 7.0823E+01 | 8.3256E+01 | 5.5200E+01 | 1.1857E+02 | 5.9021E+01 | 8.7878E+01 | **3.6462E+01** | 3.7159E+01 |
| F22 | Ave | 5.6000E+03 | 1.0275E+04 | 8.4160E+03 | 7.2253E+03 | 6.5539E+03 | 5.1503E+03 | 4.9175E+03 | 6.4982E+03 | **3.3994E+03** | 5.1048E+03 | 3.9555E+03 | 3.6629E+03 |
| | Std | 1.4270E+03 | **5.7512E+02** | 1.1777E+03 | 9.8091E+02 | 1.2246E+03 | 1.4109E+03 | 9.1222E+02 | 1.2615E+03 | 1.1761E+03 | 2.4377E+03 | 3.0268E+03 | 1.2172E+03 |
| F23 | Ave | **2.7816E+03** | 3.0714E+03 | 3.1365E+03 | 3.2916E+03 | 2.8309E+03 | 2.9908E+03 | 2.8298E+03 | 3.0889E+03 | 3.0304E+03 | 3.0065E+03 | 2.8036E+03 | 2.7880E+03 |
| | Std | 6.0544E+01 | 8.5263E+01 | 2.0203E+02 | 2.1001E+02 | 7.9693E+01 | 1.3022E+02 | 1.0269E+02 | 1.3474E+02 | 1.3877E+02 | 1.2282E+02 | **5.2603E+01** | 5.6091E+01 |
| F24 | Ave | 2.9643E+03 | 3.2471E+03 | 3.2422E+03 | 3.4791E+03 | 2.9921E+03 | 3.1596E+03 | 2.9608E+03 | 3.3061E+03 | 3.1518E+03 | 3.1606E+03 | 2.9517E+03 | **2.9494E+03** |
| | Std | 1.3145E+02 | 7.1059E+01 | 1.2210E+02 | 2.3066E+02 | 5.7286E+01 | 1.6350E+02 | 7.2485E+01 | 1.5576E+02 | 1.1740E+02 | 1.8211E+02 | 6.9748E+01 | **5.0311E+01** |
| F25 | Ave | 3.0208E+03 | 3.6714E+03 | 3.2205E+03 | 3.0164E+03 | 3.3123E+03 | 2.9469E+03 | 3.0167E+03 | 3.0259E+03 | 3.0993E+03 | 3.0702E+03 | 2.9370E+03 | **2.9087E+03** |
| | Std | 2.6646E+02 | 1.3565E+03 | 5.4746E+02 | 1.8523E+02 | 3.3116E+03 | 7.5839E+01 | 1.9266E+02 | 5.8987E+02 | 4.0774E+02 | 8.3552E+02 | 1.5869E+02 | **4.0561E+01** |
| F26 | Ave | **5.0588E+03** | 7.9045E+03 | 8.7849E+03 | 7.7998E+03 | 5.8921E+03 | 7.0150E+03 | 5.5155E+03 | 7.3735E+03 | 6.5270E+03 | 6.7746E+03 | 5.5256E+03 | 5.3374E+03 |
| | Std | 6.0573E+02 | 9.3900E+02 | 1.2774E+03 | 1.9442E+03 | 8.4773E+02 | 1.4483E+03 | 1.3756E+03 | 1.7459E+03 | 1.4978E+03 | 1.8342E+03 | 7.4729E+02 | **5.7230E+02** |
| F27 | Ave | 3.2705E+03 | 3.5432E+03 | 3.4774E+03 | 3.6188E+03 | **3.2474E+03** | 3.2949E+03 | 3.3298E+03 | 3.6443E+03 | 3.4492E+03 | 3.3307E+03 | 3.2980E+03 | 3.2736E+03 |
| | Std | **8.7344E+01** | 2.6798E+02 | 6.2245E+02 | 6.1833E+02 | 1.1357E+02 | 2.0707E+02 | 1.5657E+02 | 4.3875E+02 | 2.4570E+02 | 2.4823E+02 | 1.1470E+02 | 1.0710E+02 |
| F28 | Ave | 3.5612E+03 | 4.4333E+03 | 3.9953E+03 | 3.5030E+03 | 4.4304E+03 | 3.3162E+03 | 3.3944E+03 | 3.3516E+03 | 3.8405E+03 | 3.6552E+03 | 3.3635E+03 | **3.2952E+03** |
| | Std | 3.4742E+02 | 8.3449E+02 | 6.0103E+02 | 3.9109E+02 | 1.7225E+03 | **1.9350E+02** | 4.8079E+02 | 9.3552E+02 | 5.6756E+02 | 1.9125E+03 | 4.0149E+02 | 2.3555E+02 |
| F29 | Ave | 3.9735E+03 | 5.2094E+03 | 5.4046E+03 | 5.1221E+03 | 4.1937E+03 | 4.3967E+03 | 4.3817E+03 | 4.5706E+03 | 4.8403E+03 | 4.4849E+03 | 4.0925E+03 | **3.9647E+03** |
| | Std | 3.4888E+02 | 1.0270E+03 | 1.7393E+03 | 1.0818E+03 | 5.1866E+02 | 4.3411E+02 | 7.4497E+02 | 8.3075E+02 | 1.7602E+03 | 1.0720E+03 | 4.3798E+02 | **3.3917E+02** |
| F30 | Ave | 1.1195E+07 | 1.8918E+08 | 7.5216E+07 | 8.6254E+06 | 8.4937E+05 | 1.4797E+06 | 5.8505E+06 | 8.9697E+05 | 3.3755E+07 | 4.3829E+06 | 2.0597E+05 | **1.7091E+05** |
| | Std | 6.1634E+07 | 3.6240E+08 | 1.9305E+08 | 4.4409E+07 | 4.1963E+08 | 1.2776E+07 | 7.4631E+07 | 5.1313E+07 | 9.0563E+07 | 5.4511E+07 | 7.5222E+06 | **3.1519E+06** |
| (W\|T\|L) | | (1\|29\|0) | (0\|10\|20) | (0\|24\|6) | (0\|28\|2) | (1\|28\|1) | (0\|30\|0) | (2\|28\|0) | (0\|29\|1) | (0\|30\|0) | (0\|30\|0) | (1\|29\|0) | (**25**\|5\|0) |

Table 6

Experimental results of 12 algorithms on CEC 2017 (Dim=100).

| ID | | GWO | SCA | WOA | HHO | MFO | AVOA | CSA | CPSOGSA | AO | DBO | SO | MISO |
|---|---|---|---|---|---|---|---|---|---|---|---|---|---|
| F1 | Ave | 5.6540E+10 | 2.1991E+11 | 1.1212E+11 | 4.8728E+10 | 1.4948E+11 | 1.3438E+10 | 5.9170E+10 | 7.5125E+10 | 9.3880E+10 | 7.7619E+10 | 1.4604E+10 | **1.8907E+09** |
| | Std | 9.5886E+09 | 1.8556E+10 | 8.9489E+09 | 7.7529E+09 | 4.7609E+10 | 4.7707E+09 | 1.0166E+10 | 1.8764E+10 | 9.2052E+09 | 5.8938E+10 | 3.3685E+09 | **1.2204E+09** |
| F2 | Ave | 9.7969E+147 | 1.1384E+161 | 5.4334E+173 | 1.2898E+157 | 3.7537E+163 | 9.3194E+140 | 4.6039E+146 | 2.3783E+170 | 4.5203E+157 | 3.4707E+162 | 9.3088E+135 | **1.2452E+124** |
| | Std | 5.3660E+148 | Inf | Inf | Inf | Inf | 5.0744E+141 | 1.6332E+147 | Inf | Inf | Inf | 4.7532E+136 | **6.5605E+124** |
| F3 | Ave | 5.3514E+05 | 6.2788E+05 | 9.3488E+05 | 3.6868E+05 | 1.0319E+06 | 3.9744E+05 | 4.6070E+05 | 8.2653E+05 | **3.5713E+05** | 7.6963E+05 | 3.7359E+05 | 6.8083E+05 |
| | Std | 8.3589E+04 | 8.3954E+04 | 1.4989E+05 | 9.8277E+04 | 1.7224E+05 | 1.0701E+05 | 7.8614E+04 | 1.2177E+05 | **9.2871E+03** | 3.2629E+05 | 3.8776E+04 | 1.1546E+05 |
| F4 | Ave | 6.2160E+03 | 5.1808E+04 | 2.2473E+04 | 9.6400E+03 | 2.5687E+04 | 2.5795E+03 | 8.5632E+03 | 1.7317E+04 | 1.8403E+04 | 1.4857E+04 | 2.7054E+03 | **1.2876E+03** |
| | Std | 2.7778E+03 | 9.1863E+03 | 3.7985E+03 | 1.8486E+03 | 1.2311E+04 | 5.0716E+02 | 1.9388E+03 | 6.3317E+03 | 3.7778E+03 | 1.5567E+04 | 4.4074E+02 | **1.5618E+02** |
| F5 | Ave | 1.2471E+03 | 2.0387E+03 | 1.9868E+03 | 1.6706E+03 | 1.9301E+03 | 1.4255E+03 | 1.5417E+03 | 1.8218E+03 | 1.6693E+03 | 1.7453E+03 | 1.1840E+03 | **1.0501E+03** |
| | Std | 6.3601E+01 | 5.3860E+01 | 1.4567E+02 | **4.0358E+01** | 1.6890E+02 | 7.2779E+01 | 8.4053E+01 | 1.2020E+02 | 6.3533E+01 | 2.1779E+02 | 4.8042E+01 | 6.6792E+01 |
| F6 | Ave | 6.4606E+02 | 7.0454E+02 | 7.0561E+02 | 6.9003E+02 | 6.8502E+02 | 6.6781E+02 | 6.7236E+02 | 6.8573E+02 | 6.8917E+02 | 6.7743E+02 | 6.4870E+02 | **6.3318E+02** |
| | Std | 4.4034E+00 | 5.9182E+00 | 8.6416E+00 | **3.9289E+00** | 7.5197E+00 | 4.1997E+00 | 4.2346E+00 | 6.0450E+00 | 4.1800E+00 | 1.0229E+01 | 4.9517E+00 | 6.0805E+00 |
| F7 | Ave | 2.1706E+03 | 4.1257E+03 | 3.8379E+03 | 3.7570E+03 | 5.4538E+03 | 3.2052E+03 | 3.3542E+03 | 6.6725E+03 | 3.5308E+03 | 2.9308E+03 | 2.2261E+03 | **1.6544E+03** |
| | Std | 1.6185E+02 | 2.4867E+02 | 1.5982E+02 | 1.4648E+02 | 8.0013E+02 | 1.4811E+02 | 2.0693E+02 | 4.1125E+02 | 1.3658E+02 | 2.3326E+02 | **1.0249E+02** | 1.0269E+02 |
| F8 | Ave | 1.5638E+03 | 2.4481E+03 | 2.3801E+03 | 2.1289E+03 | 2.2817E+03 | 1.8610E+03 | 1.8894E+03 | 2.1597E+03 | 2.1508E+03 | 2.1381E+03 | 1.4838E+03 | **1.4009E+03** |
| | Std | 1.1593E+02 | 8.9046E+01 | 8.7206E+01 | **5.5687E+01** | 1.6012E+02 | 8.9719E+01 | 1.0261E+02 | 1.5419E+02 | 6.7944E+01 | 2.4121E+02 | 7.6151E+01 | 7.5578E+01 |
| F9 | Ave | 4.0235E+04 | 9.1325E+04 | 7.6881E+04 | 6.9856E+04 | 5.9003E+04 | 3.0424E+04 | 3.7161E+04 | 4.5282E+04 | 6.5496E+04 | 7.4454E+04 | 2.8742E+04 | **1.2143E+04** |
| | Std | 1.2434E+04 | 9.4362E+03 | 1.4736E+04 | 4.5371E+03 | 1.0179E+04 | 3.0659E+03 | 4.8786E+03 | 4.4376E+03 | 6.4277E+03 | 1.0394E+04 | 7.2060E+03 | **2.9499E+03** |
| F10 | Ave | 1.9818E+04 | 3.2838E+04 | 2.9528E+04 | 2.4629E+04 | 2.0446E+04 | 1.8723E+04 | 2.4513E+04 | **1.7143E+04** | 2.5361E+04 | 2.7720E+04 | 3.1230E+04 | 1.8165E+04 |
| | Std | 5.0370E+03 | **4.9076E+02** | 1.3460E+03 | 1.8322E+03 | 1.5899E+03 | 2.0265E+03 | 2.0922E+03 | 1.3220E+03 | 2.1422E+03 | 4.8502E+03 | 1.6895E+03 | 1.6152E+03 |
| F11 | Ave | **9.1679E+04** | 1.7683E+05 | 3.1703E+05 | 1.4431E+05 | 2.2684E+05 | 1.0809E+05 | 1.2705E+05 | 2.1729E+05 | 3.6671E+05 | 2.1578E+05 | 1.3854E+05 | 1.2117E+05 |
| | Std | **1.6269E+04** | 2.9329E+04 | 1.0652E+05 | 4.0309E+04 | 6.6192E+04 | 2.3932E+04 | 2.7764E+04 | 5.8579E+04 | 6.7949E+04 | 5.5917E+04 | 2.4845E+04 | 3.3032E+04 |
| F12 | Ave | 1.2534E+10 | 1.0085E+11 | 3.2186E+10 | 1.2280E+10 | 4.5442E+10 | 1.3255E+09 | 7.3669E+09 | 1.6539E+10 | 3.3005E+10 | 8.9725E+09 | 1.7080E+09 | **2.5662E+08** |
| | Std | 7.2719E+09 | 1.2526E+10 | 7.6221E+09 | 4.2305E+09 | 2.0716E+10 | 5.5036E+08 | 2.1464E+09 | 8.1623E+09 | 8.1350E+09 | 9.1331E+09 | 7.4614E+08 | **1.3152E+08** |
| F13 | Ave | 1.5465E+09 | 1.8641E+10 | 3.0038E+09 | 2.6503E+08 | 4.6347E+09 | 5.3149E+05 | 9.0121E+07 | 1.0315E+09 | 3.4309E+09 | 3.2839E+08 | 4.3647E+06 | **4.5253E+05** |
| | Std | 1.0659E+09 | 3.8010E+09 | 1.1436E+09 | 1.4148E+08 | 4.9196E+09 | 1.4074E+06 | 7.5592E+07 | 1.0559E+09 | 9.2168E+08 | 2.8719E+08 | 4.1719E+06 | **5.2537E+05** |
| F14 | Ave | 1.1336E+07 | 6.7675E+07 | 2.4713E+07 | 9.7786E+06 | 1.8890E+07 | 9.0511E+06 | 6.5560E+06 | 8.1846E+06 | 2.1944E+07 | 1.7232E+07 | 7.1620E+06 | **5.1674E+06** |
| | Std | 5.7487E+06 | 3.5402E+07 | 9.3611E+06 | **3.1306E+06** | 1.4125E+07 | 4.1356E+06 | 4.2921E+06 | 7.0804E+06 | 9.4468E+06 | 8.7785E+06 | 3.2121E+06 | 4.2818E+06 |
| F15 | Ave | 3.6606E+08 | 6.9049E+09 | 4.8062E+08 | 1.8484E+07 | 1.8397E+09 | 1.2001E+05 | 3.3613E+05 | 1.0677E+08 | 5.4868E+08 | 8.8851E+07 | 4.8129E+05 | **1.1116E+05** |
| | Std | 6.7767E+08 | 1.6726E+09 | 2.2612E+08 | 1.5392E+07 | 2.3301E+09 | 2.7097E+05 | 2.2873E+05 | 3.7229E+08 | 3.5637E+08 | 1.3482E+08 | 5.6199E+05 | **1.1483E+05** |
| F16 | Ave | 6.6456E+03 | 1.4921E+04 | 1.6441E+04 | 1.0419E+04 | 8.7365E+03 | 7.8072E+03 | 9.1659E+03 | 8.4259E+03 | 1.1637E+04 | 9.4257E+03 | 7.1715E+03 | **6.1916E+03** |
| | Std | 6.6062E+02 | 1.2226E+03 | 1.8288E+03 | 1.2656E+03 | 1.2405E+03 | 1.1571E+03 | 1.1869E+03 | 9.8679E+02 | 1.1610E+03 | 1.4387E+03 | 1.1573E+03 | **5.6139E+02** |
| F17 | Ave | 5.8092E+03 | 6.7067E+04 | 2.8652E+04 | 8.1335E+03 | 3.2148E+04 | 6.3340E+03 | 6.7582E+03 | 7.2174E+03 | 1.8072E+04 | 9.4955E+03 | 5.5827E+03 | **5.0209E+03** |
| | Std | 1.0552E+03 | 6.5494E+04 | 3.5683E+04 | 1.4502E+03 | 8.6207E+04 | 6.8083E+02 | 8.3966E+02 | 7.3934E+02 | 8.9681E+03 | 9.7225E+02 | **3.9198E+02** | 5.5196E+02 |
| F18 | Ave | 9.0292E+06 | 1.3242E+08 | 2.1812E+07 | 1.2351E+07 | 2.8449E+07 | 6.8143E+06 | 8.8152E+06 | 8.2782E+06 | 2.3625E+07 | 2.7778E+07 | 1.0591E+07 | **5.7450E+06** |
| | Std | 4.9339E+06 | 5.8038E+07 | 1.2359E+07 | 7.5191E+06 | 2.8044E+07 | 4.5468E+06 | 4.4645E+06 | 5.0103E+06 | 1.1722E+07 | 1.5069E+07 | 3.9743E+06 | **3.2629E+06** |
| F19 | Ave | 2.2017E+08 | 5.3881E+09 | 4.1591E+08 | 3.3707E+07 | 1.4619E+09 | 3.8410E+06 | 4.0926E+07 | 2.1620E+08 | 5.5300E+08 | 7.2901E+07 | 2.5194E+06 | **8.2436E+05** |
| | Std | 2.9456E+08 | 1.6753E+09 | 2.0604E+08 | 1.4667E+07 | 1.4344E+09 | 3.1437E+06 | 3.8128E+07 | 5.4054E+08 | 2.7681E+08 | 5.8245E+07 | 2.1461E+06 | **9.4208E+05** |
| F20 | Ave | 5.8789E+03 | 8.0612E+03 | 6.9870E+03 | 6.2599E+03 | 5.9312E+03 | 6.0186E+03 | 5.9232E+03 | 5.7813E+03 | 6.2713E+03 | 7.1226E+03 | 7.3200E+03 | **5.5048E+03** |
| | Std | 1.3017E+03 | **3.5407E+02** | 6.9010E+02 | 5.6772E+02 | 5.6416E+02 | 6.1625E+02 | 1.3006E+03 | 4.5110E+02 | 5.1133E+02 | 8.7738E+02 | 4.7633E+02 | 3.6958E+02 |
| F21 | Ave | 3.0867E+03 | 4.1934E+03 | 4.4631E+03 | 4.3580E+03 | 3.7979E+03 | 3.7132E+03 | 3.4572E+03 | 4.2397E+03 | 4.3025E+03 | 4.0417E+03 | 3.1057E+03 | **2.9665E+03** |
| | Std | 1.0848E+02 | 1.0058E+02 | 1.9169E+02 | 2.0509E+02 | 1.6771E+02 | 2.0350E+02 | 1.0973E+02 | 2.4157E+02 | 2.5125E+02 | 1.8348E+02 | 7.3983E+01 | **6.9033E+01** |
| F22 | Ave | 2.2082E+04 | 3.5360E+04 | 3.1712E+04 | 2.7143E+04 | 2.2087E+04 | 2.1825E+04 | 2.7154E+04 | 2.0585E+04 | 2.7572E+04 | 2.8663E+04 | 3.3310E+04 | **2.0076E+04** |
| | Std | 3.5665E+03 | **7.5890E+02** | 1.4800E+03 | 1.3030E+03 | 1.4271E+03 | 1.9210E+03 | 1.8863E+03 | 1.7469E+03 | 1.6074E+03 | 4.5160E+03 | 1.8108E+03 | 2.0583E+03 |
| F23 | Ave | 3.7297E+03 | 5.2300E+03 | 5.2984E+03 | 5.8292E+03 | 3.9176E+03 | 4.3771E+03 | 4.2312E+03 | 5.1923E+03 | 5.0060E+03 | 4.7147E+03 | 3.7305E+03 | **3.6182E+03** |
| | Std | 1.2794E+02 | 1.4105E+02 | 2.2531E+02 | 4.7898E+02 | 1.2550E+02 | 1.8014E+02 | 2.0007E+02 | 3.6489E+02 | 2.6409E+02 | 1.9886E+02 | **8.2174E+01** | 9.4262E+01 |
| F24 | Ave | 4.5212E+03 | 7.2782E+03 | 6.7999E+03 | 8.5011E+03 | 4.5425E+03 | 5.3320E+03 | 5.3301E+03 | 6.5852E+03 | 6.8378E+03 | 6.1983E+03 | 4.8200E+03 | **4.4096E+03** |
| | Std | **1.6426E+02** | 2.8510E+02 | 4.4835E+02 | 6.3757E+02 | 1.8291E+02 | 3.2480E+02 | 2.9049E+02 | 4.8279E+02 | 4.5983E+02 | 5.1152E+02 | 2.3094E+02 | 1.9660E+02 |
| F25 | Ave | 7.4386E+03 | 2.2396E+04 | 1.1220E+04 | 6.7670E+03 | 2.0085E+04 | 4.7568E+03 | 8.5984E+03 | 1.0391E+04 | 9.5029E+03 | 8.7254E+03 | 5.6307E+03 | **4.1412E+03** |
| | Std | 6.9144E+02 | 2.6350E+03 | 1.1167E+03 | 6.8890E+02 | 8.0034E+03 | 3.6115E+02 | 9.1995E+02 | 1.6431E+03 | 9.4797E+02 | 5.1682E+03 | 5.2772E+02 | **1.9613E+02** |
| F26 | Ave | 1.7541E+04 | 4.1521E+04 | 3.8267E+04 | 3.1303E+04 | 2.0037E+04 | 2.6239E+04 | 2.6658E+04 | 3.3051E+04 | 3.3565E+04 | 2.6677E+04 | 1.9839E+04 | **1.6637E+04** |
| | Std | **1.6282E+03** | 2.3592E+03 | 3.1781E+03 | 2.4646E+03 | 1.9754E+03 | 3.4511E+03 | 3.5697E+03 | 2.9320E+03 | 2.5711E+03 | 3.4182E+03 | 1.7190E+03 | 1.8515E+03 |
| F27 | Ave | 4.3167E+03 | 8.3374E+03 | 6.3671E+03 | 6.9793E+03 | 4.0613E+03 | 4.4131E+03 | 4.8774E+03 | 5.9089E+03 | 7.2914E+03 | 4.5682E+03 | 4.2599E+03 | **4.0375E+03** |
| | Std | 1.9435E+02 | 6.0394E+02 | 1.1595E+03 | 1.0701E+03 | 2.2282E+02 | 2.9278E+02 | 4.5821E+02 | 7.9612E+02 | 8.8707E+02 | 4.3792E+02 | 1.4849E+02 | **1.3762E+02** |
| F28 | Ave | 9.5807E+03 | 2.8044E+04 | 1.4351E+04 | 9.4679E+03 | 1.9723E+04 | **5.9422E+03** | 1.0213E+04 | 1.0056E+04 | 1.4675E+04 | 1.6778E+04 | 9.8357E+03 | 7.2106E+03 |
| | Std | 1.6163E+03 | 2.3766E+03 | 1.2105E+03 | 9.9438E+02 | 2.3624E+03 | **5.5150E+02** | 1.1976E+03 | 2.1928E+03 | 1.2365E+03 | 6.9790E+03 | 1.3536E+03 | 1.3241E+03 |
| F29 | Ave | 9.4828E+03 | 3.5526E+04 | 1.9802E+04 | 1.3159E+04 | 1.1576E+04 | 9.9805E+03 | 1.3578E+04 | 1.3487E+04 | 1.7456E+04 | 1.2267E+04 | 8.8809E+03 | **8.1596E+03** |
| | Std | 9.4755E+02 | 1.3891E+04 | 2.8667E+03 | 1.1716E+03 | 3.4383E+03 | 1.1476E+03 | 1.4475E+03 | 1.8350E+03 | 3.5812E+03 | 1.8306E+03 | 7.4822E+02 | **6.3219E+02** |
| F30 | Ave | 1.6061E+09 | 1.2616E+10 | 3.1856E+09 | 7.6680E+08 | 2.3078E+09 | 6.9144E+07 | 6.0464E+08 | 1.3207E+09 | 4.2183E+09 | 2.9705E+08 | 2.4017E+07 | **6.7504E+06** |
| | Std | 1.3648E+09 | 2.1667E+09 | 1.4826E+09 | 4.7967E+08 | 1.2817E+09 | 4.2580E+07 | 3.0096E+08 | 8.8938E+08 | 1.4792E+09 | 1.6087E+08 | 1.7453E+07 | **3.4855E+06** |
| (W\|T\|L) | | (1\|29\|0) | (0\|9\|21) | (0\|26\|4) | (1\|27\|2) | (0\|29\|1) | (2\|28\|0) | (0\|30\|0) | (1\|28\|1) | (0\|29\|1) | (0\|30\|0) | (0\|30\|0) | (**25**\|5\|0) |

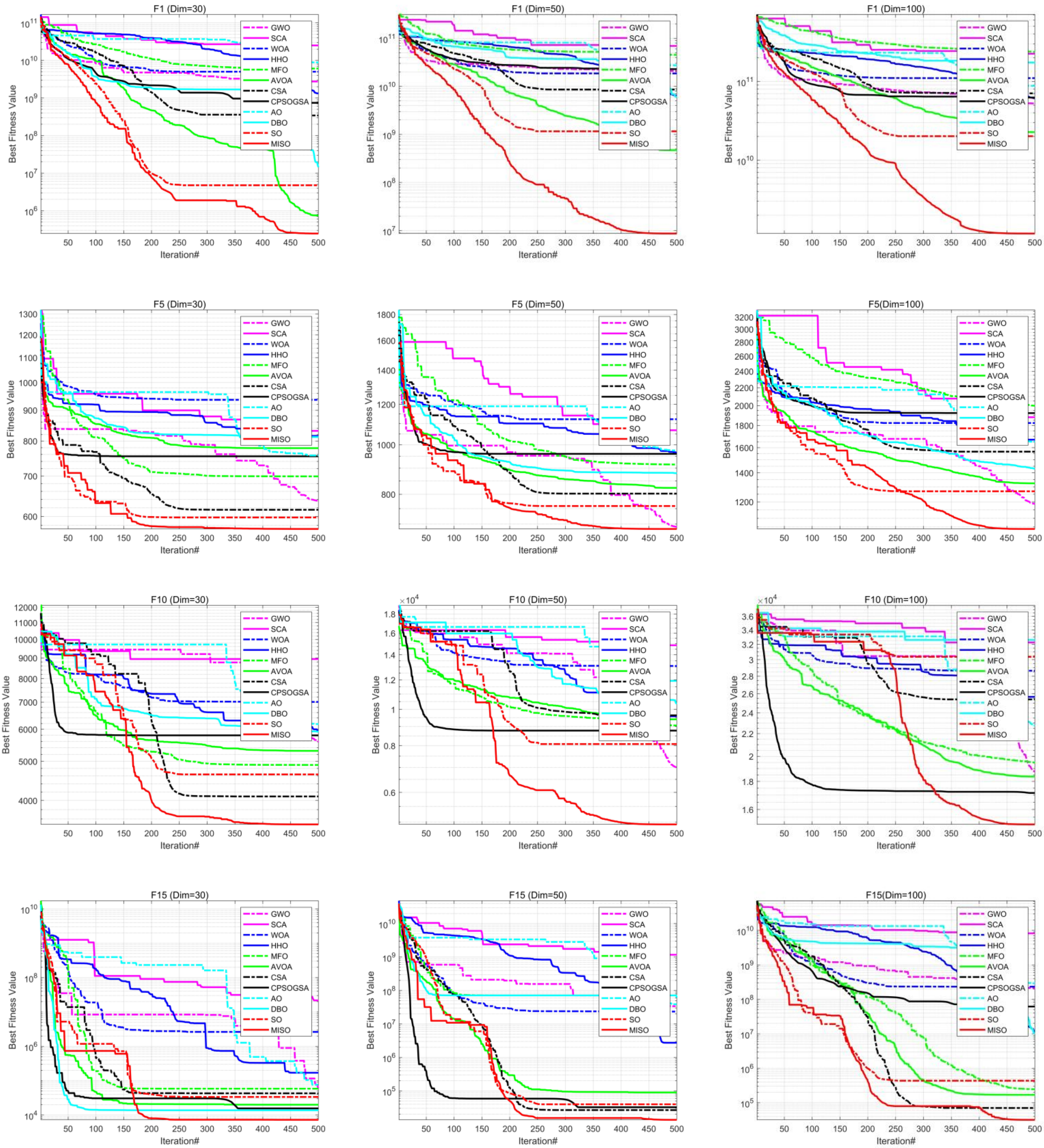
F1 (Dim=30)
F1 (Dim=50)
F1 (Dim=100)
F5 (Dim=30)
F5 (Dim=50)
F5(Dim=100)
F10 (Dim=30)
F10 (Dim=50)
F10 (Dim=100)
F15 (Dim=30)
F15 (Dim=50)
F15(Dim=100)
Best Fitness Value
Iteration#
GWO
SCA
WOA
HHO
MFO
AVOA
CSA
CPSOGSA
AO
DBO
SO
MISO

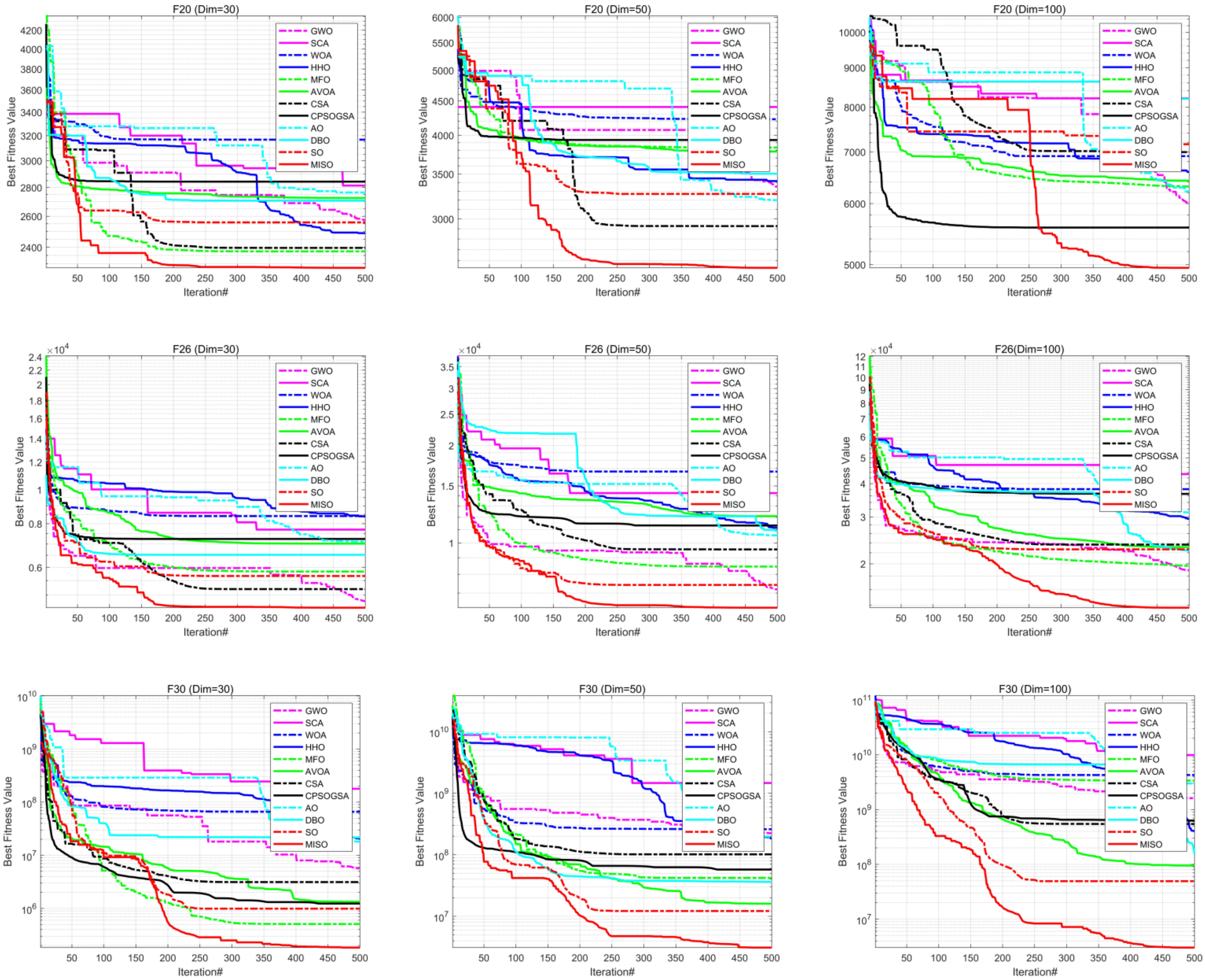

Fig. 7. Comparison of convergence curves of 12 algorithms on CEC2017.

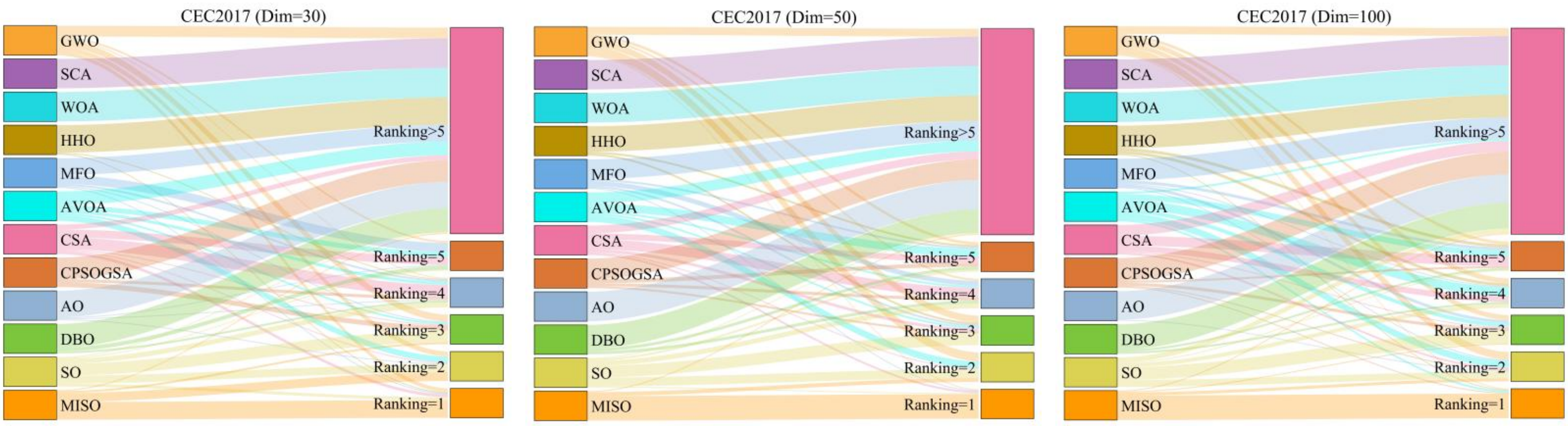

Fig. 8. The ranking Sankey of different competitors on CEC2017.

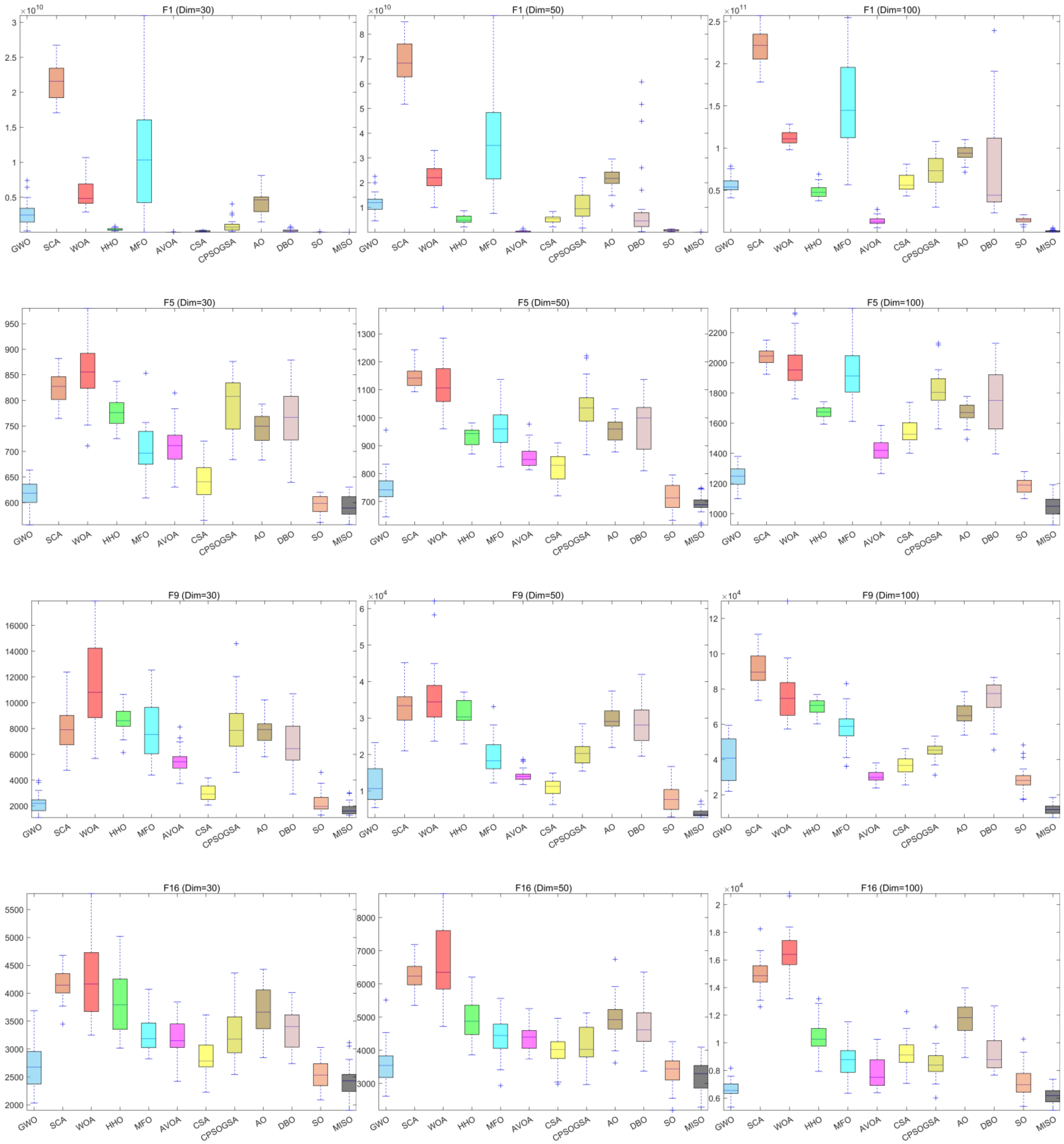
F1 (Dim=30)
F1 (Dim=50)
F1 (Dim=100)
F5 (Dim=30)
F5 (Dim=50)
F5 (Dim=100)
F9 (Dim=30)
F9 (Dim=50)
F9 (Dim=100)
F16 (Dim=30)
F16 (Dim=50)
F16 (Dim=100)
GWO
SCA
WOA
HHO
MFO
AVOA
CSA
CPSOGSA
AO
DBO
SO
MISO

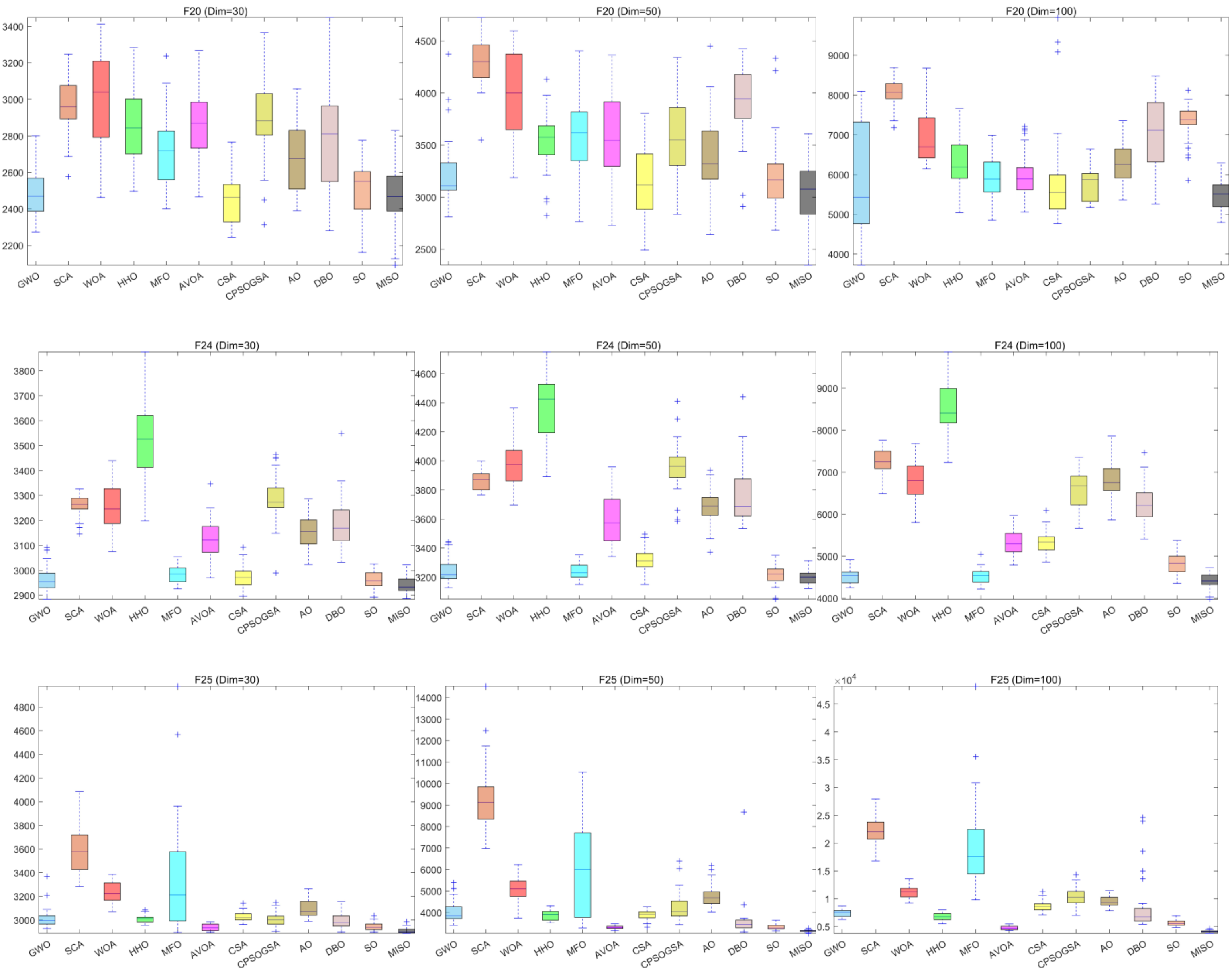

Fig. 9. Boxplot analysis of competitor algorithms on the CEC2017 test suite.

4.4.2 Comparison with other competitive algorithms in CEC 2022

We employ the latest CEC2022 suite to evaluate various competitors with dimensions set to 10, 20, respectively. The experimental results are reported in Table 7 and Table 8. The statistical results show that as the dimension increases, MISO achieves the best results in 7 and 9 out of 12 cases, respectively, without any failure. It reveals the good robustness of MISO. Fig. 10 illustrates the convergence rate of different competitors in the optimization process, and it can be seen that MISO has a strong convergence performance. Fig. 11 displays the Friedman ranking of MISO versus other optimizers on the test set using stacked bar chart. In all dimensions, MISO received the most first places. In order to compare the algorithm stability, Fig. 12 shows the results of the 12 algorithms on the CEC2022 test set in the form of box plots, and it can be seen that MISO performance is the most stable. As a result, MISO outperforms other algorithms, making it the best performer overall among the considered test functions and competing algorithms.

Table 7

Experimental results of 12 algorithms on CEC2022 (Dim=10).

| ID | | GWO | SCA | WOA | HHO | MFO | AVOA | CSA | CPSOGSA | AO | DBO | SO | MISO |
|---|---|---|---|---|---|---|---|---|---|---|---|---|---|
| F1 | Ave | 2.9470E+03 | 2.4903E+03 | 2.4799E+04 | 1.0960E+03 | 8.4302E+03 | 4.7220E+02 | 6.5469E+02 | 2.6256E+03 | 4.8794E+03 | 1.4222E+03 | 7.9639E+02 | **4.0919E+02** |
| | Std | 2.0400E+03 | 1.1029E+03 | 1.1608E+04 | 5.2816E+02 | 8.4252E+03 | 3.1262E+02 | 4.3636E+02 | 3.3130E+03 | 2.8121E+03 | 1.2965E+03 | 7.4182E+02 | **1.3040E+02** |
| F2 | Ave | 4.2679E+02 | 4.8634E+02 | 4.8707E+02 | 4.4462E+02 | 4.1855E+02 | 4.1793E+02 | 4.0508E+02 | 4.3027E+02 | 4.5806E+02 | 4.4327E+02 | 4.0462E+02 | **4.0400E+02** |
| | Std | 2.3390E+01 | 2.1292E+01 | 1.0939E+02 | 3.8996E+01 | 1.7936E+01 | 2.9637E+01 | 7.7285E+00 | 3.4338E+01 | 7.5721E+01 | 6.2900E+01 | 4.6144E+00 | **2.9882E+00** |
| F3 | Ave | 6.0142E+02 | 6.2127E+02 | 6.4241E+02 | 6.4085E+02 | 6.0300E+02 | 6.2367E+02 | 6.0611E+02 | 6.4136E+02 | 6.1992E+02 | 6.1026E+02 | 6.0186E+02 | **6.0019E+02** |
| | Std | 2.1029E+00 | 4.1038E+00 | 1.5850E+01 | 1.1774E+01 | 4.9145E+00 | 1.4686E+01 | 3.7434E+00 | 1.2980E+01 | 6.7363E+00 | 8.7592E+00 | 2.4252E+00 | **2.4149E-01** |
| F4 | Ave | 8.1716E+02 | 8.4415E+02 | 8.4063E+02 | 8.2731E+02 | 8.3089E+02 | 8.3186E+02 | **8.1128E+02** | 8.4043E+02 | 8.2305E+02 | 8.3380E+02 | 8.1522E+02 | 8.1533E+02 |
| | Std | 8.8163E+00 | 6.5942E+00 | 1.5067E+01 | 8.8555E+00 | 1.4490E+01 | 1.2620E+01 | **4.9461E+00** | 1.2945E+01 | 7.3091E+00 | 1.3556E+01 | 5.5065E+00 | 5.0657E+00 |
| F5 | Ave | 9.2640E+02 | 1.0462E+03 | 1.5213E+03 | 1.4227E+03 | 1.0639E+03 | 1.2588E+03 | 9.1460E+02 | 1.6804E+03 | 1.0453E+03 | 1.0147E+03 | 9.3331E+02 | **9.1085E+02** |
| | Std | 6.4584E+01 | 6.9495E+01 | 4.7625E+02 | 1.3384E+02 | 1.8850E+02 | 2.2515E+02 | 2.4333E+01 | 4.3675E+02 | 1.3849E+02 | 1.2375E+02 | 3.7544E+01 | **2.3462E+01** |
| F6 | Ave | 6.3138E+03 | 3.6414E+06 | 5.9151E+03 | 6.5647E+03 | 4.1158E+03 | 4.3977E+03 | **2.1338E+03** | 3.3524E+03 | 3.9195E+04 | 4.7941E+03 | 3.4434E+03 | 4.1084E+03 |
| | Std | 2.5717E+03 | 2.8278E+06 | 3.1785E+03 | 3.5119E+03 | 1.8402E+03 | 2.2517E+03 | **5.6676E+02** | 1.7021E+03 | 3.8362E+04 | 2.2436E+03 | 1.6094E+03 | 2.2381E+03 |
| F7 | Ave | 2.0328E+03 | 2.0643E+03 | 2.0848E+03 | 2.0690E+03 | 2.0364E+03 | 2.0600E+03 | 2.0266E+03 | 2.0866E+03 | 2.0526E+03 | 2.0415E+03 | 2.0294E+03 | **2.0241E+03** |
| | Std | 1.7384E+01 | 1.0615E+01 | 2.1247E+01 | 2.7758E+01 | 2.6551E+01 | 2.5729E+01 | **7.2177E+00** | 4.5696E+01 | 1.9587E+01 | 1.7673E+01 | 1.5863E+01 | 1.4021E+01 |
| F8 | Ave | 2.2288E+03 | 2.2339E+03 | 2.2364E+03 | 2.2368E+03 | 2.2252E+03 | 2.2265E+03 | **2.2210E+03** | 2.3111E+03 | 2.2299E+03 | 2.2340E+03 | 2.2221E+03 | 2.2216E+03 |
| | Std | 2.3133E+01 | 3.5387E+00 | 1.0111E+01 | 1.2660E+01 | 4.1304E+00 | 4.5766E+00 | 6.6539E+00 | 7.1113E+01 | 3.7543E+00 | 2.7283E+01 | **2.2996E+00** | 4.1240E+00 |
| F9 | Ave | 2.5807E+03 | 2.5829E+03 | 2.6160E+03 | 2.6313E+03 | 2.5340E+03 | 2.5372E+03 | 2.5331E+03 | 2.5414E+03 | 2.6198E+03 | 2.5664E+03 | 2.5302E+03 | **2.5293E+03** |
| | Std | 4.1796E+01 | 2.7446E+01 | 6.0082E+01 | 4.4940E+01 | 1.8565E+01 | 1.2379E+01 | 6.2663E+00 | 3.4959E+01 | 4.6032E+01 | 4.7191E+01 | 3.6749E+00 | **8.3255E-04** |
| F10 | Ave | 2.5700E+03 | **2.5134E+03** | 2.6840E+03 | 2.6021E+03 | 2.5310E+03 | 2.5853E+03 | 2.5398E+03 | 2.6794E+03 | 2.5688E+03 | 2.5459E+03 | 2.5193E+03 | 2.5261E+03 |
| | Std | 5.7942E+01 | **3.7520E+01** | 2.7515E+02 | 1.4796E+02 | 5.5601E+01 | 6.6054E+01 | 5.6232E+01 | 3.9121E+02 | 5.9953E+01 | 6.5252E+01 | 5.7378E+01 | 5.2402E+01 |
| F11 | Ave | 2.8025E+03 | 2.8125E+03 | 2.9467E+03 | 2.8159E+03 | 2.7674E+03 | 2.7266E+03 | **2.6508E+03** | 2.8100E+03 | 2.7194E+03 | 2.7507E+03 | 2.6802E+03 | 2.6984E+03 |
| | Std | 1.5415E+02 | 1.1915E+02 | 1.9148E+02 | 1.2737E+02 | 1.4306E+02 | 1.5295E+02 | 8.3989E+01 | 2.9608E+02 | **6.2899E+01** | 1.4651E+02 | 1.1642E+02 | 1.3424E+02 |
| F12 | Ave | 2.8677E+03 | 2.8720E+03 | 2.8987E+03 | 2.9285E+03 | **2.8638E+03** | 2.8678E+03 | 2.8641E+03 | 2.8901E+03 | 2.8685E+03 | 2.8765E+03 | 2.8756E+03 | 2.8692E+03 |
| | Std | 7.8313E+00 | 2.0013E+00 | 4.8045E+01 | 4.7140E+01 | **1.5850E+00** | 5.4405E+00 | 1.8268E+00 | 2.9852E+01 | 3.8097E+00 | 1.5087E+01 | 1.0195E+01 | 4.0484E+00 |
| (W\|T\|L) | | (0\|12\|0) | (0\|9\|3) | (0\|8\|4) | (0\|10\|2) | (1\|11\|0) | (0\|12\|0) | (3\|9\|0) | (0\|9\|3) | (0\|12\|0) | (0\|12\|0) | (1\|11\|0) | (**7**\|5\|0) |

Table 8

Experimental results of 12 algorithms on CEC2022 (Dim=20).

| ID | | GWO | SCA | WOA | HHO | MFO | AVOA | CSA | CPSOGSA | AO | DBO | SO | MISO |
|---|---|---|---|---|---|---|---|---|---|---|---|---|---|
| F1 | Ave | 1.6497E+04 | 1.9941E+04 | 3.3650E+04 | 2.5560E+04 | 5.1574E+04 | 1.9604E+04 | 1.6020E+04 | 3.6105E+04 | 5.9929E+04 | 3.6923E+04 | 2.0209E+04 | **1.4571E+04** |
| | Std | 7.0196E+03 | **4.1634E+03** | 1.3904E+04 | 8.7008E+03 | 2.0601E+04 | 8.4768E+03 | 5.8517E+03 | 1.4196E+04 | 1.4499E+04 | 1.2655E+04 | 6.3101E+03 | 7.9060E+03 |
| F2 | Ave | 5.0812E+02 | 8.2623E+02 | 6.4553E+02 | 5.6775E+02 | 5.4901E+02 | 4.7902E+02 | 5.0319E+02 | 4.8565E+02 | 5.8309E+02 | 5.0983E+02 | 4.6631E+02 | **4.5610E+02** |
| | Std | 4.2019E+01 | 8.4073E+01 | 1.0921E+02 | 4.9054E+01 | 1.2789E+02 | 3.5202E+01 | 3.4759E+01 | 3.5585E+01 | 5.6279E+01 | 6.6872E+01 | 2.3764E+01 | **1.8163E+01** |
| F3 | Ave | 6.0617E+02 | 6.4827E+02 | 6.6799E+02 | 6.6214E+02 | 6.1785E+02 | 6.5136E+02 | 6.2373E+02 | 6.6373E+02 | 6.4748E+02 | 6.3628E+02 | 6.0887E+02 | **6.0501E+02** |
| | Std | **3.6129E+00** | 7.1944E+00 | 1.6204E+01 | 8.3029E+00 | 9.3727E+00 | 1.3250E+01 | 6.7315E+00 | 9.3437E+00 | 1.1085E+01 | 1.1164E+01 | 6.6476E+00 | 4.3346E+00 |
| F4 | Ave | 8.6794E+02 | 9.5619E+02 | 9.4020E+02 | 8.9092E+02 | 8.9809E+02 | 8.8535E+02 | 8.4250E+02 | 9.0641E+02 | 8.8766E+02 | 9.1252E+02 | 8.4285E+02 | **8.3974E+02** |
| | Std | 3.2577E+01 | 1.6083E+01 | 3.6291E+01 | 1.1611E+01 | 2.9866E+01 | 1.8092E+01 | 1.2471E+01 | 2.6428E+01 | 1.8512E+01 | 3.2004E+01 | **9.1091E+00** | 1.2495E+01 |
| F5 | Ave | 1.2428E+03 | 2.7525E+03 | 4.4035E+03 | 2.9407E+03 | 3.0051E+03 | 2.3726E+03 | 1.3030E+03 | 3.5086E+03 | 2.5023E+03 | 2.0548E+03 | 1.2668E+03 | **1.1215E+03** |
| | Std | 2.6726E+02 | 6.6026E+02 | 1.8842E+03 | 4.0681E+02 | 7.9886E+02 | 3.1410E+02 | 2.3934E+02 | 7.7419E+02 | 3.2403E+02 | 6.0139E+02 | 2.0919E+02 | **1.5210E+02** |
| F6 | Ave | 2.9981E+06 | 1.8841E+08 | 6.1271E+06 | 2.4532E+05 | 1.2937E+07 | **5.2052E+03** | 5.3927E+03 | 6.7467E+03 | 4.9451E+05 | 8.0436E+05 | 1.0297E+04 | 7.1859E+03 |
| | Std | 1.1877E+07 | 1.0974E+08 | 4.2067E+06 | 1.3429E+05 | 1.1506E+07 | 4.9616E+03 | 4.9629E+03 | 5.5457E+03 | 3.1265E+05 | 1.7423E+06 | 4.4228E+03 | **4.1099E+03** |
| F7 | Ave | 2.1077E+03 | 2.1514E+03 | 2.2204E+03 | 2.1991E+03 | 2.1243E+03 | 2.1606E+03 | **2.0766E+03** | 2.2133E+03 | 2.1257E+03 | 2.1349E+03 | 2.0843E+03 | 2.0775E+03 |
| | Std | 5.3941E+01 | 2.4486E+01 | 7.7865E+01 | 7.4582E+01 | 5.5416E+01 | 5.0338E+01 | **1.8316E+01** | 8.7914E+01 | 3.1646E+01 | 5.3029E+01 | 3.5101E+01 | 3.0521E+01 |
| F8 | Ave | 2.2549E+03 | 2.2837E+03 | 2.3347E+03 | 2.3017E+03 | 2.2584E+03 | 2.2726E+03 | 2.2440E+03 | 2.4937E+03 | 2.2750E+03 | 2.3167E+03 | 2.2441E+03 | **2.2337E+03** |
| | Std | 4.7611E+01 | 3.3658E+01 | 1.0017E+02 | 8.3987E+01 | 4.7727E+01 | 5.3787E+01 | 3.7248E+01 | 1.5500E+02 | 7.8580E+01 | 7.9516E+01 | 3.5296E+01 | **1.3655E+01** |
| F9 | Ave | 2.5191E+03 | 2.6380E+03 | 2.6175E+03 | 2.5544E+03 | 2.5106E+03 | 2.4903E+03 | 2.5165E+03 | 2.4905E+03 | 2.6095E+03 | 2.5143E+03 | 2.4816E+03 | **2.4809E+03** |
| | Std | 2.4930E+01 | 3.1385E+01 | 5.4468E+01 | 3.8425E+01 | 2.0992E+01 | 8.7209E+00 | 1.9234E+01 | 1.8709E+01 | 4.6250E+01 | 2.5592E+01 | 1.8153E+00 | **3.0246E-01** |
| F10 | Ave | 3.4907E+03 | 3.5988E+03 | 4.6014E+03 | 4.2858E+03 | 3.7683E+03 | 3.4679E+03 | 3.6820E+03 | 4.7473E+03 | 3.2949E+03 | 3.3076E+03 | 3.0360E+03 | **2.7627E+03** |
| | Std | 8.2482E+02 | 1.7375E+03 | 1.3273E+03 | 1.0478E+03 | 1.0473E+03 | 6.9913E+02 | 1.1287E+03 | 7.4605E+02 | 1.2306E+03 | 1.0763E+03 | 3.4173E+02 | **2.6378E+02** |
| F11 | Ave | 3.5587E+03 | 5.2624E+03 | 3.9712E+03 | 3.6012E+03 | 4.2106E+03 | 2.9356E+03 | 3.0041E+03 | 3.0202E+03 | 4.0111E+03 | 3.1335E+03 | 2.9428E+03 | **2.8823E+03** |
| | Std | 2.9181E+02 | 4.6313E+02 | 6.2257E+02 | 7.5256E+02 | 8.4601E+02 | 1.4330E+02 | 2.1409E+02 | 3.1229E+02 | 4.9853E+02 | 2.4973E+02 | **6.9518E+01** | 7.5915E+01 |
| F12 | Ave | 2.9824E+03 | 3.0890E+03 | 3.1095E+03 | 3.2474E+03 | **2.9569E+03** | 2.9944E+03 | 2.9951E+03 | 3.3141E+03 | 3.0613E+03 | 3.0412E+03 | 3.0138E+03 | 2.9918E+03 |
| | Std | 3.8484E+01 | 3.8318E+01 | 1.1861E+02 | 1.5451E+02 | **1.4361E+01** | 4.9967E+01 | 3.3459E+01 | 2.1006E+02 | 5.4147E+01 | 6.0136E+01 | 3.8848E+01 | 2.9217E+01 |
| (W\|T\|L) | | (1\|11\|0) | (0\|7\|5) | (0\|9\|3) | (0\|12\|0) | (1\|11\|0) | (1\|11\|0) | (0\|12\|0) | (0\|9\|3) | (0\|11\|1) | (0\|12\|0) | (0\|12\|0) | (**9**\|3\|0) |

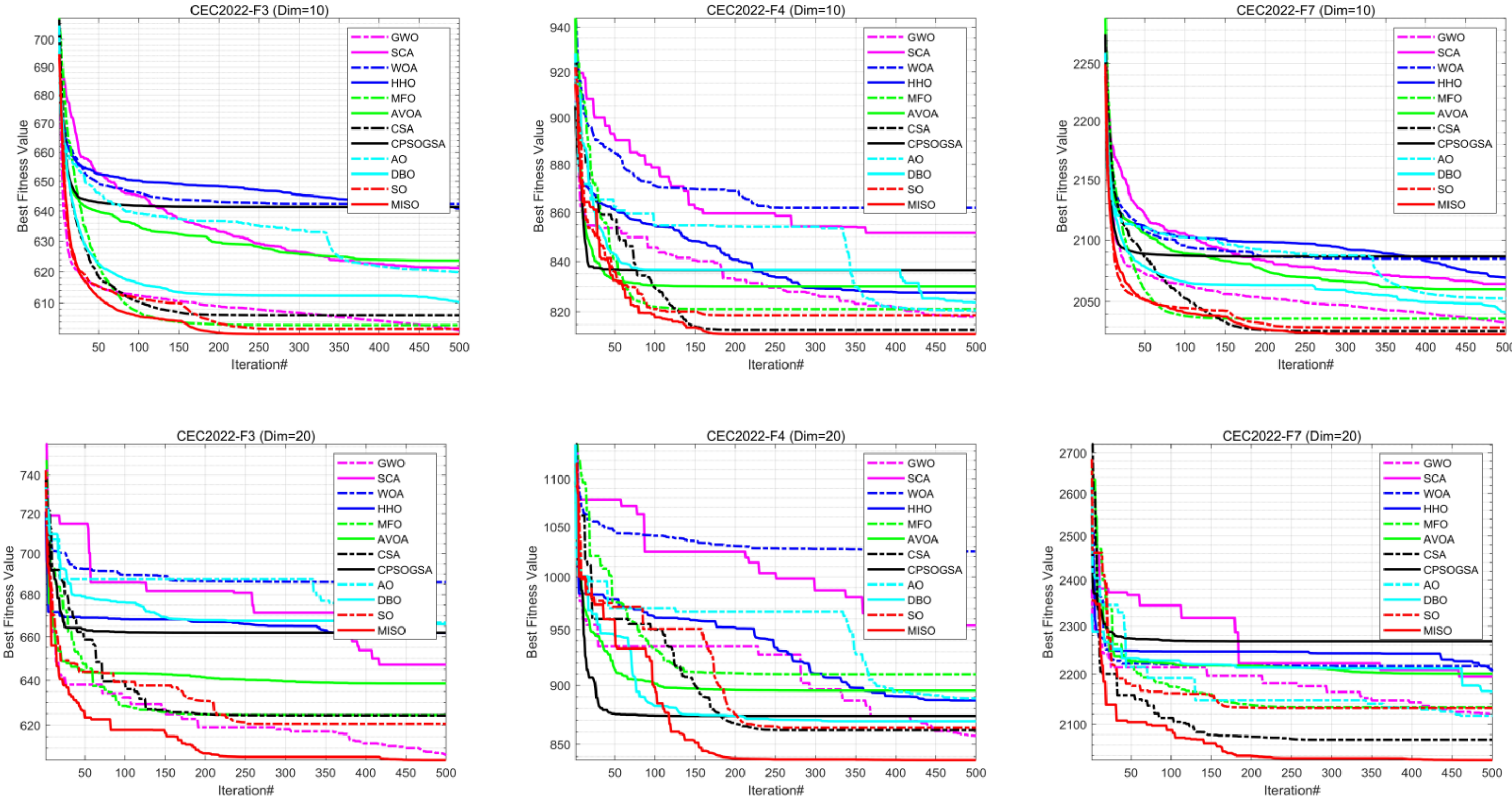

Fig. 10. Comparison of convergence curves of 12 algorithms on CEC2022.

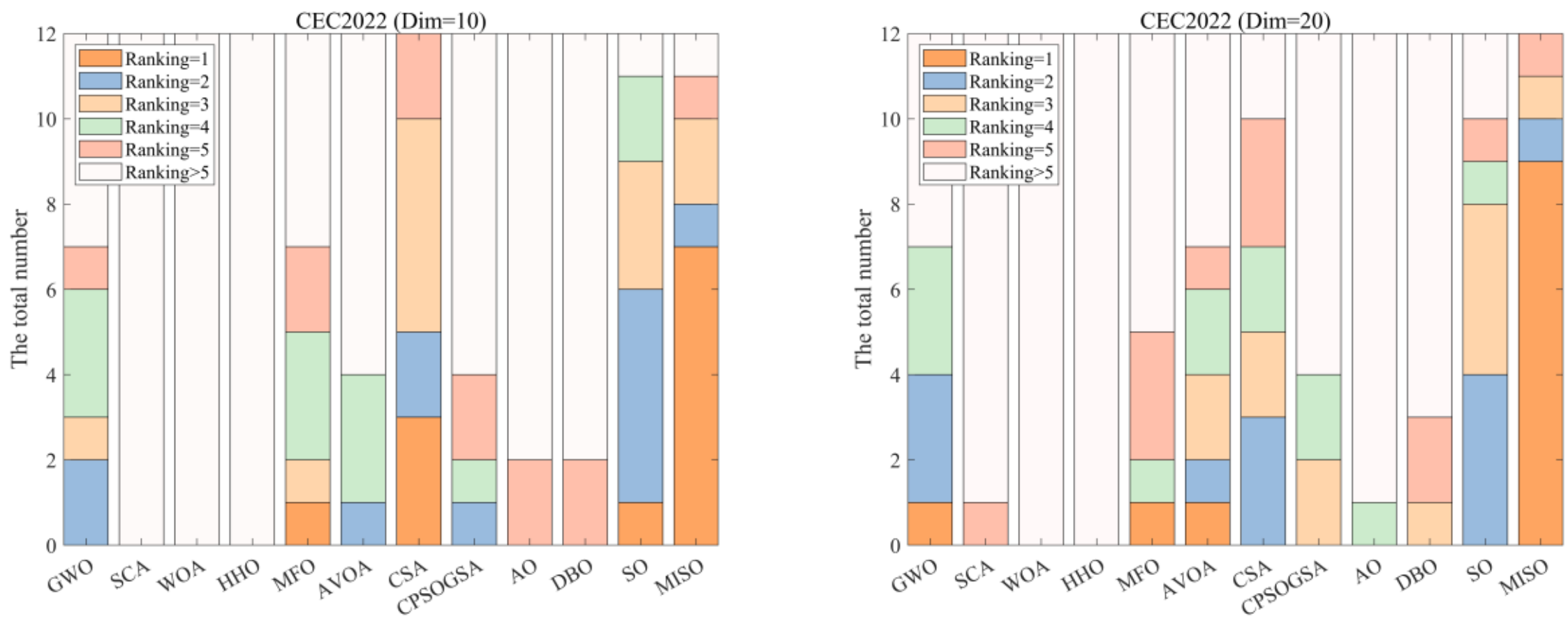

Fig. 11. The ranking stacked bar chart of different competitors on CEC2022.

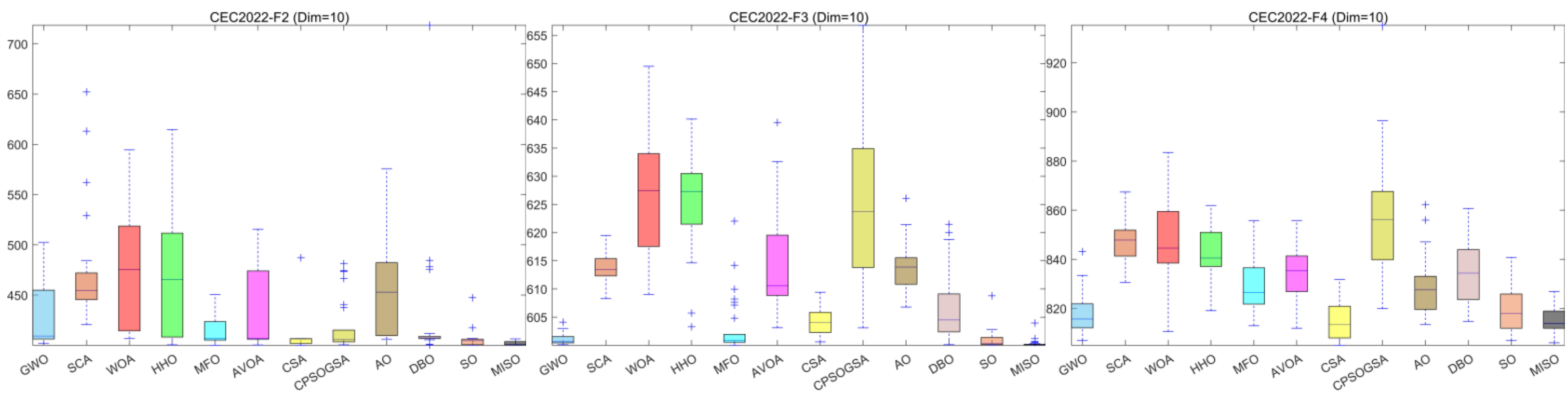

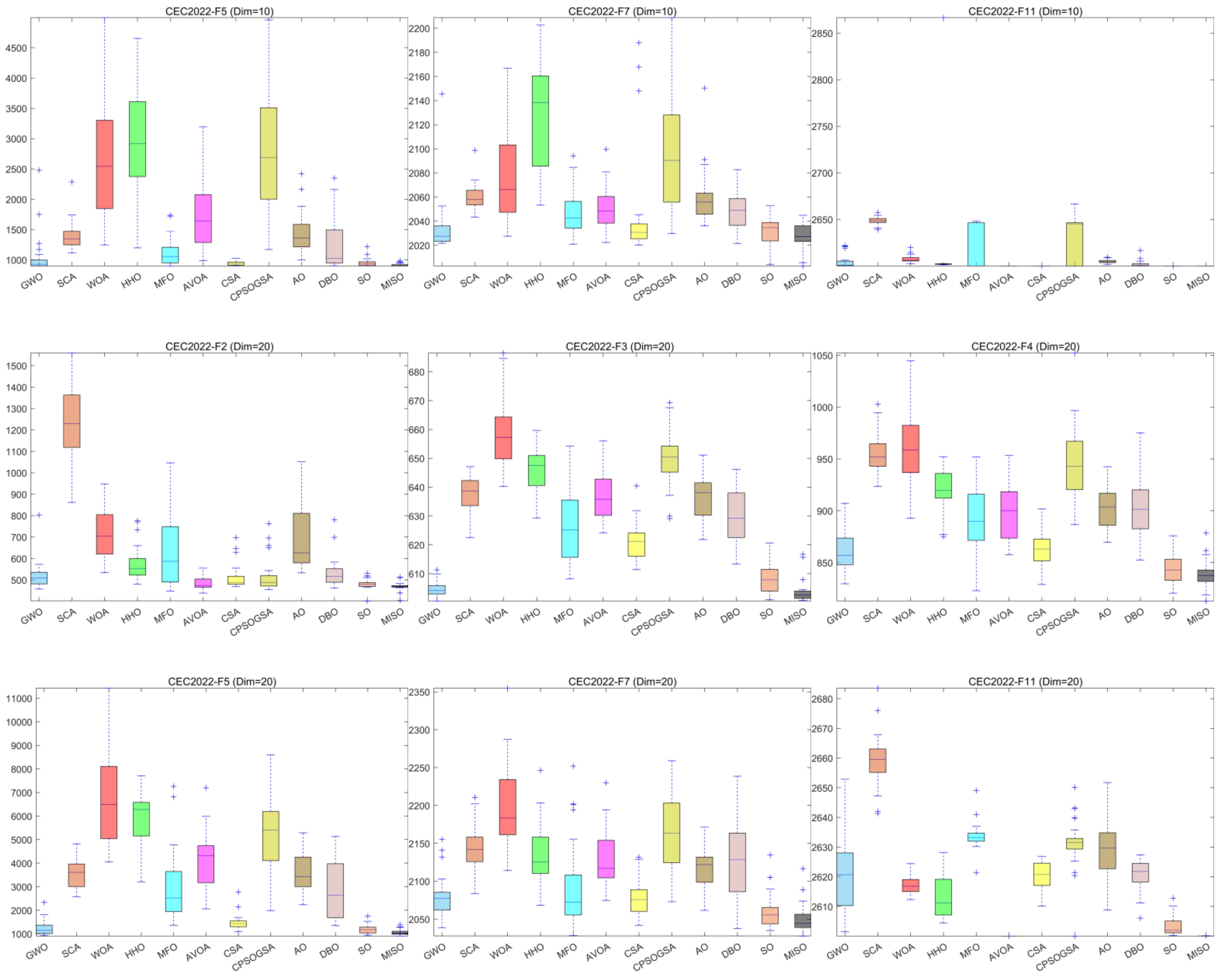

Fig. 12. Boxplot analysis of competitor algorithms on the CEC2022 test suite.

### 4.5 Statistical analysis

In this section, we statistically analyze the experimental results of MISO and other algorithms using the Wilcoxon test and Friedman test to identify any significant differences.

#### 4.5.1 Wilcoxon rank sum test

We conduct nonparametric rank-sum tests [95] to compare MISO with other algorithms, and the results are presented in Table 9, Table 10, Table 11, Table 12 and Table 13. When the p value is less than 0.05, there is a significant difference between the competitive algorithm and MISO. On the contrary, there is no significant difference, and results without a difference are highlighted in bold. We use the “+/=/-” notation to represent whether MISO outperforms, approximates, or underperforms the competition. From the experimental results in Table 14, it is clear that the gap between the MISO algorithm and the other algorithms gradually increases as the problem dimension increases. Synthesizing the analysis in the previous sections, MISO is clearly different from other competitors, and its comprehensive performance is the most prominent.

Table 9

P-value of 12 algorithms on CEC 2017 (Dim=30).

| ID | GWO | SCA | WOA | HHO | MFO | AVOA | CSA | CPSOGSA | AO | DBO | SO |
|---|---|---|---|---|---|---|---|---|---|---|---|
| F1 | 3.0199E-11 | 3.0199E-11 | 3.0199E-11 | 3.0199E-11 | 3.0199E-11 | **5.8945E-01** | 3.0199E-11 | 3.0199E-11 | 3.0199E-11 | 3.0199E-11 | 4.9752E-11 |
| F2 | 3.0199E-11 | 3.0199E-11 | 3.0199E-11 | 3.0199E-11 | 1.0937E-10 | 2.7829E-07 | 3.3384E-11 | 3.0199E-11 | 3.0199E-11 | 3.0199E-11 | 7.1186E-09 |
| F3 | 6.5277E-08 | **2.0095E-01** | 1.0937E-10 | 5.9673E-09 | 4.4440E-07 | 1.0105E-08 | 6.9724E-03 | 9.8329E-08 | 1.9963E-05 | **4.2896E-01** | 1.1143E-03 |
| F4 | 6.7220E-10 | 3.0199E-11 | 3.0199E-11 | 3.6897E-11 | 1.6947E-09 | 4.0330E-03 | 9.8329E-08 | 2.8716E-10 | 3.0199E-11 | 6.0658E-11 | 1.3853E-06 |
| F5 | 3.3242E-06 | 3.0199E-11 | 3.0199E-11 | 3.0199E-11 | 3.0199E-11 | 6.6955E-11 | 4.1997E-10 | 3.0199E-11 | 3.0199E-11 | 3.0199E-11 | 4.0330E-03 |
| F6 | 3.8481E-03 | 3.0199E-11 | 3.0199E-11 | 3.0199E-11 | 3.6897E-11 | 3.0199E-11 | 3.0199E-11 | 3.0199E-11 | 3.0199E-11 | 3.6897E-11 | 5.1857E-07 |
| F7 | 3.1967E-09 | 3.0199E-11 | 3.0199E-11 | 3.0199E-11 | 3.3384E-11 | 3.0199E-11 | 3.0199E-11 | 3.0199E-11 | 3.0199E-11 | 6.6955E-11 | 9.7555E-10 |
| F8 | 1.2212E-02 | 3.0199E-11 | 3.0199E-11 | 5.4941E-11 | 4.5043E-11 | 6.1210E-10 | 1.0576E-03 | 3.6897E-11 | 3.6897E-11 | 3.6897E-11 | **1.1536E-01** |
| F9 | 6.3560E-05 | 3.0199E-11 | 3.0199E-11 | 3.0199E-11 | 3.0199E-11 | 3.0199E-11 | 6.6955E-11 | 3.0199E-11 | 3.0199E-11 | 3.0199E-11 | 4.9426E-05 |
| F10 | 7.0430E-07 | 3.0199E-11 | 3.0199E-11 | 1.9568E-10 | 1.5581E-08 | 1.2023E-08 | 8.8910E-10 | 3.3520E-08 | 8.9934E-11 | 3.1589E-10 | 3.3281E-03 |
| F11 | 1.0937E-10 | 3.0199E-11 | 3.0199E-11 | 3.0939E-06 | 9.9186E-11 | **8.3026E-01** | 2.4386E-09 | 1.3853E-06 | 3.0199E-11 | 9.7555E-10 | 3.9881E-04 |
| F12 | 4.5043E-11 | 3.0199E-11 | 3.0199E-11 | 3.0199E-11 | 4.9980E-09 | 7.7725E-09 | 3.6897E-11 | 5.0723E-10 | 3.0199E-11 | 8.1014E-10 | 2.8913E-03 |
| F13 | 5.0723E-10 | 3.0199E-11 | 3.0199E-11 | 3.0199E-11 | 1.4932E-04 | 1.0277E-06 | 8.6634E-05 | **5.9969E-01** | 3.0199E-11 | 5.4617E-09 | **2.3985E-01** |
| F14 | 7.1988E-05 | 1.4643E-10 | 1.7769E-10 | 1.4643E-10 | 2.8314E-08 | 1.2023E-08 | 9.0688E-03 | 6.5183E-09 | 5.4941E-11 | 7.5991E-07 | 2.6362E-02 |
| F15 | 7.3803E-10 | 3.0199E-11 | 3.0199E-11 | 1.7769E-10 | 5.5611E-04 | 2.5974E-05 | 1.7666E-03 | 4.8413E-02 | 3.6897E-11 | 4.7445E-06 | **8.0727E-01** |
| F16 | **7.4827E-02** | 3.0199E-11 | 3.0199E-11 | 3.6897E-11 | 3.0811E-08 | 6.7220E-10 | 6.3560E-05 | 5.0723E-10 | 4.9752E-11 | 6.1210E-10 | 1.4551E-02 |
| F17 | **3.8710E-01** | 3.3384E-11 | 1.4294E-08 | 5.9673E-09 | 2.5306E-04 | 1.7290E-06 | **2.0095E-01** | 5.4941E-11 | 1.5581E-08 | 2.2273E-09 | **1.0547E-01** |
| F18 | 2.1327E-05 | 3.0199E-11 | 1.5465E-09 | 2.7726E-05 | 1.0407E-04 | 5.3221E-03 | 5.0842E-03 | **8.6499E-01** | 3.1967E-09 | 1.3272E-02 | 1.6955E-02 |
| F19 | 3.1589E-10 | 3.0199E-11 | 3.0199E-11 | 3.6897E-11 | 1.4918E-06 | 5.1857E-07 | 1.2057E-10 | **4.1191E-01** | 3.3384E-11 | 7.1186E-09 | 2.6077E-02 |
| F20 | **4.8252E-01** | 5.5727E-10 | 1.4110E-09 | 1.0105E-08 | 3.0939E-06 | 9.0632E-08 | **8.6499E-01** | 1.8500E-08 | 1.0907E-05 | 3.0939E-06 | **6.2040E-01** |
| F21 | **2.0095E-01** | 3.0199E-11 | 3.0199E-11 | 3.0199E-11 | 3.3384E-11 | 4.5043E-11 | 4.9426E-05 | 3.0199E-11 | 4.0772E-11 | 3.0199E-11 | 8.3146E-03 |
| F22 | 2.2780E-05 | 3.0199E-11 | 5.0723E-10 | 2.4386E-09 | 9.2603E-09 | 1.1143E-03 | 1.0576E-03 | 8.4848E-09 | **8.5338E-01** | 3.3679E-04 | 1.8368E-02 |
| F23 | **4.2896E-01** | 3.0199E-11 | 3.0199E-11 | 3.0199E-11 | 4.9818E-04 | 4.9752E-11 | 1.5846E-04 | 3.0199E-11 | 5.4941E-11 | 4.9752E-11 | **5.0120E-02** |
| F18 | **8.1875E-01** | 3.0199E-11 | 3.0199E-11 | 3.0199E-11 | 1.1674E-05 | 1.2057E-10 | **3.2553E-01** | 3.0199E-11 | 3.0199E-11 | 3.0199E-11 | **9.1171E-01** |
| F19 | 7.3891E-11 | 3.0199E-11 | 3.0199E-11 | 4.0772E-11 | 4.6159E-10 | 1.9963E-05 | 3.0199E-11 | 1.4643E-10 | 3.0199E-11 | 5.0723E-10 | 5.2640E-04 |
| F20 | 1.6955E-02 | 3.0199E-11 | 3.0199E-11 | 3.5201E-07 | 1.2493E-05 | 1.2870E-09 | 3.0317E-02 | 6.1210E-10 | 2.2780E-05 | 1.4733E-07 | 4.6756E-02 |
| F21 | **8.5000E-02** | 3.0199E-11 | 6.7220E-10 | 3.0199E-11 | 3.8307E-05 | 2.3243E-02 | 2.4913E-06 | 3.0199E-11 | 3.0199E-11 | 3.7704E-04 | 4.4272E-03 |
| F22 | 3.4742E-10 | 3.0199E-11 | 3.0199E-11 | 8.9934E-11 | 4.5043E-11 | 3.6709E-03 | 1.3111E-08 | 4.7445E-06 | 3.0199E-11 | 1.0105E-08 | 2.0023E-06 |
| F23 | **5.8945E-01** | 3.3384E-11 | 5.4941E-11 | 2.3715E-10 | 5.8282E-03 | 3.0103E-07 | 4.0840E-05 | 2.3897E-08 | 2.6099E-10 | 2.4913E-06 | 3.5137E-02 |
| F24 | 4.0772E-11 | 3.0199E-11 | 3.0199E-11 | 3.0199E-11 | 1.0763E-02 | 1.1737E-09 | 4.0772E-11 | 2.3168E-06 | 3.0199E-11 | 5.4617E-09 | **7.0617E-01** |
| F25 | 3.0199E-11 | 3.0199E-11 | 3.0199E-11 | 3.0199E-11 | 3.0199E-11 | **5.8945E-01** | 3.0199E-11 | 3.0199E-11 | 3.0199E-11 | 3.0199E-11 | 4.9752E-11 |
| F26 | 3.0199E-11 | 3.0199E-11 | 3.0199E-11 | 3.0199E-11 | 1.0937E-10 | 2.7829E-07 | 3.3384E-11 | 3.0199E-11 | 3.0199E-11 | 3.0199E-11 | 7.1186E-09 |
| F27 | 6.5277E-08 | **2.0095E-01** | 1.0937E-10 | 5.9673E-09 | 4.4440E-07 | 1.0105E-08 | 6.9724E-03 | 9.8329E-08 | 1.9963E-05 | **4.2896E-01** | 1.1143E-03 |
| F28 | 6.7220E-10 | 3.0199E-11 | 3.0199E-11 | 3.6897E-11 | 1.6947E-09 | 4.0330E-03 | 9.8329E-08 | 2.8716E-10 | 3.0199E-11 | 6.0658E-11 | 1.3853E-06 |
| F29 | 3.3242E-06 | 3.0199E-11 | 3.0199E-11 | 3.0199E-11 | 3.0199E-11 | 6.6955E-11 | 4.1997E-10 | 3.0199E-11 | 3.0199E-11 | 3.0199E-11 | 4.0330E-03 |
| F30 | 3.8481E-03 | 3.0199E-11 | 3.0199E-11 | 3.0199E-11 | 3.6897E-11 | 3.0199E-11 | 3.0199E-11 | 3.0199E-11 | 3.0199E-11 | 3.6897E-11 | 5.1857E-07 |

Table 10

P-value of 12 algorithms on CEC 2017 (Dim=50).

| ID | GWO | SCA | WOA | HHO | MFO | AVOA | CSA | CPSOGSA | AO | DBO | SO |
|---|---|---|---|---|---|---|---|---|---|---|---|
| F1 | 3.0199E-11 | 3.0199E-11 | 3.0199E-11 | 3.0199E-11 | 3.0199E-11 | 1.7769E-10 | 3.0199E-11 | 3.0199E-11 | 3.0199E-11 | 3.0199E-11 | 3.0199E-11 |
| F2 | 1.7479E-05 | 3.0199E-11 | 3.0199E-11 | 3.0199E-11 | 4.6159E-10 | 2.3800E-03 | 8.1527E-11 | 3.0199E-11 | 3.0199E-11 | 4.9752E-11 | 2.4386E-09 |
| F3 | 1.4733E-07 | 3.2651E-02 | **1.8090E-01** | 5.5329E-08 | 4.6159E-10 | 1.4918E-06 | 9.5139E-06 | 4.5726E-09 | 3.3679E-04 | **4.0354E-01** | 2.4386E-09 |
| F4 | 3.0199E-11 | 3.0199E-11 | 3.0199E-11 | 3.0199E-11 | 3.0199E-11 | 1.5964E-07 | 3.0199E-11 | 3.0199E-11 | 3.0199E-11 | 3.3384E-11 | 7.3891E-11 |
| F5 | 8.3520E-08 | 3.0199E-11 | 3.0199E-11 | 3.0199E-11 | 3.0199E-11 | 3.0199E-11 | 3.0199E-11 | 3.0199E-11 | 3.0199E-11 | 3.0199E-11 | 1.0188E-05 |
| F6 | 2.9590E-05 | 3.0199E-11 | 3.0199E-11 | 3.0199E-11 | 3.0199E-11 | 3.0199E-11 | 3.0199E-11 | 3.0199E-11 | 3.0199E-11 | 3.0199E-11 | 1.0666E-07 |
| F7 | 1.9568E-10 | 3.0199E-11 | 3.0199E-11 | 3.0199E-11 | 3.0199E-11 | 3.0199E-11 | 3.0199E-11 | 3.0199E-11 | 3.0199E-11 | 3.0199E-11 | 3.0199E-11 |
| F8 | 7.7725E-09 | 3.0199E-11 | 3.0199E-11 | 3.0199E-11 | 3.0199E-11 | 3.0199E-11 | 1.3289E-10 | 3.0199E-11 | 3.0199E-11 | 3.0199E-11 | 2.8913E-03 |
| F9 | 3.0199E-11 | 3.0199E-11 | 3.0199E-11 | 3.0199E-11 | 3.0199E-11 | 3.0199E-11 | 3.0199E-11 | 3.0199E-11 | 3.0199E-11 | 3.0199E-11 | 2.1544E-10 |
| F10 | 2.2539E-04 | 3.0199E-11 | 3.0199E-11 | 5.4941E-11 | 2.7726E-05 | 6.0459E-07 | 4.0772E-11 | 2.6784E-06 | 3.3384E-11 | 4.5043E-11 | 1.8682E-05 |
| F11 | 3.0199E-11 | 3.0199E-11 | 3.0199E-11 | 6.6955E-11 | 3.0199E-11 | **5.8945E-01** | 3.0199E-11 | 3.0199E-11 | 3.0199E-11 | 6.6955E-11 | 1.0937E-10 |
| F12 | 3.0199E-11 | 3.0199E-11 | 3.0199E-11 | 3.0199E-11 | 3.6897E-11 | 6.0459E-07 | 4.0772E-11 | 3.0199E-11 | 3.0199E-11 | 3.0199E-11 | 7.6973E-04 |
| F13 | 3.0199E-11 | 3.0199E-11 | 3.0199E-11 | 3.0199E-11 | 1.2870E-09 | 2.2360E-02 | 1.1747E-04 | 8.8829E-06 | 3.0199E-11 | 1.4643E-10 | 2.2539E-04 |
| F14 | 2.0023E-06 | 3.0199E-11 | 3.6897E-11 | 3.0199E-11 | 7.2208E-06 | 5.4617E-09 | **9.3519E-01** | 1.3853E-06 | 3.0199E-11 | 7.7725E-09 | **1.9073E-01** |
| F15 | 1.3289E-10 | 3.0199E-11 | 3.0199E-11 | 3.0199E-11 | 8.1975E-07 | 2.7829E-07 | 6.5486E-04 | 1.1567E-07 | 3.0199E-11 | 7.7725E-09 | **4.2039E-01** |
| F16 | **9.9258E-02** | 3.0199E-11 | 3.0199E-11 | 3.0199E-11 | 4.9752E-11 | 2.3715E-10 | 6.0104E-08 | 9.2603E-09 | 1.0937E-10 | 5.4941E-11 | 1.9883E-02 |
| F17 | **7.5059E-01** | 3.0199E-11 | 4.9752E-11 | 4.3106E-08 | 6.1210E-10 | 2.1959E-07 | 2.3885E-04 | 1.4110E-09 | 7.1186E-09 | 2.1544E-10 | **2.0621E-01** |
| F18 | 5.2650E-05 | 3.0199E-11 | 3.0199E-11 | 9.8329E-08 | 1.4067E-04 | 7.9590E-03 | **3.3285E-01** | **5.3951E-01** | 1.0702E-09 | 2.7829E-07 | 4.0330E-03 |
| F19 | 3.6897E-11 | 3.0199E-11 | 3.0199E-11 | 3.0199E-11 | 1.7294E-07 | 2.6099E-10 | 4.0772E-11 | 7.5991E-07 | 3.0199E-11 | 4.9980E-09 | **5.0120E-02** |
| F20 | **9.3519E-01** | 3.0199E-11 | 9.9186E-11 | 6.5277E-08 | 6.0104E-08 | 5.0922E-08 | **6.1001E-01** | 3.0811E-08 | 6.0459E-07 | 1.4643E-10 | **2.2257E-01** |
| F21 | 3.8349E-06 | 3.0199E-11 | 3.0199E-11 | 3.0199E-11 | 3.0199E-11 | 3.0199E-11 | 4.9752E-11 | 3.0199E-11 | 3.0199E-11 | 3.0199E-11 | 2.5101E-02 |
| F22 | 8.6844E-03 | 3.0199E-11 | 3.0199E-11 | 5.4941E-11 | 3.0059E-04 | 8.8411E-07 | 3.1589E-10 | 1.2493E-05 | 5.4941E-11 | 1.6947E-09 | 6.0971E-03 |
| F23 | **1.3345E-01** | 3.0199E-11 | 3.0199E-11 | 3.0199E-11 | 1.4294E-08 | 3.0199E-11 | 2.8716E-10 | 3.0199E-11 | 3.0199E-11 | 3.0199E-11 | 2.0523E-03 |
| F18 | **8.5338E-01** | 3.0199E-11 | 3.0199E-11 | 3.0199E-11 | 1.1738E-03 | 4.9752E-11 | 6.0104E-08 | 3.0199E-11 | 3.0199E-11 | 3.0199E-11 | **3.4029E-01** |
| F19 | 3.0199E-11 | 3.0199E-11 | 3.0199E-11 | 3.0199E-11 | 3.0199E-11 | 2.6099E-10 | 3.0199E-11 | 3.0199E-11 | 3.0199E-11 | 5.4941E-11 | 4.0772E-11 |
| F20 | 1.6955E-02 | 3.0199E-11 | 3.0199E-11 | 8.8910E-10 | 7.0881E-08 | 1.5465E-09 | 5.0723E-10 | 3.0199E-11 | 1.6132E-10 | 5.5727E-10 | 2.5101E-02 |
| F21 | 2.0681E-02 | 3.0199E-11 | 6.0658E-11 | 3.0199E-11 | **4.9178E-01** | 7.0430E-07 | 1.2860E-06 | 3.0199E-11 | 3.0199E-11 | 3.8053E-07 | 3.2555E-07 |
| F22 | 1.3289E-10 | 3.0199E-11 | 3.0199E-11 | 8.1527E-11 | 3.3384E-11 | 8.1465E-05 | 1.9568E-10 | 3.8249E-09 | 3.0199E-11 | 8.1014E-10 | 7.6950E-08 |
| F23 | 4.3584E-02 | 3.0199E-11 | 3.0199E-11 | 3.0199E-11 | 5.0912E-06 | 4.9980E-09 | 8.1527E-11 | 6.0658E-11 | 3.0199E-11 | 1.4110E-09 | 1.3250E-04 |
| F24 | 3.0199E-11 | 3.0199E-11 | 3.0199E-11 | 3.0199E-11 | 2.2658E-03 | 4.5043E-11 | 3.0199E-11 | 3.0199E-11 | 3.0199E-11 | 8.9934E-11 | 2.3897E-08 |
| F25 | 3.0199E-11 | 3.0199E-11 | 3.0199E-11 | 3.0199E-11 | 3.0199E-11 | 1.7769E-10 | 3.0199E-11 | 3.0199E-11 | 3.0199E-11 | 3.0199E-11 | 3.0199E-11 |
| F26 | 1.7479E-05 | 3.0199E-11 | 3.0199E-11 | 3.0199E-11 | 4.6159E-10 | 2.3800E-03 | 8.1527E-11 | 3.0199E-11 | 3.0199E-11 | 4.9752E-11 | 2.4386E-09 |
| F27 | 1.4733E-07 | 3.2651E-02 | **1.8090E-01** | 5.5329E-08 | 4.6159E-10 | 1.4918E-06 | 9.5139E-06 | 4.5726E-09 | 3.3679E-04 | **4.0354E-01** | 2.4386E-09 |
| F28 | 3.0199E-11 | 3.0199E-11 | 3.0199E-11 | 3.0199E-11 | 3.0199E-11 | 1.5964E-07 | 3.0199E-11 | 3.0199E-11 | 3.0199E-11 | 3.3384E-11 | 7.3891E-11 |
| F29 | 8.3520E-08 | 3.0199E-11 | 3.0199E-11 | 3.0199E-11 | 3.0199E-11 | 3.0199E-11 | 3.0199E-11 | 3.0199E-11 | 3.0199E-11 | 3.0199E-11 | 1.0188E-05 |
| F30 | 2.9590E-05 | 3.0199E-11 | 3.0199E-11 | 3.0199E-11 | 3.0199E-11 | 3.0199E-11 | 3.0199E-11 | 3.0199E-11 | 3.0199E-11 | 3.0199E-11 | 1.0666E-07 |

Table 11

P-value of 12 algorithms on CEC 2017 (Dim=100).

| ID | GWO | SCA | WOA | HHO | MFO | AVOA | CSA | CPSOGSA | AO | DBO | SO |
|---|---|---|---|---|---|---|---|---|---|---|---|
| F1 | 3.0199E-11 | 3.0199E-11 | 3.0199E-11 | 3.0199E-11 | 3.0199E-11 | 3.0199E-11 | 3.0199E-11 | 3.0199E-11 | 3.0199E-11 | 3.0199E-11 | 3.0199E-11 |
| F2 | 6.9125E-04 | 3.0199E-11 | 3.0199E-11 | 3.0199E-11 | 3.0199E-11 | 3.1589E-10 | 3.6897E-11 | 3.0199E-11 | 3.0199E-11 | 3.0199E-11 | 2.0338E-09 |
| F3 | 3.5708E-06 | **6.7869E-02** | 3.0811E-08 | 9.7555E-10 | 2.9215E-09 | 2.2273E-09 | 1.6947E-09 | 1.8682E-05 | 3.0199E-11 | **3.8710E-01** | 3.0199E-11 |
| F4 | 3.0199E-11 | 3.0199E-11 | 3.0199E-11 | 3.0199E-11 | 3.0199E-11 | 3.0199E-11 | 3.0199E-11 | 3.0199E-11 | 3.0199E-11 | 3.0199E-11 | 3.0199E-11 |
| F5 | 1.2057E-10 | 3.0199E-11 | 3.0199E-11 | 3.0199E-11 | 3.0199E-11 | 3.0199E-11 | 3.0199E-11 | 3.0199E-11 | 3.0199E-11 | 3.0199E-11 | 3.1967E-09 |
| F6 | 2.4386E-09 | 3.0199E-11 | 3.0199E-11 | 3.0199E-11 | 3.0199E-11 | 3.0199E-11 | 3.0199E-11 | 3.0199E-11 | 3.0199E-11 | 3.0199E-11 | 2.1544E-10 |
| F7 | 3.0199E-11 | 3.0199E-11 | 3.0199E-11 | 3.0199E-11 | 3.0199E-11 | 3.0199E-11 | 3.0199E-11 | 3.0199E-11 | 3.0199E-11 | 3.0199E-11 | 3.0199E-11 |
| F8 | 2.3897E-08 | 3.0199E-11 | 3.0199E-11 | 3.0199E-11 | 3.0199E-11 | 3.0199E-11 | 3.0199E-11 | 3.0199E-11 | 3.0199E-11 | 3.0199E-11 | 1.6813E-04 |
| F9 | 3.0199E-11 | 3.0199E-11 | 3.0199E-11 | 3.0199E-11 | 3.0199E-11 | 3.0199E-11 | 3.0199E-11 | 3.0199E-11 | 3.0199E-11 | 3.0199E-11 | 5.4941E-11 |
| F10 | **5.4933E-01** | 3.0199E-11 | 3.0199E-11 | 4.0772E-11 | 5.0912E-06 | **2.3985E-01** | 4.0772E-11 | 2.9205E-02 | 4.9752E-11 | 1.3289E-10 | 3.0199E-11 |
| F11 | 2.5974E-05 | 1.8731E-07 | 5.4941E-11 | 3.0317E-02 | 1.2023E-08 | 4.8413E-02 | **4.8252E-01** | 1.1023E-08 | 3.3384E-11 | 7.1186E-09 | 9.0688E-03 |
| F12 | 3.0199E-11 | 3.0199E-11 | 3.0199E-11 | 3.0199E-11 | 3.0199E-11 | 4.9752E-11 | 3.0199E-11 | 3.0199E-11 | 3.0199E-11 | 3.0199E-11 | 3.3384E-11 |
| F13 | 3.0199E-11 | 3.0199E-11 | 3.0199E-11 | 3.0199E-11 | 3.0199E-11 | 4.4272E-03 | 3.0199E-11 | 3.0199E-11 | 3.0199E-11 | 3.0199E-11 | 2.0338E-09 |
| F14 | 5.4620E-06 | 3.0199E-11 | 1.2057E-10 | 3.8349E-06 | 1.6980E-08 | 1.2477E-04 | **9.0490E-02** | **6.5671E-02** | 2.8716E-10 | 3.9648E-08 | 3.0339E-03 |
| F15 | 3.0199E-11 | 3.0199E-11 | 3.0199E-11 | 3.0199E-11 | 1.7769E-10 | **8.7710E-02** | 2.3168E-06 | 1.5846E-04 | 3.0199E-11 | 3.0199E-11 | 2.0023E-06 |
| F16 | 8.6844E-03 | 3.0199E-11 | 3.0199E-11 | 3.0199E-11 | 2.1544E-10 | 1.3111E-08 | 4.9752E-11 | 2.8716E-10 | 3.0199E-11 | 3.0199E-11 | 2.3885E-04 |
| F17 | 8.5641E-04 | 3.0199E-11 | 3.0199E-11 | 3.3384E-11 | 3.0199E-11 | 7.3803E-10 | 9.7555E-10 | 6.0658E-11 | 3.0199E-11 | 3.0199E-11 | 4.6390E-05 |
| F18 | 4.8560E-03 | 3.0199E-11 | 1.2870E-09 | 2.3168E-06 | 3.4971E-09 | **5.4933E-01** | 1.0576E-03 | 2.1506E-02 | 6.7220E-10 | 2.6099E-10 | 6.5261E-07 |
| F19 | 3.0199E-11 | 3.0199E-11 | 3.0199E-11 | 3.0199E-11 | 3.0199E-11 | 1.0666E-07 | 7.3891E-11 | 4.9752E-11 | 3.0199E-11 | 4.9752E-11 | 1.4298E-05 |
| F20 | **9.2344E-01** | 3.0199E-11 | 4.0772E-11 | 7.5991E-07 | 1.5969E-03 | 5.2640E-04 | **5.7929E-01** | 1.6285E-02 | 2.3768E-07 | 2.6695E-09 | 4.9752E-11 |
| F21 | 1.1077E-06 | 3.0199E-11 | 3.0199E-11 | 3.0199E-11 | 3.0199E-11 | 3.0199E-11 | 3.0199E-11 | 3.0199E-11 | 3.0199E-11 | 3.0199E-11 | 2.3897E-08 |
| F22 | 2.8913E-03 | 3.0199E-11 | 3.0199E-11 | 1.3289E-10 | 1.6351E-05 | 8.1200E-04 | 8.9934E-11 | **1.7613E-01** | 8.9934E-11 | 3.8202E-10 | 3.0199E-11 |
| F23 | 5.8737E-04 | 3.0199E-11 | 3.0199E-11 | 3.0199E-11 | 1.0937E-10 | 3.0199E-11 | 3.0199E-11 | 3.0199E-11 | 3.0199E-11 | 3.0199E-11 | 3.5923E-05 |
| F18 | **7.2446E-02** | 3.0199E-11 | 3.0199E-11 | 3.0199E-11 | 2.9205E-02 | 3.0199E-11 | 3.0199E-11 | 3.0199E-11 | 3.0199E-11 | 3.0199E-11 | 1.1567E-07 |
| F19 | 3.0199E-11 | 3.0199E-11 | 3.0199E-11 | 3.0199E-11 | 3.0199E-11 | 2.9215E-09 | 3.0199E-11 | 3.0199E-11 | 3.0199E-11 | 3.0199E-11 | 3.0199E-11 |
| F20 | **9.3341E-02** | 3.0199E-11 | 3.0199E-11 | 3.0199E-11 | 1.5964E-07 | 4.6159E-10 | 3.3384E-11 | 3.0199E-11 | 3.0199E-11 | 3.3384E-11 | 1.4733E-07 |
| F21 | 5.5329E-08 | 3.0199E-11 | 3.0199E-11 | 3.0199E-11 | **1.0000E+00** | 1.8731E-07 | 2.8716E-10 | 3.0199E-11 | 3.0199E-11 | 4.3106E-08 | 9.5332E-07 |
| F22 | 1.1937E-06 | 3.0199E-11 | 3.0199E-11 | 3.0811E-08 | 3.0199E-11 | 2.1265E-04 | 2.4386E-09 | 2.1540E-06 | 3.0199E-11 | 3.9648E-08 | 2.6015E-08 |
| F23 | 5.5999E-07 | 3.0199E-11 | 3.0199E-11 | 3.0199E-11 | 1.5465E-09 | 4.6856E-08 | 3.0199E-11 | 3.0199E-11 | 3.0199E-11 | 1.6132E-10 | 8.1200E-04 |
| F24 | 3.0199E-11 | 3.0199E-11 | 3.0199E-11 | 3.0199E-11 | 3.0199E-11 | 3.0199E-11 | 3.0199E-11 | 3.0199E-11 | 3.0199E-11 | 3.0199E-11 | 1.5465E-09 |
| F25 | 3.0199E-11 | 3.0199E-11 | 3.0199E-11 | 3.0199E-11 | 3.0199E-11 | 3.0199E-11 | 3.0199E-11 | 3.0199E-11 | 3.0199E-11 | 3.0199E-11 | 3.0199E-11 |
| F26 | 6.9125E-04 | 3.0199E-11 | 3.0199E-11 | 3.0199E-11 | 3.0199E-11 | 3.1589E-10 | 3.6897E-11 | 3.0199E-11 | 3.0199E-11 | 3.0199E-11 | 2.0338E-09 |
| F27 | 3.5708E-06 | **6.7869E-02** | 3.0811E-08 | 9.7555E-10 | 2.9215E-09 | 2.2273E-09 | 1.6947E-09 | 1.8682E-05 | 3.0199E-11 | **3.8710E-01** | 3.0199E-11 |
| F28 | 3.0199E-11 | 3.0199E-11 | 3.0199E-11 | 3.0199E-11 | 3.0199E-11 | 3.0199E-11 | 3.0199E-11 | 3.0199E-11 | 3.0199E-11 | 3.0199E-11 | 3.0199E-11 |
| F29 | 1.2057E-10 | 3.0199E-11 | 3.0199E-11 | 3.0199E-11 | 3.0199E-11 | 3.0199E-11 | 3.0199E-11 | 3.0199E-11 | 3.0199E-11 | 3.0199E-11 | 3.1967E-09 |
| F30 | 2.4386E-09 | 3.0199E-11 | 3.0199E-11 | 3.0199E-11 | 3.0199E-11 | 3.0199E-11 | 3.0199E-11 | 3.0199E-11 | 3.0199E-11 | 3.0199E-11 | 2.1544E-10 |

Table 12

P-value of 12 algorithms on CEC 2022(Dim=10).

| ID | GWO | SCA | WOA | HHO | MFO | AVOA | CSA | CPSOGSA | AO | DBO | SO |
|---|---|---|---|---|---|---|---|---|---|---|---|
| F1 | 1.1023E-08 | 3.6897E-11 | 3.0199E-11 | 1.0702E-09 | 3.6459E-08 | **6.2040E-01** | **5.3685E-02** | 3.5708E-06 | 3.0199E-11 | 1.4298E-05 | 7.6588E-05 |
| F2 | 1.4110E-09 | 3.0199E-11 | 3.0199E-11 | 2.5721E-07 | 3.2409E-08 | **4.5530E-01** | **6.7350E-01** | 2.1536E-03 | 1.2057E-10 | 2.1532E-10 | **5.6922E-01** |
| F3 | 2.4913E-06 | 3.0199E-11 | 3.0199E-11 | 3.0199E-11 | 6.5486E-04 | 3.0199E-11 | 3.0199E-11 | 3.0199E-11 | 3.0199E-11 | 9.9186E-11 | 1.8916E-04 |
| F4 | **6.8432E-01** | 3.0199E-11 | 2.8716E-10 | 7.6950E-08 | 7.5991E-07 | 6.5277E-08 | 1.7666E-03 | 6.0658E-11 | 5.6073E-05 | 1.4727E-07 | **8.7663E-01** |
| F5 | 1.2212E-02 | 1.9568E-10 | 4.0772E-11 | 3.0199E-11 | 1.3367E-05 | 6.6955E-11 | 4.2175E-04 | 3.3384E-11 | 5.5727E-10 | 7.1186E-09 | 6.7362E-06 |
| F6 | 4.4592E-04 | 3.0199E-11 | 7.2884E-03 | 2.8913E-03 | **8.4180E-01** | **5.4933E-01** | 4.6390E-05 | **3.8710E-01** | 3.0199E-11 | **2.1156E-01** | 4.0354E-02 |
| F7 | 1.5014E-02 | 4.6159E-10 | 6.0658E-11 | 2.6695E-09 | 4.4272E-03 | 1.6980E-08 | 5.5699E-03 | 8.1014E-10 | 1.8500E-08 | 2.1327E-05 | **1.6238E-01** |
| F8 | 1.6351E-05 | 3.0199E-11 | 4.5043E-11 | 6.6955E-11 | 6.0971E-03 | 9.5139E-06 | **7.4827E-02** | 5.4941E-11 | 3.4742E-10 | 7.5991E-07 | **9.2344E-01** |
| F9 | 3.0123E-11 | 3.0123E-11 | 3.0123E-11 | 3.0123E-11 | **3.1586E-01** | 7.7578E-09 | 4.0671E-11 | 3.3320E-03 | 3.0123E-11 | 7.8954E-06 | 3.1265E-02 |
| F10 | **3.1119E-01** | 3.5923E-05 | 8.8829E-06 | 1.0277E-06 | 4.2175E-04 | 1.3853E-06 | 1.1711E-02 | 3.5012E-03 | 1.3367E-05 | 3.7704E-04 | 4.5000E-02 |
| F11 | 2.6243E-03 | 1.7836E-04 | 3.5201E-07 | 1.2493E-05 | **1.2597E-01** | **3.8710E-01** | 1.1143E-03 | **8.4180E-01** | 9.4683E-03 | 3.6439E-02 | **3.7904E-01** |
| F12 | 1.4423E-03 | 1.7836E-04 | 2.4327E-05 | 7.7725E-09 | 1.6956E-08 | 2.5101E-02 | 1.1567E-07 | 2.3885E-04 | **2.6433E-01** | 3.3874E-02 | 6.3772E-03 |

Table 13

P-value of 12 algorithms on CEC 2022(Dim=20).

| ID | GWO | SCA | WOA | HHO | MFO | AVOA | CSA | CPSOGSA | AO | DBO | SO |
|---|---|---|---|---|---|---|---|---|---|---|---|
| F1 | **2.5188E-01** | 5.5611E-04 | 6.0104E-08 | 5.8587E-06 | 3.4742E-10 | 5.8282E-03 | 3.8781E-03 | 1.4294E-08 | 3.0199E-11 | 1.2023E-08 | 3.0339E-03 |
| F2 | 2.8314E-08 | 3.0199E-11 | 1.9568E-10 | 4.5043E-11 | 5.4620E-06 | 5.2640E-04 | 3.3520E-08 | 8.8829E-06 | 4.5043E-11 | 2.3168E-06 | 6.0971E-03 |
| F3 | **5.9428E-02** | 3.0199E-11 | 3.0199E-11 | 3.0199E-11 | 4.5726E-09 | 3.0199E-11 | 1.0937E-10 | 3.0199E-11 | 3.0199E-11 | 3.3384E-11 | 2.2658E-03 |
| F4 | 5.6073E-05 | 3.0199E-11 | 3.0199E-11 | 3.0199E-11 | 5.9673E-09 | 7.3891E-11 | **3.8710E-01** | 4.0772E-11 | 4.0772E-11 | 5.4941E-11 | **2.3399E-01** |
| F5 | **1.2235E-01** | 3.0199E-11 | 3.0199E-11 | 3.0199E-11 | 5.4941E-11 | 3.0199E-11 | 7.2951E-04 | 3.0199E-11 | 3.0199E-11 | 2.6695E-09 | 2.7548E-03 |
| F6 | 1.7294E-07 | 3.0199E-11 | 3.0199E-11 | 3.0199E-11 | 2.5101E-02 | **2.0095E-01** | **4.7335E-01** | **7.3940E-01** | 3.0199E-11 | 4.2175E-04 | 4.2067E-02 |
| F7 | 2.8129E-02 | 1.0702E-09 | 9.9186E-11 | 1.4643E-10 | 3.1821E-04 | 5.4617E-09 | 5.3121E-03 | 1.3289E-10 | 8.8411E-07 | 1.7479E-05 | **7.9585E-01** |
| F8 | 1.8368E-02 | 2.1544E-10 | 7.7725E-09 | 3.0103E-07 | 1.1228E-02 | 3.5923E-05 | **9.3341E-02** | 9.9186E-11 | 1.4298E-05 | 6.7362E-06 | **2.8378E-01** |
| F9 | 3.0199E-11 | 3.0199E-11 | 3.0199E-11 | 3.0199E-11 | 1.5465E-09 | 4.1997E-10 | 3.0199E-11 | 2.6099E-10 | 3.0199E-11 | 4.0772E-11 | 3.9881E-04 |
| F10 | 2.1566E-03 | **4.1191E-01** | 2.7726E-05 | 5.5999E-07 | 9.5207E-04 | 1.4932E-04 | 1.1711E-02 | 1.4110E-09 | **8.5338E-01** | **3.0418E-01** | 6.9125E-04 |
| F11 | 3.0199E-11 | 3.0199E-11 | 3.0199E-11 | 3.0199E-11 | 1.9568E-10 | **2.5805E-01** | 6.5261E-07 | 4.2064E-02 | 3.0199E-11 | 9.8329E-08 | 3.1967E-09 |
| F12 | **5.1877E-02** | 4.1997E-10 | 1.6062E-06 | 5.4941E-11 | 2.5721E-07 | **3.6322E-01** | **7.0617E-01** | 8.9934E-11 | 1.0666E-07 | 2.2539E-04 | 1.7649E-02 |

Table 14

Wilcoxon rank sum test statistical results.

| MISO VS. | CEC2017 (Dim=30) | CEC2017 (Dim=50) | CEC2017 (Dim=100) | CEC2022 (Dim=10) | CEC2022 (Dim=20) |
|---|---|---|---|---|---|
| GWO | 22/0/8 | 25/0/5 | 26/0/4 | 10/0/2 | 8/0/4 |
| SCA | 28/0/2 | 30/0/0 | 28/0/2 | 12/0/0 | 11/0/1 |
| WOA | 30/0/0 | 28/0/2 | 30/0/0 | 12/0/0 | 12/0/0 |
| HHO | 30/0/0 | 30/0/0 | 30/0/0 | 12/0/0 | 12/0/0 |
| MFO | 30/0/0 | 29/0/1 | 29/0/1 | 9/0/3 | 12/0/0 |
| AVOA | 27/0/3 | 29/0/1 | 27/0/3 | 8/0/4 | 9/0/3 |
| CSA | 27/0/3 | 27/0/3 | 27/0/3 | 9/0/3 | 8/0/4 |
| CPSOGSA | 27/0/3 | 29/0/1 | 28/0/2 | 10/0/2 | 11/0/1 |
| AO | 29/0/1 | 30/0/0 | 30/0/0 | 11/0/1 | 11/0/1 |
| DBO | 28/0/2 | 28/0/2 | 28/0/2 | 11/0/1 | 11/0/1 |
| SO | 22/0/8 | 24/0/6 | 30/0/0 | 7/0/5 | 9/0/3 |
| Overall(+/=/-) | **300**/0/30 | **309**/0/21 | **313**/0/17 | **111**/0/21 | **114**/0/18 |

4.5.2 Friedman mean rank test

The nonparametric Friedman mean rank test [96] is employed to rank the experimental results of MISO and other algorithms on the CEC2017 and CEC2022 test sets. The experimental results are presented in Table 15. Obviously, MISO consistently ranks first, signifying that our proposed optimizer outperforms other competitive algorithms on the considered test suites.

Table 15

The Friedman mean rank test was performed on the considered test suite.

| Suites | CEC 2017 | | | | | | CEC 2022 | | | |
|---|---|---|---|---|---|---|---|---|---|---|
| Dimensions | 30 | | 50 | | 100 | | 10 | | 20 | |
| Algorithms | Ave. Rank | Overall Rank | Ave. Rank | Overall Rank | Ave. Rank | Overall Rank | Ave. Rank | Overall Rank | Ave. Rank | Overall Rank |
| GWO | 4.40 | 4 | 4.17 | 3 | 4.13 | 4 | 5.42 | 4 | 4.67 | 4 |
| SCA | 11.00 | 12 | 11.33 | 12 | 11.27 | 12 | 9.92 | 10 | 9.75 | 10 |
| WOA | 10.83 | 11 | 10.67 | 11 | 10.40 | 11 | 10.83 | 10 | 10.75 | 10 |
| HHO | 8.90 | 10 | 7.67 | 9 | 6.97 | 6 | 9.33 | 9 | 9.00 | 9 |
| MFO | 6.57 | 6 | 6.97 | 6 | 7.87 | 9 | 5.42 | 4 | 6.75 | 6 |
| AVOA | 5.07 | 5 | 4.60 | 4 | 3.53 | 2 | 6.42 | 4 | 4.75 | 4 |
| CSA | 4.37 | 3 | 4.60 | 4 | 5.20 | 5 | 2.67 | 2 | 4.08 | 3 |
| CPSOGSA | 7.50 | 8 | 7.47 | 8 | 7.00 | 7 | 8.17 | 5 | 8.33 | 4 |
| AO | 8.20 | 9 | 8.87 | 10 | 9.03 | 10 | 8.08 | 4 | 8.33 | 4 |
| DBO | 6.60 | 7 | 7.30 | 7 | 7.23 | 8 | 6.58 | 3 | 6.58 | 3 |
| SO | 2.87 | 2 | 3.03 | 2 | 3.97 | 3 | 3.00 | 2 | 3.42 | 2 |
| MISO | **1.70** | **1** | **1.33** | **1** | **1.40** | **1** | **2.17** | **1** | **1.58** | **1** |

### 4.6 Impact analysis of the proposed strategy

This section demonstrates the validity of the proposed three strategies in SO. The availability of the proposed strategy is assessed by using CEC2017 (100 dim) and CEC2022 (20 dim) test sets. The tested approaches include SO, MISO, DSO, LSO, and BSO. The experimental results are plotted in Fig. 13. The unimodal function curves clearly demonstrate that these methods enhance the original SO and are practical. Nevertheless, for multimodal functions, a single enhancement strategy is not effective, and only the MISO algorithm, which combines the three improvement technologies, outperforms the SO in avoiding local optima and premature convergence. In conclusion, the combined use of the three strategies for improving SO proves to be effective.

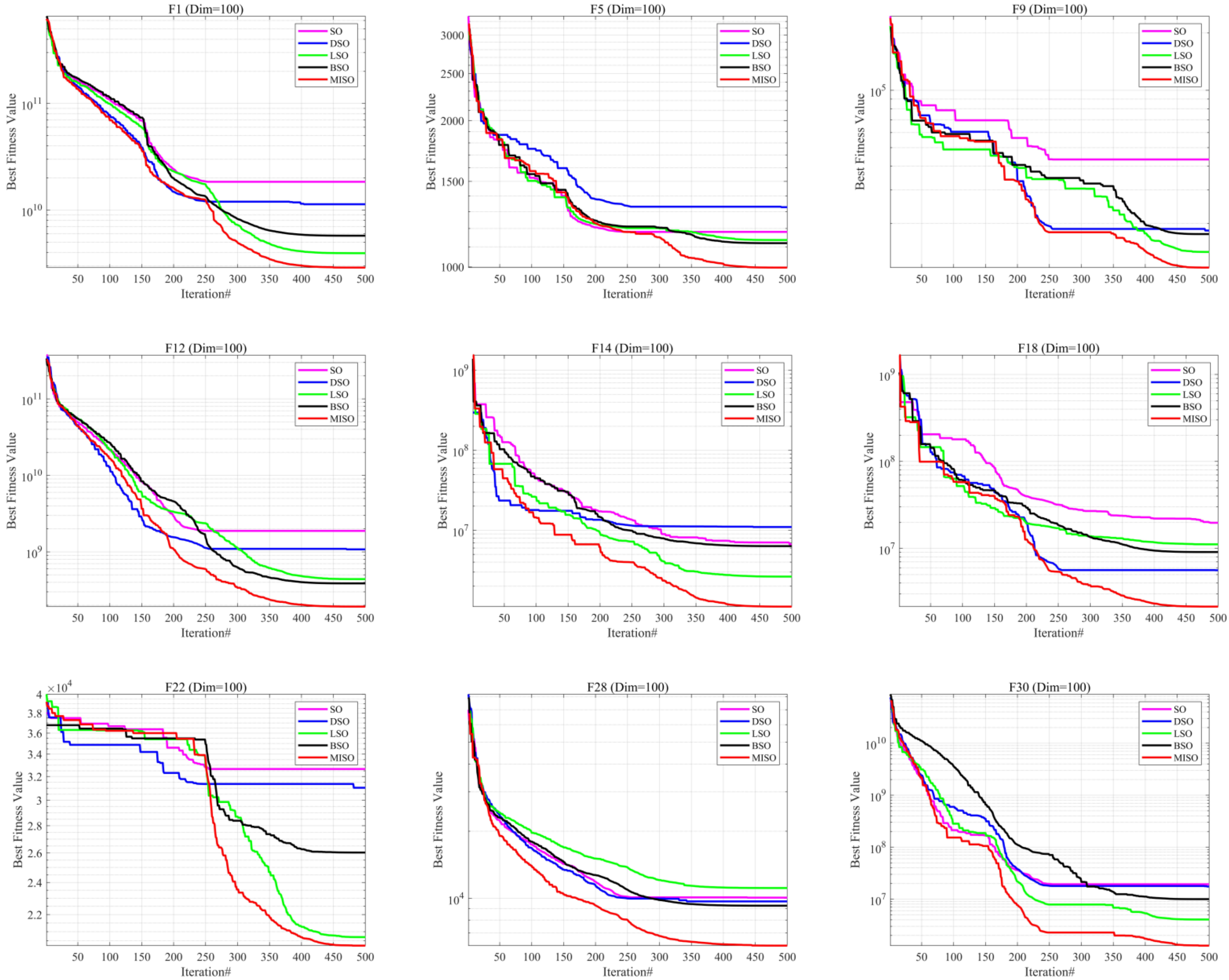

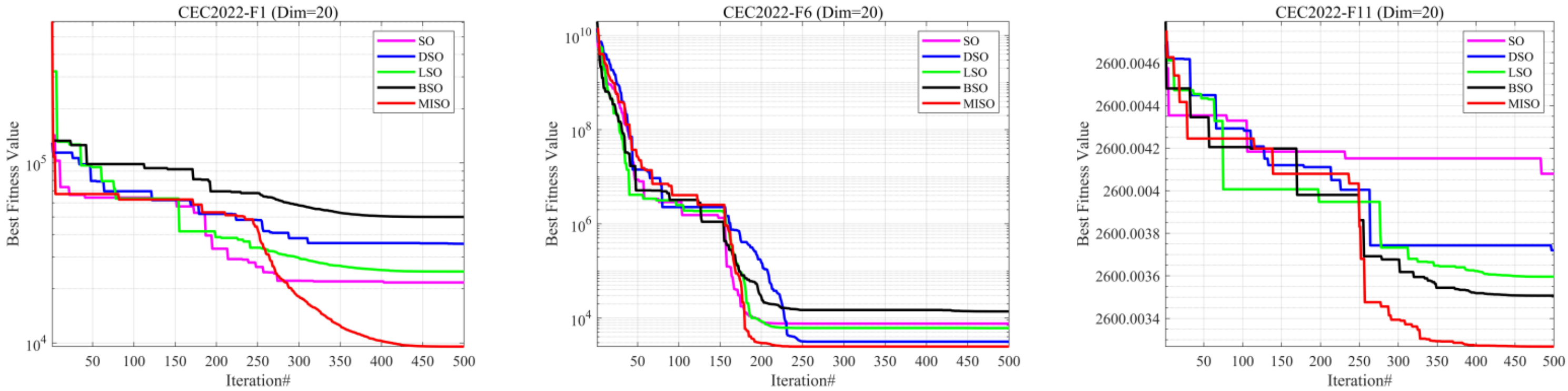

Fig. 13. Comparison of different improvement strategies.

### 4.7 Time Complexity Analysis of MISO and SO

In order to obtain a metaheuristic algorithm with good performance, it is necessary to reduce its time complexity as much as possible to minimize the cost overhead. Through the above experimental results, we can see that the proposed MISO algorithm performs better than SO. However, although MISO is superior in performance, we expect that its time complexity will not exceed that of SO. Therefore, in this section, we test the time complexity of MISO and SO. In Table 16, when the number of populations is set to 30 and the maximum number of iterations is set to 500, we perform 30 independent runs of SO and MISO and record their required time. As can be seen from the table, the computation time of MISO is lower than that of SO under the same setting. In summary, the proposed MISO algorithm not only improves the performance of SO algorithm, but also reduces the time complexity of SO algorithm.

Table 16

Execution time comparison between IMSO and SO.

| CEC2017(Dim=30) | F1 | F2 | F3 | F4 | F5 | F6 | F7 | F8 | F9 | F10 |
|---|---|---|---|---|---|---|---|---|---|---|
| SO | 2.3 | 2.48 | 2.33 | 2.3 | 2.74 | 4.14 | 2.87 | 2.84 | 2.84 | 3.14 |
| MISO | 1.63 | 1.68 | 1.54 | 1.54 | 1.97 | 3.17 | 2.01 | 2.02 | 2.02 | 2.31 |
| CEC2017(Dim=30) | F11 | F12 | F13 | F14 | F15 | F16 | F17 | F18 | F19 | F20 |
| SO | 2.58 | 2.84 | 2.64 | 3.17 | 2.54 | 2.73 | 3.69 | 2.75 | 9.32 | 3.88 |
| MISO | 1.75 | 1.99 | 1.77 | 2.29 | 1.71 | 1.88 | 2.74 | 1.88 | 7.81 | 2.93 |
| CEC2017(Dim=30) | F21 | F22 | F23 | F24 | F25 | F26 | F27 | F28 | F29 | F30 |
| SO | 4.96 | 5.69 | 6.25 | 6.79 | 6.03 | 7.32 | 8.2 | 7.26 | 5.64 | 11.28 |
| MISO | 3.9 | 4.42 | 4.91 | 5.39 | 4.71 | 6 | 6.76 | 5.89 | 4.39 | 9.53 |
| CEC2017(Dim=50) | F1 | F2 | F3 | F4 | F5 | F6 | F7 | F8 | F9 | F10 |
| SO | 4.13 | 4.46 | 4.08 | 4.11 | 4.9 | 7.44 | 7.63 | 12.08 | 12.66 | 14.14 |
| MISO | 3.19 | 3.39 | 7.41 | 8.46 | 9.35 | 15.82 | 9.89 | 9.81 | 9.89 | 11.63 |
| CEC2017(Dim=50) | F11 | F12 | F13 | F14 | F15 | F16 | F17 | F18 | F19 | F20 |
| SO | 11.3 | 12.89 | 11.39 | 13.81 | 11.1 | 11.92 | 16.92 | 11.82 | 37.42 | 18.18 |
| MISO | 9.11 | 10.59 | 9.33 | 5.47 | 3.41 | 3.8 | 5.19 | 3.68 | 13.56 | 5.43 |
| CEC2017(Dim=50) | F21 | F22 | F23 | F24 | F25 | F26 | F27 | F28 | F29 | F30 |
| SO | 24.97 | 25 | 29.52 | 18.34 | 13.02 | 34.56 | 41.59 | 36.7 | 27.43 | 47.68 |
| MISO | 8.52 | 9.4 | 10.92 | 11.74 | 11.31 | 13.79 | 16.01 | 17.42 | 20.88 | 45.07 |
| CEC2017(Dim=100) | F1 | F2 | F3 | F4 | F5 | F6 | F7 | F8 | F9 | F10 |
| SO | 11.88 | 16.31 | 26.6 | 27.07 | 29.2 | 41.62 | 29.21 | 29.28 | 29.59 | 29.05 |
| MISO | 11.08 | 14.96 | 23.13 | 23.19 | 25.41 | 38.12 | 26.15 | 26.18 | 25.8 | 28.18 |
| CEC2017(Dim=100) | F11 | F12 | F13 | F14 | F15 | F16 | F17 | F18 | F19 | F20 |
| SO | 27.7 | 30.69 | 27.93 | 32.57 | 27.56 | 29.16 | 38.05 | 29.31 | 79.75 | 37.19 |
| MISO | 24.22 | 27.27 | 24.65 | 29.52 | 24.14 | 25.52 | 34.66 | 25.61 | 75.14 | 36.65 |
| CEC2017(Dim=100) | F21 | F22 | F23 | F24 | F25 | F26 | F27 | F28 | F29 | F30 |
| SO | 69.33 | 71.5 | 87.92 | 89.08 | 98.06 | 106.75 | 110.36 | 112.25 | 64.94 | 49.85 |
| MISO | 65.4 | 66.71 | 83.61 | 86.17 | 94.58 | 102.68 | 103.6 | 111.73 | 51.4 | 38.37 |
| CEC2022(Dim=10) | F1 | F2 | F3 | F4 | F5 | F6 | F7 | F8 | F9 | F10 |
| SO | 1.14 | 1.04 | 1.59 | 1.24 | 1.25 | 1.08 | 1.72 | 1.97 | 1.72 | 1.61 |
| MISO | 0.76 | 0.62 | 1.16 | 0.79 | 0.82 | 0.66 | 1.28 | 1.5 | 1.22 | 1.15 |
| CEC2022(Dim=10) | F11 | F12 | | | | | | | | |
| SO | 2.11 | 2.12 | | | | | | | | |
| MISO | 1.61 | 1.68 | | | | | | | | |
| CEC2022(Dim=20) | F1 | F2 | F3 | F4 | F5 | F6 | F7 | F8 | F9 | F10 |
| SO | 1.72 | 1.48 | 2.63 | 1.83 | 1.85 | 1.56 | 2.84 | 3.14 | 3.02 | 2.55 |
| MISO | 1.03 | 0.89 | 1.99 | 1.21 | 1.25 | 0.98 | 2.21 | 2.46 | 2.37 | 1.93 |
| CEC2022(Dim=20) | F11 | F12 | | | | | | | | |
| SO | 3.64 | 5.52 | | | | | | | | |
| MISO | 2.94 | 3.91 | | | | | | | | |

## 5. Application of MISO algorithm to UAV 3D path planning and engineering design problems

In this section, the application of MISO algorithm in UAV 3D path planning is described and compared with other 11 algorithms. MISO has achieved impressive results in the competition. In addition, in order to verify the wide applicability and accuracy of MISO, we applied it to six engineering application problems, and compared with other 11 optimization algorithms, MISO obtained better results.

### 5.1 3D Path planning of UAV

Unmanned Aerial Vehicles (UAV) play a vital role in many civil and military fields, and their importance and convenience are widely recognized. As a core task of the autonomous control system of unmanned Aerial Vehicle (UAV), path planning and design aims to solve a complex constrained optimization problem: finding a reliable and safe path from a starting point to a goal point under certain constraints. In recent years, with the wide application of unmanned aerial vehicle (UAV), the research on path planning problem has attracted wide attention. Therefore, we use MISO to solve the UAV path planning problem and verify the effectiveness of the algorithm. A specific mathematical model of this problem is outlined as follows.

The starting point of UAV flight is denoted as ($x_s$, $y_s$, $z_s$), the ending point remember ($x_e$, $y_e$, $z_e$). By cubic spline interpolation to generate a smooth curve, and the discrete points ($x_s$, $y_s$, $z_s$), ($x_1$, $y_1$, $z_1$), ..., ($x_{g-1}$, $y_{g-1}$, $z_{g-1}$), ($x_e$, $y_e$, $z_e$). This curve is then represented as a sequence of discrete points ($h_1$, $h_2$, ..., $h_g$), which are the coordinates of the $h_m$ ($x_m$, $y_m$, $z_m$). Therefore, the objective function of this problem can be derived, as shown in Eq. (32)

$$F_{tc} = w_1 \times F_{pc} + w_2 \times F_{hc} + w_3 \times F_{sc} \tag{32}$$

where $F_{tc}$ is the total cost, $F_{pc}$ is the cost of path length, $F_{hc}$ is the cost of height standard deviation, $F_{sc}$ is the smoothness cost of the planned path, and $w_i$ ($i$ = 1,2,3) is the weight. The constraint on the weight coefficient is given in

Eq. (33).

$$\begin{cases} w_i \geq 0 \\ \sum_{i=1}^{3} w_i = 1 \end{cases} \tag{33}$$

Under normal circumstances, the flight of UAV should save time and reduce costs as much as possible under the premise of ensuring safety, so the path length of path planning is crucial. The mathematical model is described by Eq. (34).

$$F_{pc} = \sum_{m=1}^{g-1} \| (x_{m+1}, y_{m+1}, z_{m+1}) - (x_m, y_m, z_m) \|_2 \tag{34}$$

where $(x_m, y_m, z_m)$ represents the $m^{\text{th}}$ waypoint on the planned path of the UAV.

In addition, the flight height of the drone has a great impact on the control system and safety, so we need to fully consider this factor. Eq. (35) gives the mathematical model of this problem.

$$F_{hc} = \sqrt{\sum_{m=1}^{g} (z_m - \frac{1}{n}\sum_{k=1}^{g} z_m)^2} \tag{35}$$

Finally, we also need to consider the influence of the UAV when it turns, and the mathematical model is displayed by Eq. (36).

$$F_{sc} = \sum_{m=1}^{g} \left( \cos(phi) - \frac{\varphi_{m+1} \times \varphi_m}{|\varphi_{m+1}| \times |\varphi_m|} \right) \tag{36}$$

where $\varphi_m$ represents $(x_{m+1} - x_m, y_{m+1} - y_m, z_{m+1} - z_m)$.

In summary, we can obtain the model of UAV path planning optimization problem, which is specifically expressed as Eq. (37).

$$\begin{gathered} \min_{L} F_{\mathrm{tc}}(L) \\ \mathrm{s.t.\,path}(L) \notin Ground \cup Obstacle \end{gathered} \tag{37}$$

where $L$ is the path available for UAV flight, $Ground$ and $Obstacle$ represent the ground and obstacles, respectively. In this paper, we will establish the model of the ground and obstacles according to Eq. (38).

$$z = \sin(y+1) + \sin(x) + \cos(x^2 + y^2) + 2 \times \cos(y) + \sin(x^2 + y^2) \tag{38}$$

In this study, the starting point of the UAV is set as (0,0,20), and the ending point is set as (200,200,30). A feasible and smooth path was successfully obtained by combining cubic spline interpolation with several optimization methods. The experimental parameters are consistent with those described in the previous chapters. The experimental results are obtained based on 30 independent runs of 12 optimizers. As shown in Table 17. Compared with other optimization techniques, MISO shows better performance, and SCA has the worst effect. The convergence curve is shown in Fig. 14, which clearly demonstrates the faster convergence speed of MISO. Meanwhile, Fig. 15 illustrates the paths of the 12 contestants. The experimental results show that the curves generated by the MISO algorithm are smoother, which further validates the potential of the MISO algorithm in solving practical problems.

Table 17

Experimental results of competitors in UAV 3D trajectory planning.

| Algorithms | Best | Median | Worst | Ave | Std | Friedman ranking | Wilcoxon |
|---|---|---|---|---|---|---|---|
| GWO | 234.251630 | 414.745743 | 573.017029 | 402.975017 | 65.418267 | 5 | (+) |
| SCA | 332.893006 | 434.012764 | 645.981842 | 445.959111 | 61.129866 | 12 | (+) |
| WOA | 246.382192 | 431.371087 | 639.243829 | 435.027803 | 100.561120 | 11 | (+) |
| HHO | 246.796834 | 430.637849 | 555.264678 | 431.679853 | 71.302067 | 10 | (+) |
| MFO | 228.574346 | 420.303406 | 533.231279 | 409.274588 | 67.905116 | 8 | (+) |
| AVOA | 228.590373 | 409.254095 | 628.034515 | 388.891270 | 87.310938 | 4 | (+) |
| CSA | 228.672545 | 420.105215 | 439.694505 | 356.886009 | 94.129457 | 3 | (+) |
| CPSOGSA | 228.584243 | 397.904572 | 641.185677 | 416.504615 | 105.218598 | 6 | (+) |
| AO | 253.584159 | 420.749899 | 526.066155 | 416.078473 | 62.785764 | 9 | (+) |
| DBO | 228.685189 | 420.207176 | 440.518336 | 399.358497 | 59.604683 | 7 | (+) |
| SO | 228.595583 | 376.796098 | 485.203975 | 326.804797 | 89.405980 | 2 | (+) |
| MISO | **228.568619** | **229.001349** | **413.520091** | **291.751514** | **49.080927** | **1** | — |

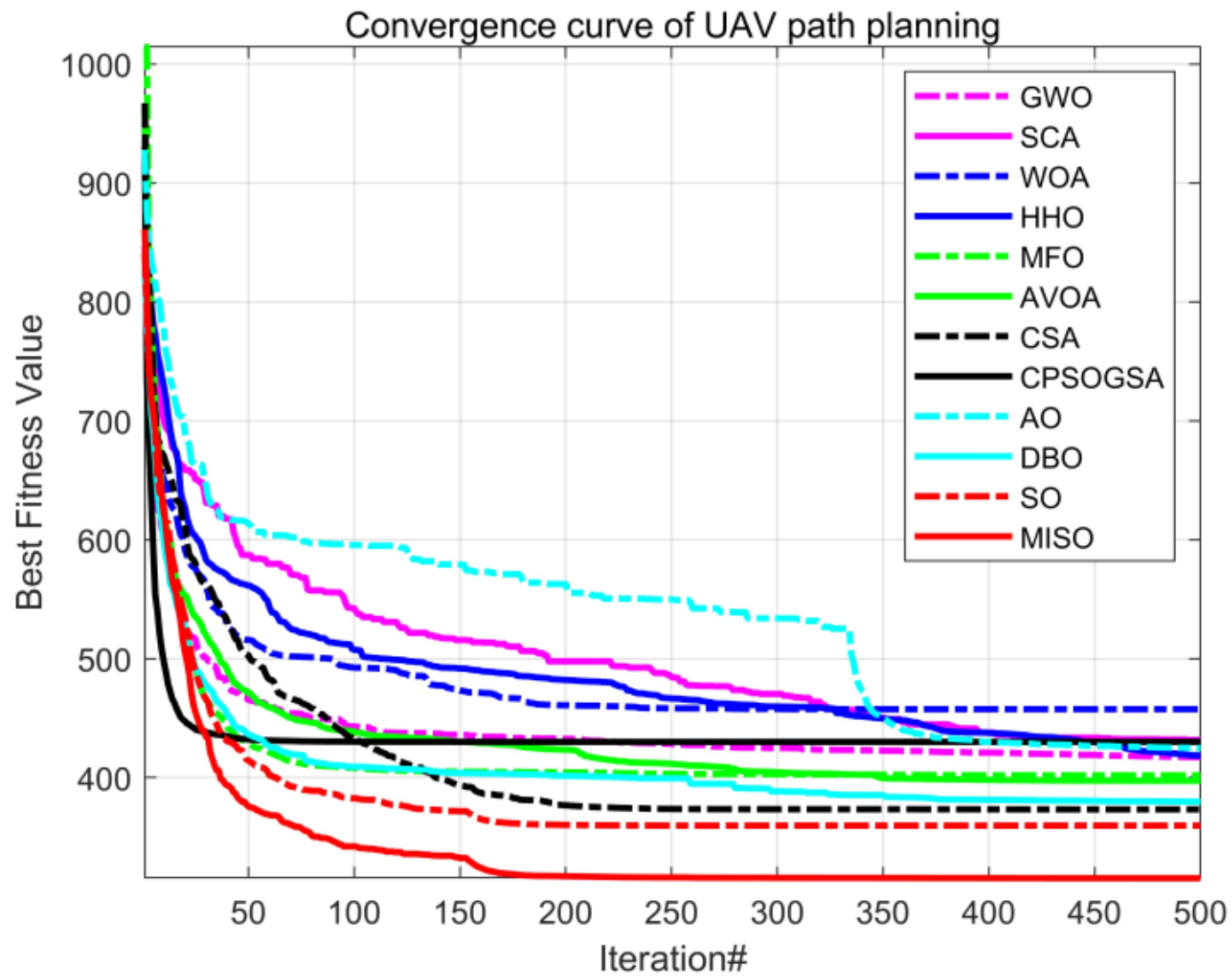

Fig. 14. Convergence curve of 12 competitors on the UAV 3D path planning problem.

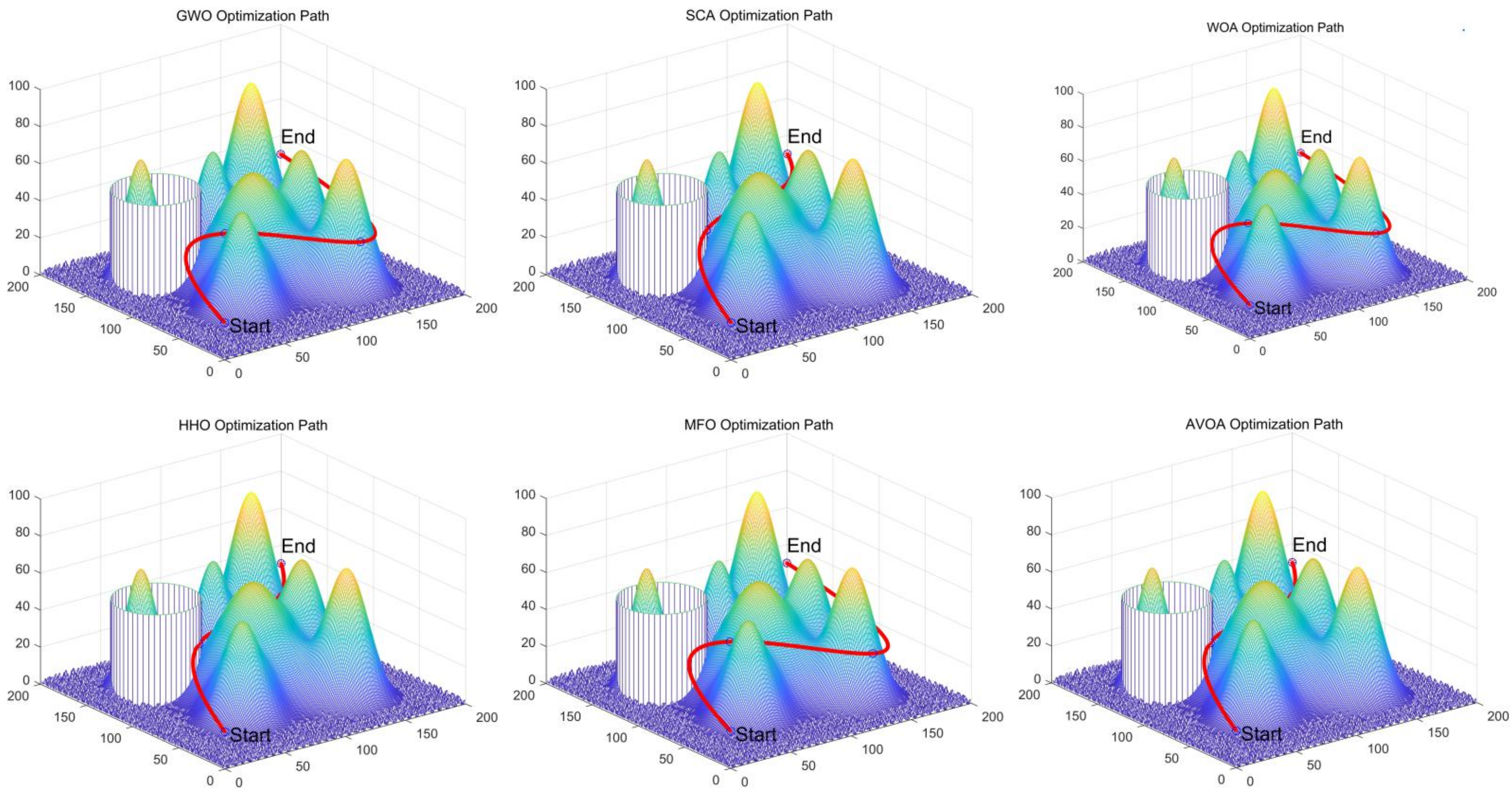

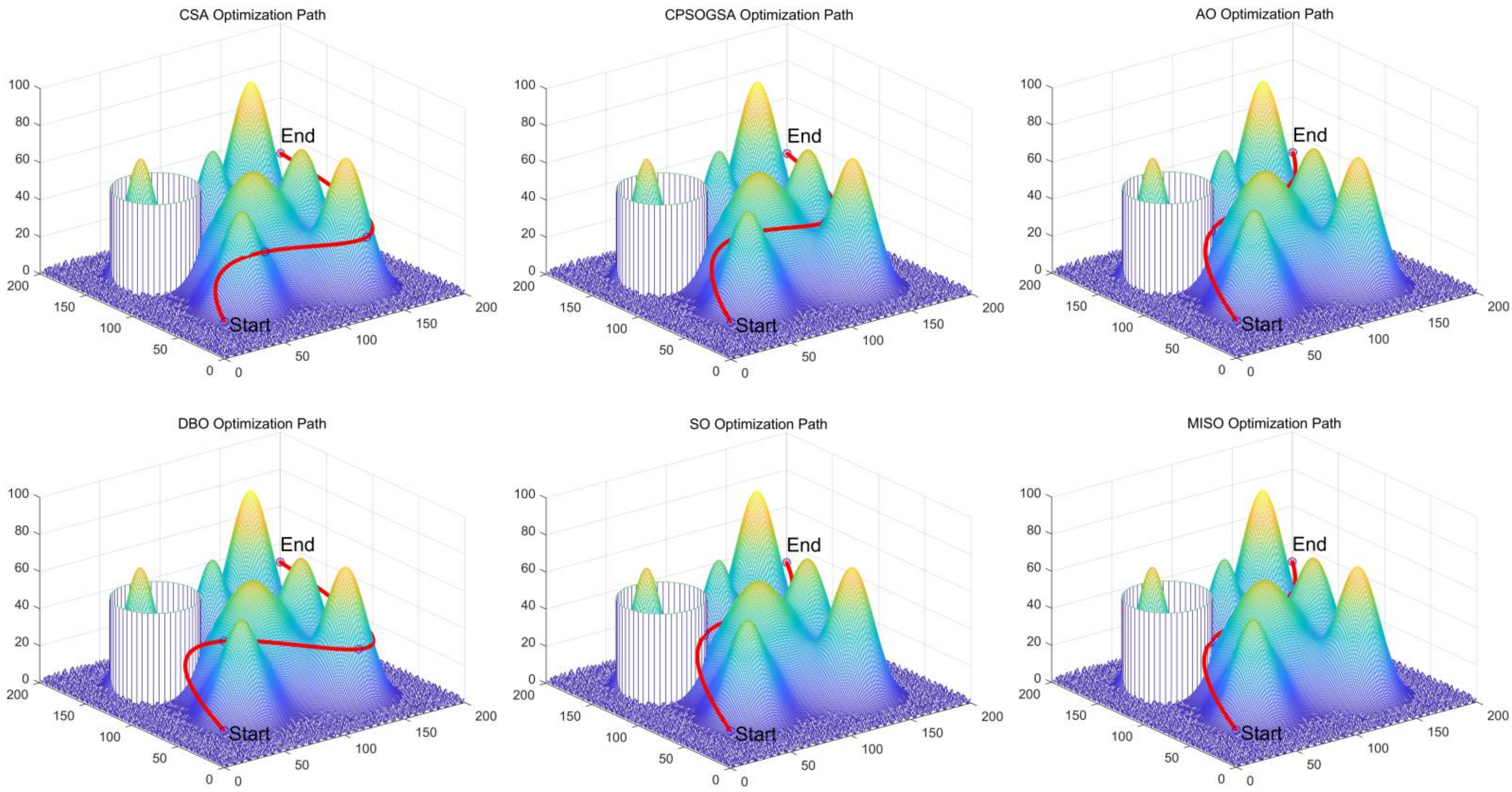

Fig. 15. Optimization path diagram for each comparison algorithm.

**5.2 The engineering design problems**

In the previous section, we used benchmark test functions to demonstrate the performance of MISO. The Friedman ranking test verifies the overall effectiveness. An efficient algorithm needs to pass the test of practical problems. Therefore, in this section, we use our proposed MISO to solve real-world applications that are more complex than test functions, which is a tremendous challenge for our optimizer. Firstly, we applied MISO to six engineering design problems, namely the welded beam design problem (WBD) [97], tension/compression spring design problem (T/CSD) [98], cantilever beam design problem (CBD) [99], rolling element bearing design (REBD) [100], speed reducer design (SRD) [101] and Three-bar truss design problem (T-bTD) [102]. We compared MISO's results with nine other well-known algorithms, including GWO [19], SCA [82], WOA [4], HHO [33], MFO [24], AVOA [42], CSA [44], CPSOGSA [92], AO [85], DBO [55], and the original SO [59]. Table 3 summarizes the parameters of these competitors. We recorded the optimal solution (Best), median (Median), worst cost (Worst), mean (Ave), standard deviation (Std), Friedman ranking, wilcoxon rank-sum test (Wilcoxon) and highlighted the best results.

5.2.1 Welded beam design problem (WBD)

The target of the welded beam design problem is to reduce the economic cost by optimizing the thickness ($h$), length ($l$), height ($t$), bottom width ($b$) of the beam and rod, as well as the weld seam. The structure of WBD is shown in Fig. 16, and its mathematical model is displayed by Eq. (39).

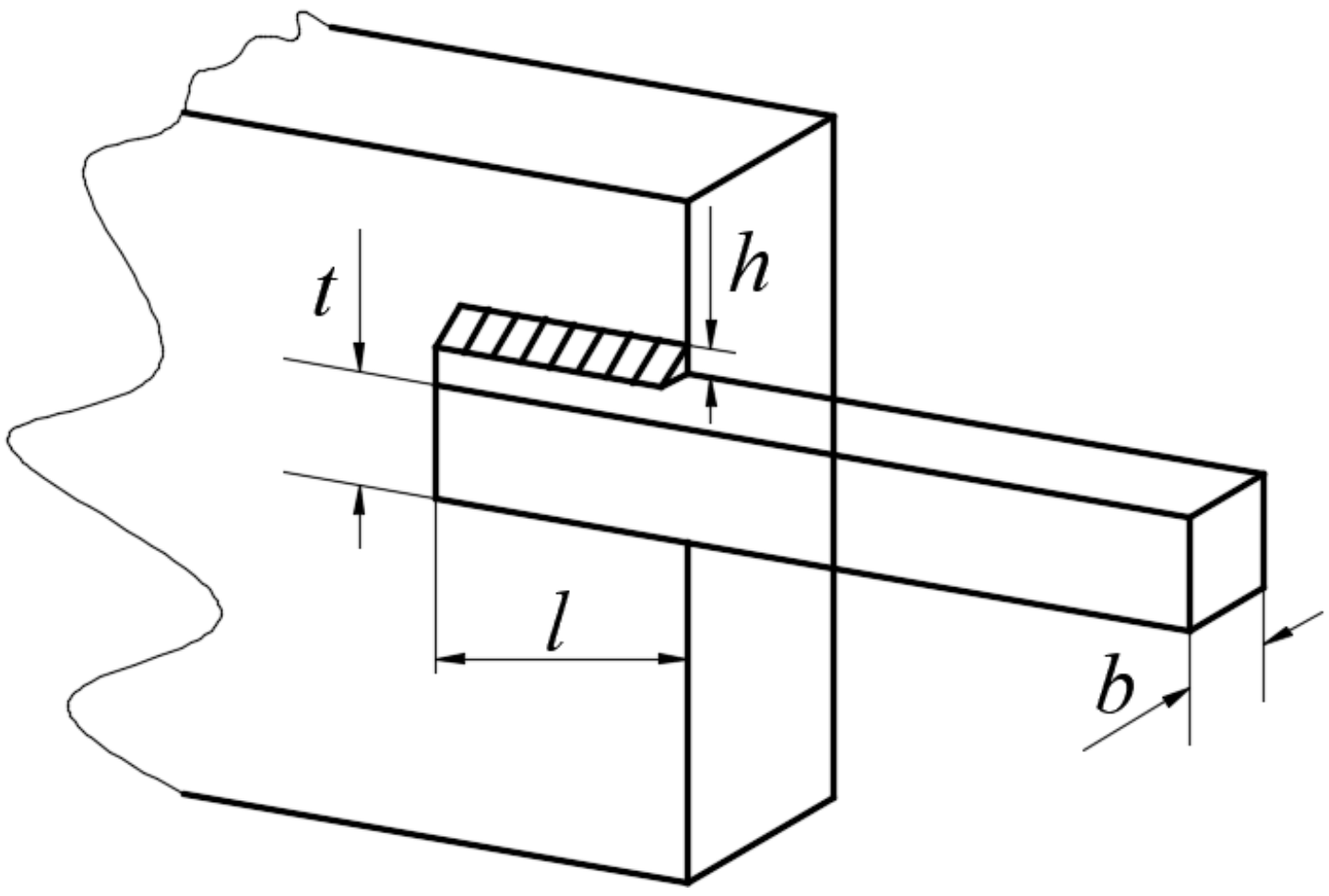

Fig. 16. Schematic representation of the WBD.

Consider: $\vec{x} = [x_1\ x_2\ x_3\ x_4] = [h\ l\ t\ b]$,

Minimize: $f(\vec{x}) = f(\vec{x}) = 1.10471x_1^2x_2 + 0.04811x_3x_4(14.0 + x_2)$,

Subject to:

$$
\begin{aligned}
&g_1(\vec{x}) = \tau(\vec{x}) - \tau_{max} \leqslant 0,\\
&g_2(\vec{x}) = \sigma(\vec{x}) - \sigma_{max} \leqslant 0,\\
&g_3(\vec{x}) = \delta(\vec{x}) - \delta_{max} \leqslant 0,\\
&g_4(\vec{x}) = x_1 - x_4 \leqslant 0,\\
&g_5(\vec{x}) = P - P_c(\vec{x}) \leqslant 0,\\
&g_6(\vec{x}) = 0.125 - x_1 \leqslant 0,\\
&g_7(\vec{x}) = 1.10471x_1^2 + 0.04811x_3x_4(14.0 + x_2) - 5.0 \leqslant 0,
\end{aligned}
$$

Parameter range: $0.1 \leqslant x_1, x_4 \leqslant 2,\ 0.1 \leqslant x_2, x_3 \leqslant 10$,

Where,

$$
\begin{aligned}
&\tau(\vec{x}) = \sqrt{(\tau')^2 + 2\tau'\tau''\frac{x_2}{2R} + (\tau')^2},\\
&\tau' = \frac{p}{\sqrt{2x_1x_2}}, \tau'' = \frac{MR}{J},\\
&M = P\left(L + \frac{x_2}{2}\right),\\
&R = \sqrt{\frac{x_2^2}{4} + \left(\frac{x_1+x_3}{2}\right)^2},\\
&J = 2\left\{\sqrt{2x_1x_2}\left[\frac{x_2^2}{4} + \left(\frac{x_1+x_3}{2}\right)^2\right]\right\},\\
&\sigma(\vec{x}) = \frac{6PL}{x_4x_3^2}, \delta(\vec{x}) = \frac{6PL^3}{Ex_3^2x_4},\\
&P_c(\vec{x}) = \frac{\sqrt{\frac{40x}{36}}}{L^2}\left(1 - \frac{x_3^2}{2L}\sqrt{\frac{E}{4G}}\right),\\
&P = 6000lb, L = 14in, \delta_{max} = 0.25in,\\
&E = 30 \times 1^6\ psi, G = 12 \times 10^6\ psi,\\
&\tau_{max} = 13600psi, \sigma_{max} = 30000psi.
\end{aligned}
\tag{39}
$$

Table 18 details the results and statistical analysis of the 12 optimizers used to solve the WBD problem. In general, the optimization results of MISO algorithm are the most prominent.

Table 18

Comparative results for WBD.

| Algorithms | Best | Median | Worst | Ave | Std | Friedman ranking | Wilcoxon |
|---|---|---|---|---|---|---|---|
| GWO | 1.671496 | 1.674258 | 1.884354 | 1.674961 | 0.046537 | 3 | (-) |
| SCA | 1.734715 | 1.823262 | 1.927289 | 1.825547 | 0.049367 | 8 | (+) |
| WOA | 1.748200 | 2.194410 | 3.947607 | 2.332473 | 0.531256 | 11 | (+) |
| HHO | 1.709655 | 1.947397 | 2.854893 | 2.081799 | 0.306001 | 9 | (+) |
| MFO | 1.670236 | 1.722715 | 2.589517 | 1.790385 | 0.213321 | 5 | (+) |
| AVOA | 1.671148 | 1.778634 | 1.826733 | 1.764999 | 0.048991 | 7 | (+) |
| CSA | **1.670219** | 1.673875 | 1.850195 | 1.697198 | 0.049240 | 2 | (-) |
| CPSOGSA | 3.057225 | 6.566424 | 17.055918 | 6.780318 | 2.657184 | 12 | (+) |
| AO | 1.776060 | 1.954713 | 2.280363 | 1.977384 | 0.133254 | 10 | (+) |
| DBO | 1.670233 | 1.751091 | 2.282913 | 1.776144 | 0.115144 | 6 | (+) |
| SO | 1.670441 | 1.680505 | 2.136209 | 1.714653 | 0.096261 | 4 | (+) |
| MISO | **1.670219** | **1.671058** | **1.818081** | **1.690940** | **0.045546** | **1** | — |

5.2.2 Tension/compression spring design problem (T/CSD)

The main objective of the tension/compression spring design problem is to minimize the weight by choosing three variables, wire diameter (*d*), average coil diameter (*D*), and number of active coils (*N*), while satisfying the usage criteria, as given in Fig. 17. The mathematical model is described by Eq. (40).

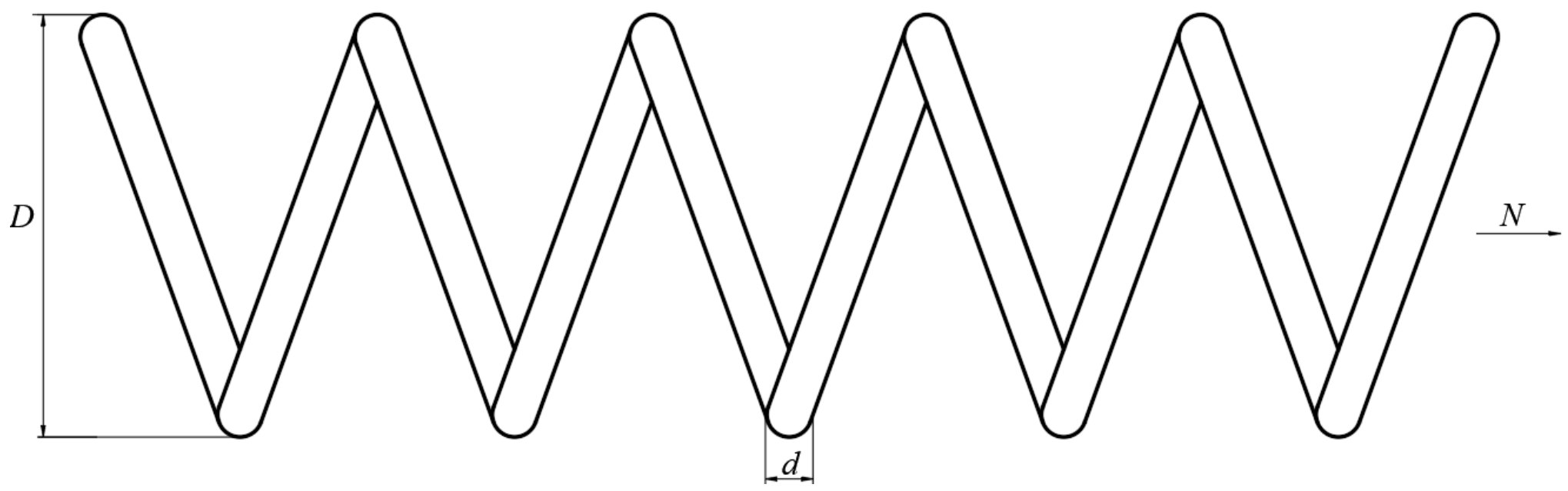

Fig. 17. Schematic representation of the T/CSD.

Fig. 11.

Consider: $\vec{x} = [x_1 \quad x_2 \quad x_3] = [d\ D\ N]$,

Minimize: $f(\vec{x}) = (x_3 + 2)x_2 x_1^2$,

Subject to:

$$g_1(\vec{x}) = 1 - \frac{x_2^3 x_3}{71785 x_1^4} \leqslant 0,$$

$$g_2(\vec{x}) = \frac{4x_2^2 - x_1 x_2}{12566(x_2 x_1^3 - x_1^4)} + \frac{1}{5108 x_1^2} \leqslant 0, \tag{40}$$

$$g_3(\vec{x}) = 1 - \frac{140.45 x_1}{x_2^2 x_3} \leqslant 0,$$

$$g_4(\vec{x}) = \frac{x_1 + x_2}{1.5} - 1 \leqslant 0$$

Parameters range: $0.05 \leqslant x_1 \leqslant 2,\ 0.25 \leqslant x_2 \leqslant 1.3,\ 2 \leqslant x_3 \leqslant 15.$

Table 19 compares MISO with the other 11 optimization algorithms from several perspectives. CSA and our optimizer obtain the minimum cost of the objective function. Moreover, MISO has the smallest median value, worst value, mean value and standard deviation, which confirms the effectiveness and robustness of the algorithm in terms of results. In addition, the results of the Wilcoxon rank-sum test also show that there is a significant difference between MISO and most of the algorithms considered.

Table 19

Comparative results for T/CSD.

| Algorithms | Best | Median | Worst | Ave | Std | Friedman ranking | Wilcoxon |
|---|---|---|---|---|---|---|---|
| GWO | 0.012695 | 0.012828 | 0.013927 | 0.012937 | 0.000289 | 5 | (+) |
| SCA | 0.012833 | 0.013145 | 0.014224 | 0.013165 | 0.000261 | 9 | (+) |
| WOA | 0.012666 | 0.013232 | 0.017777 | 0.013989 | 0.001641 | 8 | (+) |
| HHO | 0.012676 | 0.013747 | 0.017773 | 0.014063 | 0.001263 | 10 | (+) |
| MFO | **0.012665** | 0.012975 | 0.017773 | 0.013953 | 0.001983 | 6 | (+) |
| AVOA | **0.012665** | 0.012803 | 0.015703 | 0.013141 | 0.000759 | 4 | (+) |
| CSA | **0.012665** | 0.012734 | 0.014017 | 0.012895 | 0.000325 | 2 | (-) |
| CPSOGSA | 3.3799E-02 | 3.7211E+14 | 8.6555E+14 | 3.9874E+14 | 3.8049E+14 | 12 | (+) |
| AO | 0.013605 | 0.018580 | 0.024769 | 0.018867 | 0.002499 | 11 | (+) |
| DBO | 0.012710 | 0.013000 | 0.017853 | 0.013552 | 0.001550 | 7 | (+) |
| SO | 0.012666 | 0.012767 | 0.017773 | 0.013125 | 0.000993 | 3 | (+) |
| MISO | **0.012665** | **0.012684** | **0.012719** | **0.012692** | **0.000026** | **1** | — |

5.2.3 Cantilever beam design problem (CBD)

The core of the cantilever design problem is to minimize the weight of the cantilever using five hollow square blocks of equal thickness. At the same time, the height (or width) of the beam is also taken into account to ensure minimum production costs. The structure of CBD is presented in Fig. 18, and Eq. (41) illuminates the mathematical model of CBD.

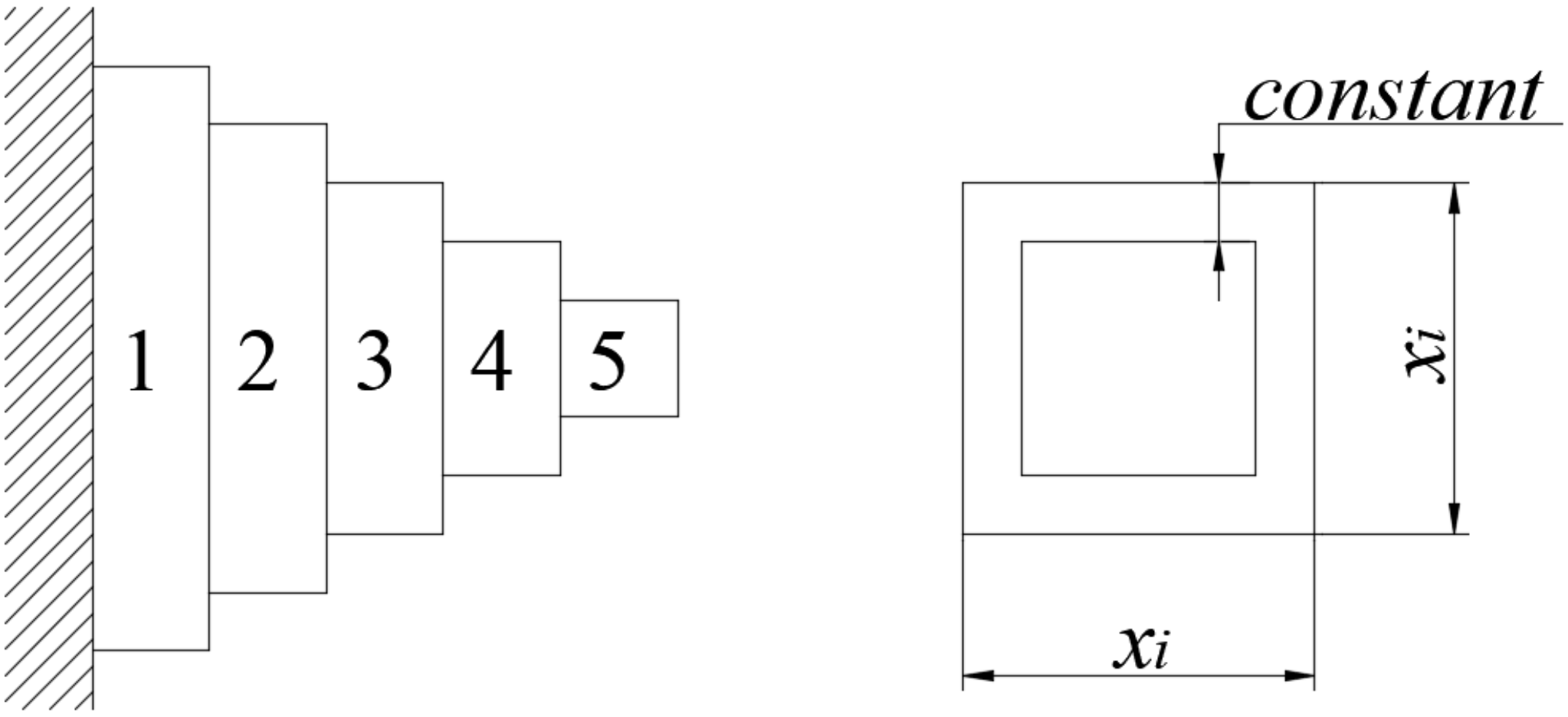

Fig. 18. Schematic representation of the CBD.

Consider: $\vec{x} = [x_1\ x_2\ x_3\ x_4\ x_5]$,

Minimize: $f(\vec{x}) = 0.6224(x_1 + x_2 + x_3 + x_4 + x_5)$,

Subject to: $$g(\vec{x}) = \frac{60}{x_1^3} + \frac{27}{x_2^3} + \frac{19}{x_3^3} + \frac{7}{x_4^3} + \frac{1}{x_5^3} - 1 \leqslant 0, \quad (41)$$

Parameter range: $0.01 \leqslant x_1, x_2, x_3, x_4, x_5 \leqslant 100$,

The optimization results of different competitors are reported in Table 20. Overall, we find that MISO performs best among all algorithms, yielding the lowest mean and standard deviation. This further validates the potential of the proposed technique for solving practical problems. Wilcoxon rank-sum test results denote a significant difference in performance between MISO and all other competitors.

Table 20

Comparison results for CBD.

| Algorithms | Best | Median | Worst | Ave | Std | Friedman ranking | Wilcoxon |
|---|---|---|---|---|---|---|---|
| GWO | 1.340587465 | 1.341507144 | 1.343053284 | 1.341616746 | 6.8083E-04 | 7 | (+) |
| SCA | 1.356827563 | 1.402434499 | 1.46458411 | 1.405234294 | 2.8017E-02 | 10 | (+) |
| WOA | 1.352746391 | 1.574699605 | 2.532955591 | 1.635925559 | 2.7476E-01 | 11 | (+) |
| HHO | 1.341269455 | 1.343349362 | 1.348583607 | 1.343911224 | 1.8397E-03 | 8 | (+) |
| MFO | 1.339974999 | 1.340514863 | 1.343582466 | 1.340893258 | 9.6060E-04 | 6 | (+) |
| AVOA | 1.340011163 | 1.340306003 | 1.3408304 | 1.340320329 | 2.1869E-04 | 5 | (+) |
| CSA | 1.339976513 | 1.339996283 | 1.340970351 | 1.340074165 | 2.2195E-04 | 2 | (+) |
| CPSOGSA | 2.72853826 | 8.443135486 | 12.56076863 | 8.315321304 | 2.1164E+00 | 12 | (+) |
| AO | 1.340929596 | 1.344591941 | 1.351686869 | 1.345084584 | 2.6374E-03 | 9 | (+) |
| DBO | 1.339972503 | 1.340094912 | 1.340353458 | 1.340104647 | 1.0230E-04 | 4 | (+) |
| SO | 1.339966404 | 1.34003546 | 1.340527629 | 1.340071901 | 1.1551E-04 | 3 | (+) |
| MISO | **1.339957649** | **1.339966738** | **1.339999494** | **1.33996924** | **9.9914E-06** | **1** | — |

5.2.4 Rolling element bearing design problem (REBD)

The design problem aims at optimizing the bearing capacity of rolling bearings and involves five design variables: pitch diameter ($D_m$), ball diameter ($D_b$), curvature coefficients of inner and outer raceways ($f_o$ and $f_i$), and total number of balls ($Z$). The other five design parameters $e$, $\epsilon$, $\zeta$, $K_{Dmax}$ and $K_{Dmin}$ appear only in constraint condition. Fig. 19 illustrates the structure of the optimization problem, whose mathematical model is given in Eq. (42).

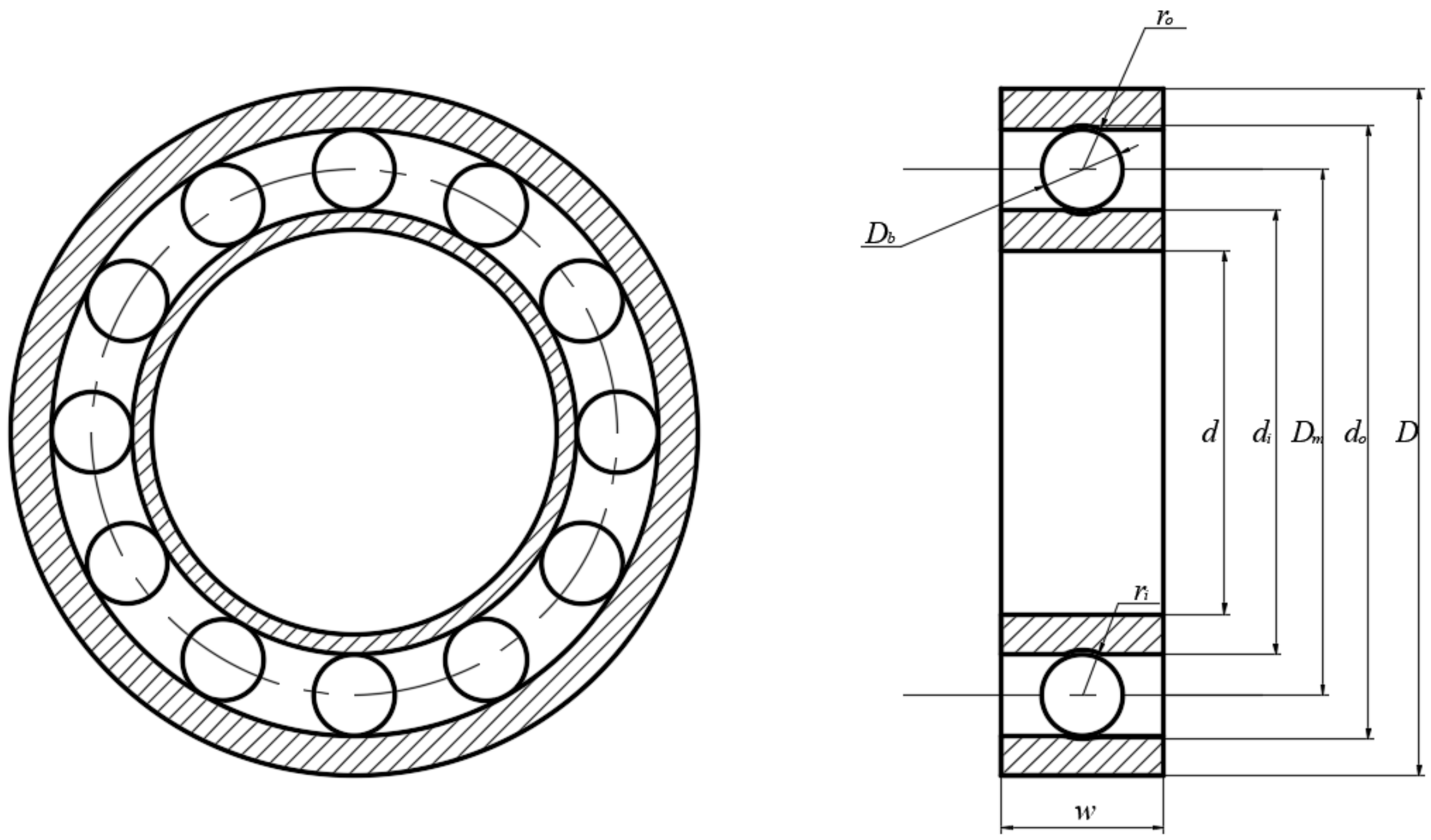

Fig. 19. Schematic representation of the REBD.

Consider: $\vec{x} = [x_1\ x_2\ x_3\ x_4\ x_5\ x_6\ x_7\ x_8\ x_9\ x_{10}] = [D_m\ D_b\ f_o\ f_i\ Z\ e\ \epsilon\ \zeta\ KD_{max}\ KD_{min}]$,

Minimize:
$$f(\vec{x}) = \begin{cases} f_c Z^{2/3} D_b^{1.8} & \text{, if } D_b \le 25.4\text{mm} \\ 3.647 f_c Z^{2/3} D_b^{1.4} & \text{, otherwise} \end{cases} \quad (42)$$

Subject to:

$$g_1(\bar{x}) = Z - \frac{\phi_0}{2\sin^{-1}(D_b/D_m)} - 1 \le 0,$$
$$g_2(\bar{x}) = K_{D\min}(D-d) - 2D_b \le 0,$$
$$g_3(\bar{x}) = 2D_b - K_{D\max}(D-d) \le 0,$$
$$g_4(\bar{x}) = D_b - {}_w \le 0,$$
$$g_5(\bar{x}) = 0.5(D+d) - D_m \le 0,$$
$$g_6(\bar{x}) = D_m - (0.5+e)(D+d) \le 0,$$
$$g_7(\bar{x}) = \epsilon D_b - 0.5(D - D_m - D_b) \le 0,$$
$$g_8(\bar{x}) = 0.515 - f_i \le 0,$$
$$g_9(\bar{x}) = 0.515 - f_0 \le 0,$$

Where,

$$f_c = 37.91\left\{1 + \left\{1.04\left(\frac{1-\gamma}{1+\gamma}\right)^{1.72}\left(\frac{f_i(2f_0-1)}{f_0(2f_i-1)}\right)^{0.41}\right\}^{10/3}\right\}^{-0.3},$$

$$\gamma = \frac{D_b\cos(\alpha)}{D_m}, f_i = \frac{r_i}{D_b}, f_0 = \frac{r_0}{D_b},$$

$$\phi_0 = 2\pi - 2\times\cos^{-1}\left(\frac{\{(D-d)/2 - 3(T/4)\}^2 + \{D/2 - (T/4) - D_b\}^2 - \{d/2 + (T/4)\}^2}{2\{(D-d)/2 - 3(T/4)\}\{D/2 - (T/4) - D_b\}}\right),$$

$$T = D - d - 2D_b, D = 160, d = 90, B_w = 30,$$

Parameters range:

$$0.5(D+d) \le D_m \le 0.6(D+d), 0.5(D+d) \le D_m \le 0.6(D+d), 4 \le Z \le 50,$$
$$0.515 \le f_i \le 0.6, 0.515 \le f_0 \le 0.6, 0.4 \le K_{Dmin} \le 0.5, \ 0.6 \le K_{Dmax} \le 0.7,$$
$$0.3 \le \epsilon \le 0.4, 0.02 \le e \le 0.1, 0.6 \le \zeta \le 0.85.$$

Table 21 compares MISO with the other 11 optimization algorithms from several aspects. The results show that MFO, DBO, SO and MISO all return the minimum cost of the objective function. In addition, MISO provides the smallest median, worst, mean, and standard deviation, which proves the effectiveness and robustness of the algorithm results. Similarly, the results of the Wilcoxon rank-sum test indicate a significant difference between RBMO and the other algorithms considered.

Table 21

Comparison results for REBD.

| Algorithms | Best | Median | Worst | Ave | Std | Friedman ranking | Wilcoxon |
|---|---|---|---|---|---|---|---|
| GWO | 16988.146584 | 17065.529761 | 17103.578455 | 17061.880824 | 25.341200 | 8 | (+) |
| SCA | 17077.008683 | 17480.554030 | 18252.040938 | 17538.831207 | 368.867884 | 10 | (+) |
| WOA | 16974.344580 | 17326.383395 | 25880.620181 | 17928.574074 | 1908.855657 | 9 | (+) |
| HHO | 16960.823578 | 17055.748892 | 25242.485983 | 17808.785095 | 1836.959124 | 5 | (+) |
| MFO | **16958.202287** | 16958.232287 | 17058.766922 | 16974.963060 | 38.118927 | 2 | (+) |
| AVOA | 16965.357475 | 17021.533779 | 17058.713724 | 17025.105337 | 21.639178 | 4 | (+) |
| CSA | 16992.608386 | 17058.766922 | 17983.253932 | 17333.028680 | 431.048658 | 7 | (+) |
| CPSOGSA | 2.4579E+04 | 1.3656E+16 | 7.5098E+16 | 1.9657E+16 | 1.9546E+16 | 12 | (+) |
| AO | 1.6991E+04 | 1.9205E+04 | 3.0229E+14 | 1.1178E+13 | 5.5313E+13 | 11 | (+) |
| DBO | **16958.202287** | 17058.766922 | 25748.569106 | 17760.267668 | 2149.291232 | 6 | (+) |
| SO | **16958.202287** | 16978.222815 | 17058.766922 | 16976.023005 | 23.633682 | 3 | (+) |
| MISO | **16958.202287** | **16958.203213** | **17017.528863** | **16968.989137** | **16.166518** | **1** | — |

5.2.5 Speed reducer design problem (SRD)

The objective of the problem is to reduce the total weight of the reducer by optimizing seven design variables, namely:

face width ($b$), tooth modulus ($m$), pint tooth number ($z$), first shaft length between bearings ($l_1$), second shaft length between bearings ($l_2$), first shaft diameter ($d_1$) and second shaft diameter ($d_2$). A schematic representation of this problem is shown in Fig. 20, and its mathematical model is expressed by Eq. (43).

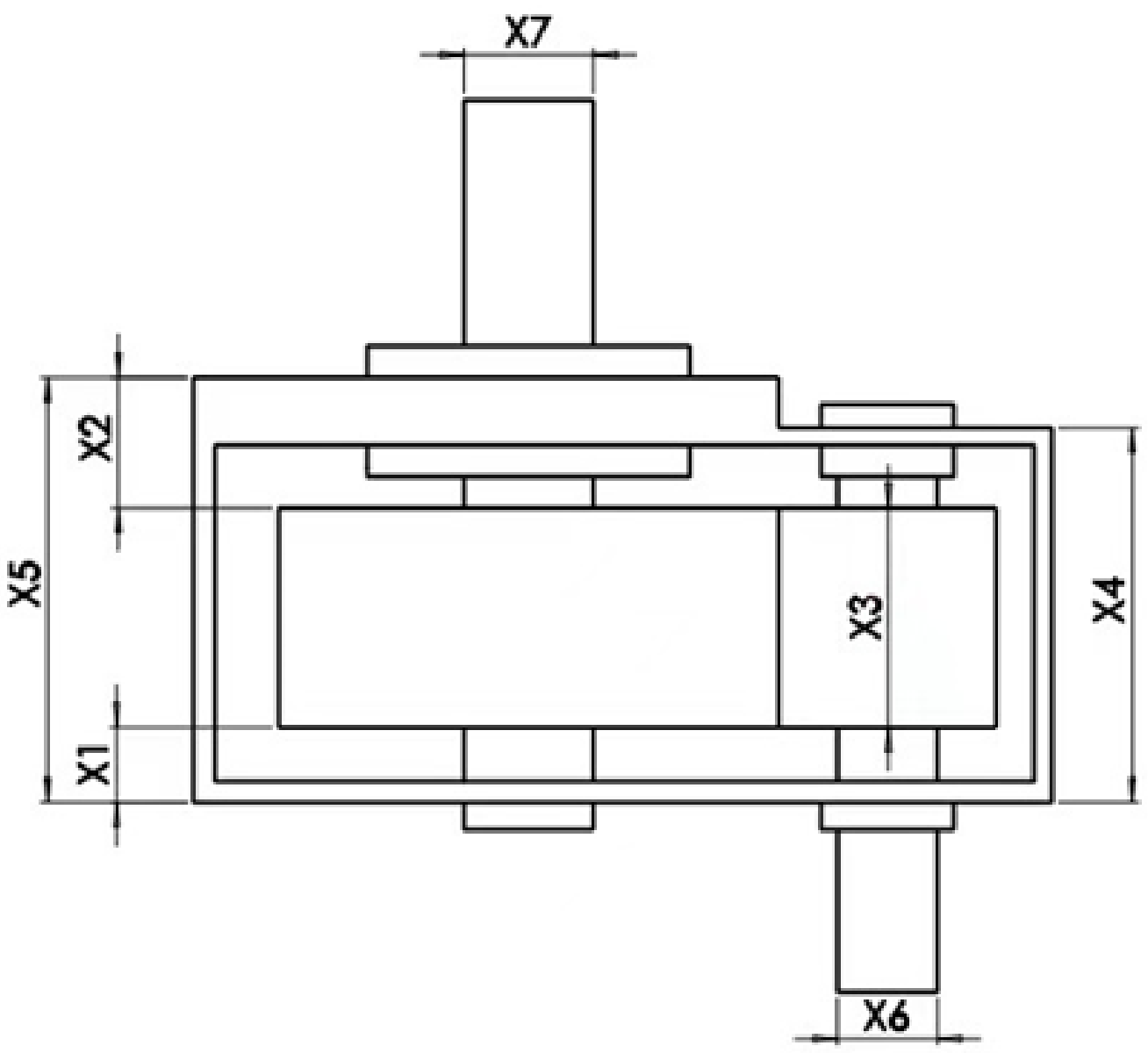

Fig. 20. Schematic representation of the SRD.

Minimize: $f(\vec{x}) = 0.7854x_1x_2^2(3.3333x_3^2 + 14.9334x_3 - 43.0934) - 1.508x_1(x_6^2 + x_7^2) + 7.4777(x_6^3 + x_7^3)$

Subject to:

$$g_1(\vec{x}) = \frac{27}{x_1x_2^2x_3} - 1 \leqslant 0,$$

$$g_2(\vec{x}) = \frac{397.5}{x_1x_2^2x_3^2} - 1 \leqslant 0,$$

$$g_3(\vec{x}) = \frac{1.93x_4^3}{x_2x_3x_6^4} - 1 \leqslant 0,$$

$$g_4(\vec{x}) = \frac{1.93x_5^3}{x_2x_3x_7^4} - 1 \leqslant 0,$$

$$g_5(\vec{x}) = \frac{\sqrt{\left(\frac{745_{x_4}}{x_2x_3}\right)^2 + 16.9 \times 10^6}}{110.0x_6^3} - 1 \leqslant 0, \tag{43}$$

$$g_6(\vec{x}) = \frac{\sqrt{\left(\frac{745_{x_4}}{x_2x_3}\right)^2 + 157.5 \times 10^6}}{85.0x_6^3} - 1 \leqslant 0,$$

$$g_7(\vec{x}) = \frac{x_2x_3}{40} - 1 \leqslant 0,$$

$$g_8(\vec{x}) = \frac{5x_2}{x_1} - 1 \leqslant 0,$$

$$g_9(\vec{x}) = \frac{x_1}{12x_2} - 1 \leqslant 0,$$

$$g_{10}(\vec{x}) = \frac{1.5x_6 + 1.9}{x_4} - 1 \leqslant 0,$$

$$g_{11}(\vec{x}) = \frac{1.1x_7 + 1.9}{x_5} - 1 \leqslant 0$$

Parameters range: $2.6 \leqslant x1 \leqslant 3.6, 0.7 \leqslant x2 \leqslant 0.8, 17 \leqslant x3 \leqslant 28, 7.3 \leqslant x4 \leqslant 8.3, 7.8 \leqslant x5 \leqslant 8.3,$
$2.9 \leqslant x \leqslant 3.9, 5.0 \leqslant x7 \leqslant 5.5.$

Table 22 displays the optimization results of each competitor for the reducer design problem. From the table, it can be clearly seen that MISO achieves the optimal value in each index, which is significantly different from all optimizers, so its performance is better than other competitive algorithms.

Table 22
Comparison results for SRD.

| Algorithms | Best | Median | Worst | Ave | Std | Friedman ranking | Wilcoxon |
|---|---|---|---|---|---|---|---|
| GWO | 3000.509207 | 3010.817317 | 3019.159648 | 3010.801637 | 5.063049 | 6 | (+) |
| SCA | 3056.071704 | 3144.748723 | 3213.212451 | 3142.617151 | 41.885023 | 8 | (+) |
| WOA | 3015.992791 | 3218.160521 | 5294.625470 | 3518.206018 | 631.222432 | 9 | (+) |
| HHO | 3004.377792 | 3781.523647 | 5349.911843 | 3882.196614 | 728.317752 | 10 | (+) |
| MFO | 2994.424466 | 2994.434756 | 3033.701596 | 3004.764673 | 16.624255 | 2 | (+) |
| AVOA | 2994.736280 | 2999.738062 | 3014.448806 | 3001.528115 | 5.717162 | 5 | (+) |
| CSA | 2994.426453 | 2996.261591 | 3034.733704 | 2999.073703 | 8.448769 | 4 | (+) |
| CPSOGSA | 3.1470E+12 | 1.4773E+14 | 1.6222E+15 | 2.2030E+14 | 3.1292E+14 | 12 | (+) |
| AO | 3082.812150 | 4315.197115 | 5524.032573 | 4280.490284 | 719.787252 | 11 | (+) |
| DBO | 2994.434466 | 3046.666743 | 3188.264544 | 3056.469800 | 59.081155 | 7 | (+) |
| SO | 2994.454467 | 2994.497141 | 3081.887954 | 2998.747732 | 16.404459 | 3 | (+) |
| MISO | **2994.424458** | **2994.424466** | **3001.413415** | **2995.345574** | **1.979894** | **1** | — |

5.2.6 Three-bar truss design problem (T-bTD)

A common three-bar truss design problem in civil engineering, the objective is to control two parameter variables to minimize the total weight of the structure. The structure is denoted by Fig. 21. Eq. (44) describes its mathematical model.

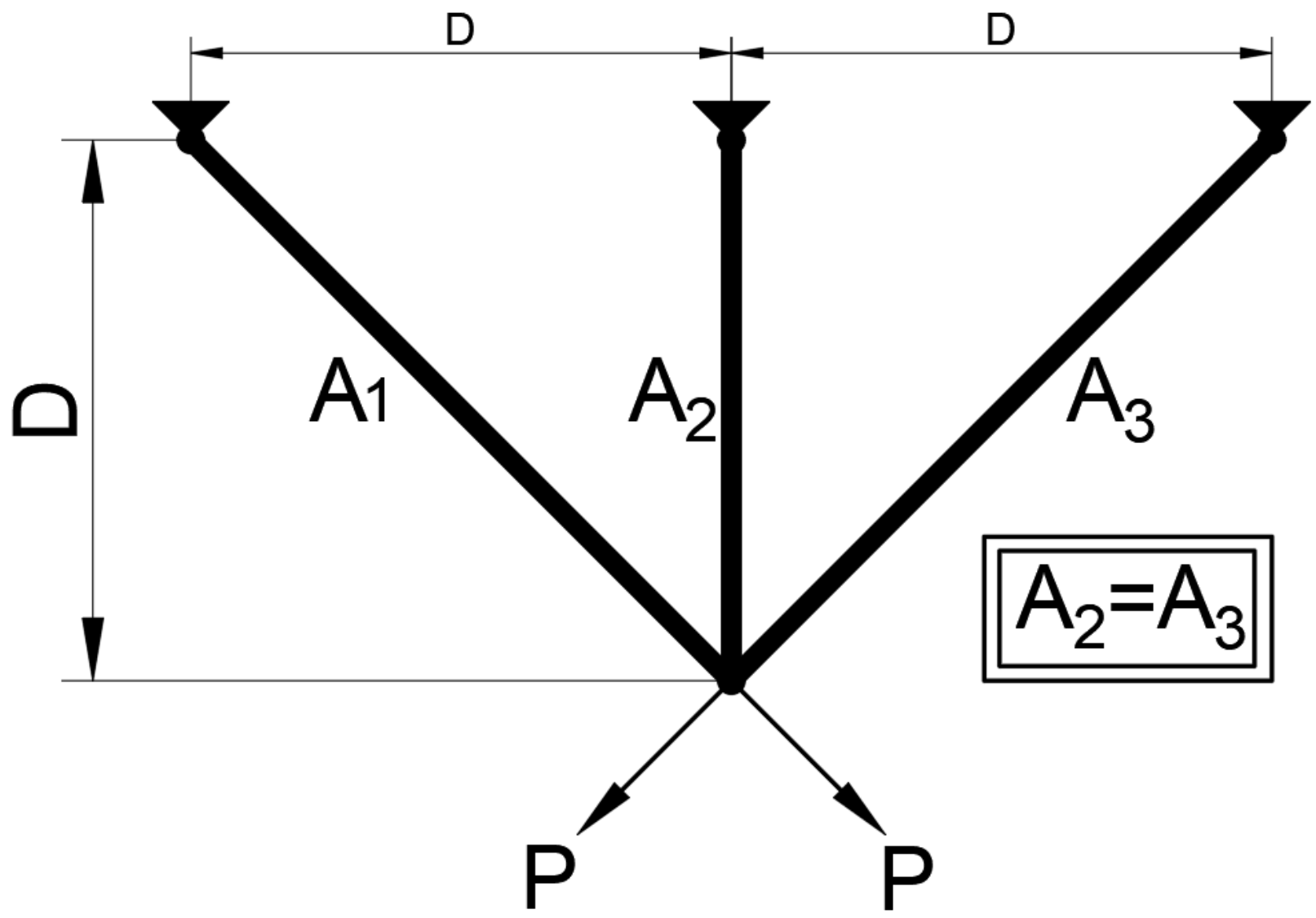

Fig. 21. Schematic representation of the T-bTD.

Consider $\vec{x} = [x_1\ x_2] = [A_1\ A_2],$ (44)
Minimize $f(\vec{x}) = l * (2\sqrt{2x_1} + x_2),$

Subject to

$$g_1(\vec{x}) = \frac{\sqrt{2}x_1+x_2}{\sqrt{2}x_1^2+2x_1x_2}P - \sigma \leqslant 0,$$

$$g_2(\vec{x}) = \frac{x_2}{\sqrt{2}x_1^2+2x_1x_2}P - \sigma \leqslant 0,$$

$$g_3(\vec{x}) = \frac{1}{\sqrt{2}x_2+x_1}P - \sigma \leqslant 0,$$

Parameter range $0 \leqslant x_1, x_2 \leqslant 1,$

Where $l = 100\text{cm}, P = 2\text{KN/cm}^2, \sigma = 2\text{KN/cm}^2.$

Table 23 reports the results for different competitors. The results show that both CSA and MISO can find the minimum cost of the objective function. In addition, MISO also provides data such as minimum median, worst value, mean value and standard deviation, which further confirms the effectiveness and robustness of its algorithm results. Similarly, the results of the Wilcoxon rank-sum test also show that there is a significant difference between MISO and the other algorithms considered.

Table 23

Comparative results for T-bTD.

| Algorithms | Best | Median | Worst | Ave | Std | Friedman ranking | Wilcoxon |
|---|---|---|---|---|---|---|---|
| GWO | 263.897790 | 1.674258 | 263.998555 | 1.694961 | 0.052537 | 6 | (+) |
| SCA | 263.897599 | 1.823262 | 282.842712 | 1.825547 | 0.043367 | 9 | (+) |
| WOA | 263.898213 | 2.194410 | 273.183301 | 2.332473 | 0.531256 | 10 | (+) |
| HHO | 263.895859 | 1.947397 | 264.865290 | 2.081799 | 0.306001 | 8 | (+) |
| MFO | 263.896052 | 1.722715 | 264.344921 | 1.790385 | 0.213321 | 7 | (+) |
| AVOA | 263.895844 | 1.778634 | 264.077144 | 1.764999 | 0.048991 | 5 | (+) |
| CSA | **263.895843** | 1.671058 | 263.915843 | 1.690940 | 0.045546 | 2 | (+) |
| CPSOGSA | 264.294360 | 6.566424 | 301.920143 | 6.780318 | 2.657184 | 12 | (+) |
| AO | 264.066179 | 1.954713 | 271.066605 | 1.977384 | 0.133254 | 11 | (+) |
| DBO | 263.895844 | 1.751091 | 263.924978 | 1.776144 | 0.115144 | 4 | (+) |
| SO | 263.895849 | 1.680505 | 263.925095 | 1.714653 | 0.096261 | 3 | (+) |
| MISO | **263.895843** | **1.670875** | **263.900825** | **1.690198** | **0.043240** | **1** | — |

In Sections 5.2, we validate MISO against other 11 advanced algorithms on six real engineering problems. To show the comparative effects more clearly, we show the ranking of the algorithms in different engineering problems using radar plots. As shown in Fig. 22, the smaller image area of the algorithm represents the better effect of the algorithm in the six engineering problems. It can be seen from the figure that MISO can reach the first in every engineering problem, which indicates that it not only shows good results but also has high stability when solving practical problems. However, the optimization results of the SO algorithm are not ideal. Although the optimization results of SO are better than other algorithms in some engineering problems, it is far less effective than MISO. By improving the local exploration method on the basis of SO, MISO achieves higher search accuracy. Therefore, we can conclude that the improvement strategy carried out for SO is effective and improves the performance of SO to a large extent. The experiments in this section fully verify the applicability of MISO.

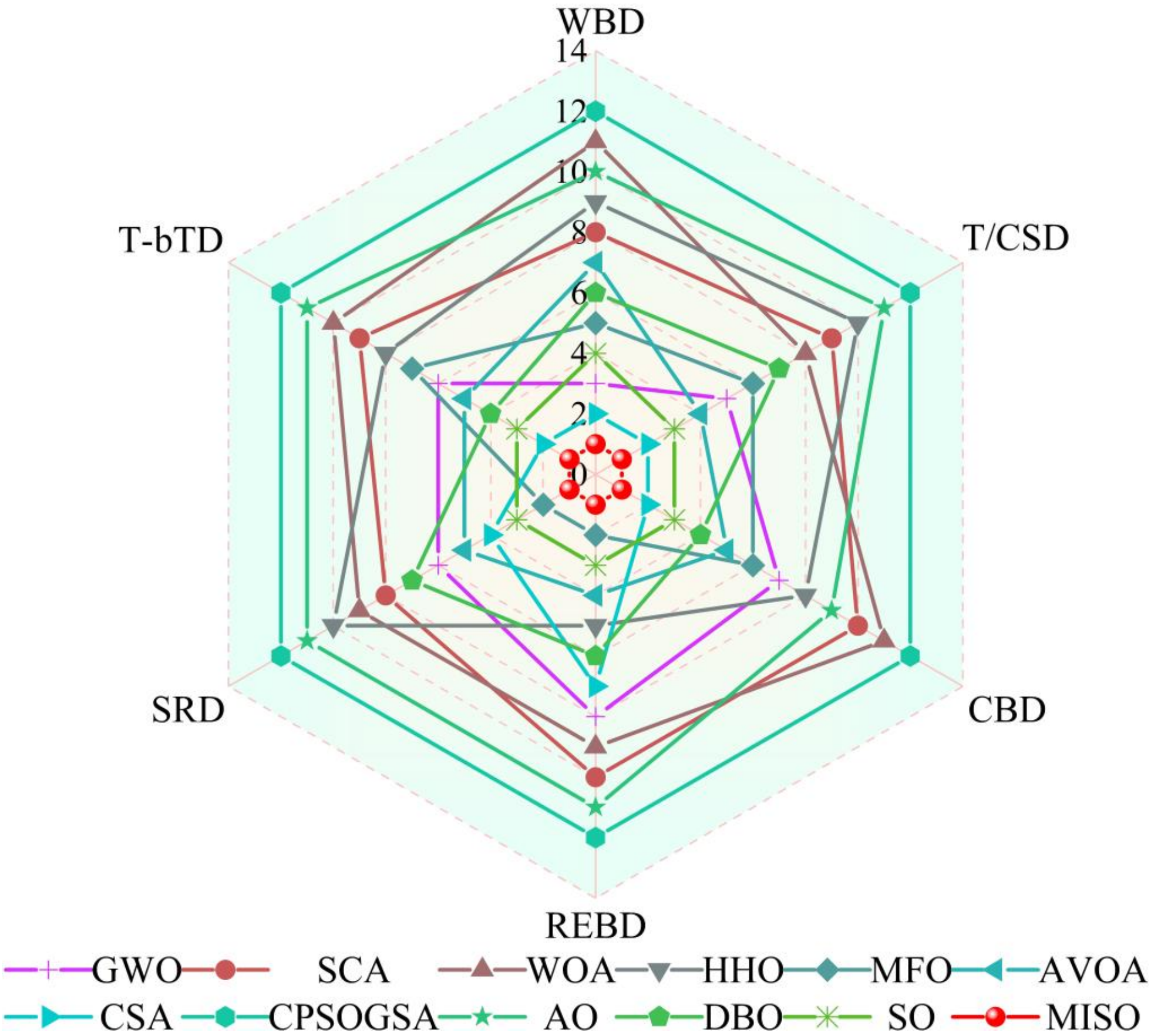

Fig. 22. Comparison of algorithm ranking in 6 engineering problems.

**6. Conclusion and prospect**

In this paper, a Multi-strategy Improved SO (MISO) algorithm is proposed. Aiming at the vulnerability of the SO algorithm to local optima in the iterative process, we propose a new random perturbation strategy based on sine function to enhance the SO, called DSO, which helps the algorithm to jump out of local optimum. The random walk strategy based on the scale factor and the leader enables the male snake leader to obtain flight ability, called LSO, to ensure accuracy while improving convergence speed. Female snake elite with adaptive position update and Brownian motion (BSO) can improve the efficiency of exploring the search space. Furthermore, ablation experiments were executed to demonstrate the validity of the proposed method. The performance of MISO is assessed using 30 CEC2017 and 12 CEC2022 test functions. Compared with other competitive algorithms, MISO displays preferable performance on test functions with different dimensions. We also apply MISO to handle UAV 3D path planning problem and 6 real optimization problems, and perform Wilcoxon and Friedman tests to statistically analyze the experimental results. The statistical analysis confirms that MISO outperforms other competitors and is significantly competitive. Considering the outstanding performance of this algorithm, we intend to apply the enhancement technology of MISO to improve other optimizers in future work. In addition, we will explore population initialization and boundary control methods in the enhancement of MISO. Moreover, we aim to extend the application of MISO to address other practical problems, including fault diagnosis and data mining, extracting the parameters of solar photovoltaic power generation systems and workshop multi-agent task allocation, and point cloud registration, etc.

**Funding**
This research was funded by the Startup Fund for Advanced Talents of Putian grant number 2023141, Scientific Research (on Science and Technology) Projects for Young and Middle-aged Teachers in Fujian grant number JZ240055.

**Data Availability**
All data generated or analysed during this study are included in this published article.

**Conflict of Interest**
The authors declare that they have no conflict of interest.

**Competing Interests**
The authors have no relevant financial or non-financial interests to disclose.

**Ethical Approval**
His article does not contain any studies with human participants or animals performed by any of the authors.

**Informed Consent**
This article does not contain any studies with human participants. So informed consent is not applicable here.